\documentclass[letterpaper]{article} % DO NOT CHANGE THIS
\usepackage{aaai2026}  % DO NOT CHANGE THIS
\usepackage{times}  % DO NOT CHANGE THIS
\usepackage{helvet}  % DO NOT CHANGE THIS
\usepackage{courier}  % DO NOT CHANGE THIS
\usepackage[hyphens]{url}  % DO NOT CHANGE THIS
\usepackage{graphicx} % DO NOT CHANGE THIS
\def\UrlFont{\rm}  % DO NOT CHANGE THIS
\usepackage{natbib}  % DO NOT CHANGE THIS AND DO NOT ADD ANY OPTIONS TO IT
\usepackage{caption} % DO NOT CHANGE THIS AND DO NOT ADD ANY OPTIONS TO IT
\usepackage{tabularx}
\usepackage{booktabs}
\usepackage{siunitx}
\usepackage{multirow}
\usepackage{makecell}
\usepackage{colortbl}
\usepackage{amssymb}
\usepackage{pifont}
\usepackage[linesnumbered,ruled,vlined]{algorithm2e}
\usepackage{amsmath}
\usepackage{subcaption}
\usepackage{tikz}

\usepackage{dashrule}
\usepackage[most]{tcolorbox}
\usetikzlibrary{shadows, backgrounds}
\usepackage[dvipsnames]{xcolor}

\usepackage[table]{xcolor}
\newcolumntype{Y}{>{\centering\arraybackslash}X} % 居中列

\definecolor{lightblue}{rgb}{1,1,1}
\definecolor{darkgreen}{rgb}{0, 0, 0}
\definecolor{darkred}{rgb}{0,0,0}
\newcommand{\xmark}{\ding{55}} 

\title{\textsc{RareAgents}: Autonomous Multi-disciplinary Team for Rare Disease Diagnosis and Treatment}
\author{
    Xuanzhong Chen\textsuperscript{\rm 1}\equalcontrib,
    Ye Jin\textsuperscript{\rm 2}\equalcontrib,
    Xiaohao Mao\textsuperscript{\rm 1},
    Lun Wang\textsuperscript{\rm 2},
    Shuyang Zhang\textsuperscript{\rm 2}\footnotemark[2],
    Ting Chen\textsuperscript{\rm 1}\thanks{Corresponding authors.}
}
\affiliations{
    \textsuperscript{\rm 1}Tsinghua University,
    \textsuperscript{\rm 2}Peking Union Medical College Hospital\\
    cxz23@mails.tsinghua.edu.cn, tingchen@tsinghua.edu.cn
}

\usepackage{bibentry}
\begin{document}

\maketitle

\begin{abstract}
Rare diseases, despite their low individual incidence, collectively impact around 300 million people worldwide due to the vast number of diseases. The involvement of multiple organs and systems, and the shortage of specialized doctors with relevant experience, make diagnosing and treating rare diseases more challenging than common diseases. Recently, agents powered by large language models (LLMs) have demonstrated notable applications across various domains. In the medical field, some agent methods have outperformed direct prompts in question-answering tasks from medical examinations. However, current agent frameworks are not well-adapted to real-world clinical scenarios, especially those involving the complex demands of rare diseases. To bridge this gap, we introduce \textbf{RareAgents}, the first LLM-driven multi-disciplinary team decision-support tool designed specifically for the complex clinical context of rare diseases. \textit{RareAgents} integrates advanced Multidisciplinary Team (MDT) coordination, memory mechanisms, and medical tools utilization, leveraging Llama-3.1-8B/70B as the base model. Experimental results show that \textit{RareAgents} outperforms state-of-the-art domain-specific models, GPT-4o, and current agent frameworks in diagnosis and treatment for rare diseases. Furthermore, we contribute a novel rare disease dataset, \textsc{MIMIC-IV-Ext-Rare}, to facilitate further research in this field.
\end{abstract}


\section{Introduction}

Rare diseases are defined as disorders with low prevalence, typically affecting fewer than 1 in 2,000 individuals in Europe or fewer than 1 in 1,500 individuals in the United States~\cite{valdez2016public}. Despite their rarity, more than 7,000 rare diseases have been identified, impacting approximately 300 million people worldwide~\cite{nguengang2020estimating}. Rare diseases often present with complex and heterogeneous symptoms that overlap with common diseases. As a result, patients frequently experience several years of misdiagnosis, referred to as a "diagnostic odyssey"~\cite{schieppati2008rare}. Such delays not only limit access to timely and effective treatments but also cause the worsening of the disease. On the other hand, while deep learning models have shown promise in medication recommendation~\cite{shang2019gamenet, yang2021safedrug}, their performance for rare diseases remains suboptimal. Experimental studies on the MIMIC-III and MIMIC-IV datasets~\cite{johnson2016mimic, johnson2023mimic} reveal that current state-of-the-art models for drug recommendation are substantially less effective for rare diseases than common ones~\cite{zhao2024leave}. 

\begin{figure}[t]
	\centering
	\includegraphics[width=7.5cm]{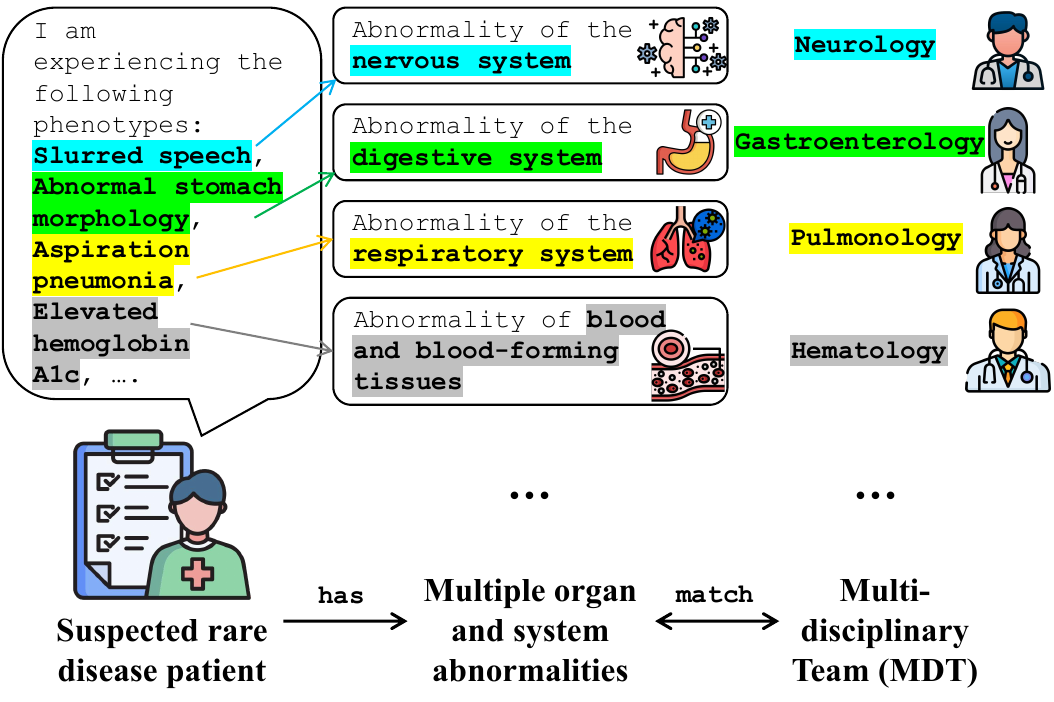}
	\caption{A rare disease patient with multi-organ / multi-system abnormalities necessitates a multi-disciplinary team for comprehensive diagnosis and treatment.}
	\label{fig:mdt}

\end{figure}

Large language models, trained on massive and diverse text corpora, demonstrate remarkable potential across a wide range of natural language interaction tasks~\cite{achiam2023gpt, dubey2024llama}. In particular, LLM-based agents exhibit impressive capabilities in augmented reasoning and problem-solving within complex environments~\cite{wang2024survey}. In the domain of rare diseases, RareBench~\cite{chen2024rarebench} introduced the first benchmark to evaluate LLMs in phenotype extraction and differential diagnosis. Experimental results indicate that advanced LLMs, such as GPT-4~\cite{achiam2023gpt}, can achieve notable diagnostic accuracy under zero-shot settings, even outperforming human specialist physicians for certain rare diseases. 

As shown in Figure~\ref{fig:mdt}, patients with rare diseases often experience symptoms affecting multiple organs and systems, indicating the critical need for expertise from multiple related specialties to achieve accurate diagnosis and personalized treatment plans~\cite{xie2023patient}. In clinical practice, this integrated approach is known as multi-disciplinary team (MDT) care, with the central objective of synthesizing their diagnostic insights and therapeutic proposals from diverse experts to formulate a comprehensive management strategy that prioritizes treatment steps and resolves potential conflicts among recommendations.

Although several multi-agent frameworks have been proposed for general medical applications (as summarized in Table~\ref{tab:comparison}), these methods primarily show improved performance in tasks like multiple-choice question answering (MCQA)~\cite{tang-etal-2024-medagents, jin2024agentmd} and basic question answering (QA)~\cite{kimmdagents}. For these tasks, candidate options are provided or the decision-making is confined to limited and small scopes~\cite{li2024agent}, but these settings differ significantly from the complex real-world clinical scenarios. Moreover, existing approaches tend to emphasize planning capabilities while focusing less on the integration, which can be achieved through memory usage and tool utilization. Additionally, the definition of different agent roles is frequently left to LLMs themselves, leading to potential hallucinations in medical contexts~\cite{lee2023benefits}.

%\begin{table}[t]
%	\centering
%	\scriptsize
%	\setlength{\tabcolsep}{3pt}  % 调整列间距
%	\renewcommand{\arraystretch}{0.92}  % 调整行间距
%	\begin{tabularx}{\columnwidth}{l|c|c|c|l|l}
%		\toprule
%		\textbf{Method} & \makecell[c]{\textbf{Plan-} \\ \textbf{ning}} & \makecell[c]{\textbf{Me-} \\ \textbf{mory}} & \makecell[c]{\textbf{Tool} \\ \textbf{Use}}  & \makecell[c]{\textbf{Multi-Agent} \\ \textbf{Roles}} & \textbf{Medical Scenario} \\ \midrule
%		\makecell[l]{MedAgents \\ \cite{tang-etal-2024-medagents}} & \checkmark & \xmark & \xmark & LLM-generated & Medical MCQA \\
%		\makecell[l]{Agent Hospital \\ \cite{li2024agent}} & \checkmark & \checkmark & \xmark & Pre-defined & Medical MCQA \\
%		\makecell[l]{MDAgents \\ \cite{kimmdagents}} & \checkmark & \xmark & \xmark & LLM-generated & Medical QA / VQA \\
%		\makecell[l]{AgentMD \\ \cite{jin2024agentmd}} & \checkmark & \xmark & \checkmark & Single-Agent & Medical MCQA \\
%		\makecell[l]{\textbf{RareAgents} \\ \textbf{(ours)}} & \checkmark & \checkmark & \checkmark & Pre-defined & \makecell[l]{Complex Clinical QA \\ (Rare Disease \\ Diagnosis+Treatment)} \\ 
%		\bottomrule
%	\end{tabularx}
%	\caption{Characteristics of different medical LLM agent methods: inclusion of planning, memory, and tool usage, along with multi-agent role definition ways and target medical scenarios.}
%	\label{tab:comparison}
%\end{table}

\begin{table}[t]
	\centering
	\small
	\setlength{\tabcolsep}{2pt} % 微调列间距
	\renewcommand{\arraystretch}{0.92}  % 调整行间距
	\begin{tabularx}{\columnwidth}{@{}>{\raggedright}p{1.6cm} *{3}{c} >{\centering\arraybackslash}p{2.1cm} >{\centering\arraybackslash}X@{}} % 精确控制列宽
			\toprule
			\textbf{Method} & \makecell[c]{\textbf{Plan-} \\ \textbf{ning}} & \makecell[c]{\textbf{Me-} \\ \textbf{mory}} & \makecell[c]{\textbf{Tool} \\ \textbf{Use}}  & \makecell[c]{\textbf{Multi-Agent} \\ \textbf{Roles}} & \makecell[c]{ \textbf{Medical} \\ \textbf{Scenario}} \\ \midrule
			\makecell[l]{MedAgents \\ \scriptsize{\cite{tang-etal-2024-medagents}}} & \checkmark & \xmark & \xmark & LLM-generated &  MCQA \\
			\makecell[l]{Agent Hospital \\ \scriptsize{\cite{li2024agent}}} & \checkmark & \checkmark & \xmark & Pre-defined &  MCQA \\
			\makecell[l]{MDAgents \\ \scriptsize{\cite{kimmdagents}}} & \checkmark & \xmark & \xmark & LLM-generated & MCQA / VQA \\
			\makecell[l]{AgentMD \\ \scriptsize{\cite{jin2024agentmd}}} & \checkmark & \xmark & \checkmark & Single-Agent &  MCQA \\
			\makecell[l]{\textbf{RareAgents} \\ \textbf{(ours)}} & \checkmark & \checkmark & \checkmark & Pre-defined & \makecell[l]{Open-ended \\ Complex QA} \\ 
			\bottomrule
		\end{tabularx}
	\caption{Characteristics of different medical LLM agent methods: inclusion of planning, memory, and tool usage, along with multi-agent role definition ways and target medical scenarios.}
	\label{tab:comparison}
\end{table}

To address these challenges, we propose \textbf{RareAgents}, a patient-centered autonomous MDT framework customized for real-world rare disease patients, taking advantage of the planning, memory, and tool-using capabilities of LLM agents. As illustrated in Figure~\ref{fig:overall}, a patient first conveys his personal profile, including symptoms and diagnosis / treatment requests, to an \textbf{Attending Physician Agent}.  Then, this agent assembles an MDT of specialists from a predefined pool of physician agents, designed with dynamic long-term memory and the ability to utilize specialized medical tools. This enhances the performance of LLMs in diagnosing and treating rare diseases, ultimately offering more accurate and personalized medical care for patients.

Overall, our contributions are three-fold: (1) We propose \textbf{RareAgents}, a novel patient-centered multi-disciplinary agent-based framework for enhanced diagnosis and treatment of rare diseases. Each physician agent within \textit{RareAgents} is equipped with dynamic long-term memory, and can effectively utilize a wide range of medical tools, simulating the behavior of a human doctor. Additionally, \textit{RareAgents} is a plug-and-play framework, easily extensible for various medical decision-making scenarios. (2) We evaluate \textit{RareAgents} using Llama-3.1 models (8B and 70B), demonstrating superior diagnostic performance and improved accuracy in medication recommendations compared to state-of-the-art (SOTA) domain-specific models, GPT-4o, and existing medical agent frameworks. We also validate the effectiveness of each module within the \textit{RareAgents} architecture. (3) To the best of our knowledge, this work first extends the medication recommendation task in MIMIC-IV to the LLM agent framework. Furthermore, we compile a rare disease medication recommendation dataset, \textsc{MIMIC-IV-Ext-Rare}, by mapping disease codes and applying rigorous filtering to MIMIC-IV. This dataset contains 4,760 rare disease patients with 18,522 admission records, providing a valuable resource for the rare disease research community.

%%%%%%%%%%%%%%%%%%%%%%%%%%%%%%%%%%%%%%%%%%%%%%%%%%%%%%%%%%%%%%%%%%%%%%%%

\section{Related Work}

\paragraph{LLM-based Agents}
Large language models (LLMs) as agents have demonstrated remarkable capabilities in reasoning and decision-making within complex interactive environments~\cite{liuagentbench}. The concept of generative agents, which first simulated human behavior~\cite{park2023generative}, has evolved into sophisticated frameworks. LLM-based agents are typically composed of three key components: planning~\cite{yao2023react}, memory~\cite{zhong2024memorybank}, and tool-using~\cite{schick2023toolformer}. Existing agent frameworks can be broadly categorized into two paradigms: single and multi-agent systems~\cite{li2023camel}. Among these, role-playing~\cite{shanahan2023role} is a widely adopted approach that assigns agents distinct personalities or roles, allowing them to adapt to specific task scenarios. LLM-based agents have shown significant potential in applications across domains such as education~\cite{zhang2024simulating}, finance, and healthcare~\cite{mehandru2024evaluating}.

\paragraph{Medical Agents}
Med-Palm~\cite{singhal2023large} and Med-Gemini~\cite{saab2024capabilities} have demonstrated promising single-agent capabilities as medical domain LLMs. Beyond this, MedAgents~\cite{tang-etal-2024-medagents} introduces a multi-disciplinary collaboration framework for medical question-answering by leveraging the planning capabilities of multiple agents. MDAgents~\cite{kimmdagents} adaptively adjust to the difficulty of medical questions and extend to visual-question-answering tasks. AI Hospital~\cite{fan2025ai} evaluates the performance of large language models (LLMs) as doctors in symptom collection, examination recommendation, and diagnostic decision-making. Agent Hospital~\cite{li2024agent} creates a virtual hospital environment that simulates task stratification within medical workflows. Furthermore, current applications of medical agents encompass a range of scenarios, including clinical triage~\cite{lu2024triageagent}, electronic health record reasoning~\cite{shi2024ehragent}, and medical imaging analysis~\cite{li2024mmedagent}.

\begin{table*}[ht]
	\centering
	\footnotesize
	\setlength{\tabcolsep}{1mm}
	\renewcommand{\arraystretch}{1.1} % Adjust row separation
		\begin{tabular}{l|cl|c}
			\toprule
			\multicolumn{2}{c|}{\textbf{\textsc{RareBench-Public}}~\cite{chen2024rarebench}}                  & \multicolumn{2}{c}{\textbf{\textsc{MIMIC-IV-Ext-Rare}}~\cite{johnson2023mimic}}                             \\ 
			\midrule
			Type of Clinical Task              & \multicolumn{1}{c|}{Differential Diagnosis} & Type of Clinical Task                & \multicolumn{1}{c}{Medication Recommendation} \\ 
			Patient Data Source             & \multicolumn{1}{c|}{Multi-center}           & Patient Data Source              & \multicolumn{1}{c}{BIDMC of Boston}                     \\
			\# of Rare Disease Patients     & \multicolumn{1}{c|}{1,197}                       & \# of Visits / \# of Rare Disease Patients &                     18,522 / 4,760        \\
			\# of Rare Diseases             & \multicolumn{1}{c|}{498}                       & Disease / Procedure / Medication Space Size &  8,922 / 3,920 / 122                    \\
			Symptom / Disease Space Size & \multicolumn{1}{c|}{17,232 / 9,260}                   & Avg. / Max \# of Visits        &     3.89 / 74                      \\
			Avg. / Max \# of Symptoms per Case & \multicolumn{1}{c|}{12.66 / 96}           & Avg. / Max \# of Diseases per Visit &     16.99 / 39                     \\
			Avg. / Max \# of Diseases per Case & \multicolumn{1}{c|}{1.42 / 26}           & Avg. / Max \# of Procedures per Visit &   2.82 / 32                     \\
			Avg. / Max \# of Cases per Disease & \multicolumn{1}{c|}{3.40 / 148}          & Avg. / Max \# of Medications per Visit &   11.27 / 65                    \\ 
			\bottomrule
		\end{tabular}
	\caption{Statistics of \textsc{RareBench-Public} and \textsc{MIMIC-IV-Ext-Rare} datasets; processing details are in the supplement.}
	\label{table:dataset}
\end{table*}

\paragraph{AI Models for Rare Diseases}
Most AI diagnostic models for rare diseases primarily rely on phenotypic and genotypic information~\cite{javed2014phen, robinson2020lirical}, utilizing statistical and machine learning approaches~\cite{yang2015phenolyzer, kohler2017phenomizer2, zhai2023phen2disease}. RareBERT~\cite{prakash2021rarebert} introduces a Transformer-based model to identify rare disease patients. In the realm of LLMs, dynamic few-shot prompting methods~\cite{chen2024rarebench} have been explored to enhance diagnostic performance. RAREMed~\cite{zhao2024leave} focuses on fairness in drug recommendation systems and proposes a novel approach to improving therapeutic recommendations for rare disease patients. PhenoBrain~\cite{phenobrain} designs a workflow for phenotype extraction and differential diagnosis, enabling an end-to-end diagnostic process based on patients' electronic health records (EHRs).

%%%%%%%%%%%%%%%%%%%%%%%%%%%%%%%%%%%%%%%%%%%%%%%%%%%%%%%%%%%%%%%%%%%%%%%%

\section{Problem Formulation and Datasets}

\subsection{Definition of Rare Disease Tasks}

As indicated in Table~\ref{tab:comparison}, current medical agent frameworks typically formulate tasks as \textbf{multiple-choice questions or limited-answer problems}. However, real-world clinical scenarios are far more complex. To better simulate these conditions, we provide only the patient's profile records $\mathcal{R}$ and ask the agent to make decisions $\mathcal{A}$ based on the specific task demands ($query$). For rare disease diagnosis and treatment, we define the following task scenarios:

\textbf{Differential Diagnosis}\ \ The goal of differential diagnosis for rare diseases is to identify a specific rare disease by distinguishing it from other disorders with similar symptoms. This task focuses on \textbf{phenotype-based differential diagnosis}. Specifically, the patient's profile $\mathcal{R}$ is represented as a set of symptoms ($\{s_n\}$): $\mathcal{R} = \{s_1, s_2, \ldots, s_n\ |\  query=diagnosis\}$. \textbf{No candidate disease list is provided, nor is it explicitly stated that the patient has a rare disease.} The agent relies solely on the symptom information to reason and predict the most likely diagnoses (e.g., the top 10 potential diseases): $\mathcal{A}_{diagnosis} = \{d_1, d_2, \ldots, d_{10}\}$.

\textbf{Medication Recommendation}\ \ This task involves patients who may have multiple admission visits for extended medical treatments. During each visit, the patient’s profile $\mathcal{R}$ comprises a sequence of diagnosed diseases ($\{d_j\}$) and procedures ($\{p_k\}$), along with a full set of available medications $\mathcal{M}$: $\mathcal{R} = \{\{d_i\}_{i=1}^j; \{p_i\}_{i=1}^k; \mathcal{M}\ |\  query=treatment\}$, where $\mathcal{M}$ can include hundreds of drugs (e.g., $|\mathcal{M}|=122$). The objective is to give the optimal combination of medications to match the patient's treatment needs (\textbf{exponential complexity}):  $\mathcal{A}_{treatment} = \{m_1, m_2, \ldots, m_l\}\subset\mathcal{M}$.

\subsection{Datasets}

This research uses two publicly available datasets, RareBench~\cite{chen2024rarebench} and MIMIC-IV~\cite{johnson2023mimic}, for distinct tasks. RareBench is primarily employed for rare disease differential diagnosis, whereas MIMIC-IV supports various medical tasks, including medication recommendation. From MIMIC-IV, we derive MIMIC-IV-Ext-Rare, a specialized dataset for medication recommendations tailored to rare disease patients. Detailed statistics for both datasets are presented in Table~\ref{table:dataset}.

\textbf{\textsc{RareBench-Public}}\ \ RareBench is a multi-center dataset comprising rare disease patient data from Europe, China, and Canada. It is specifically designed to evaluate LLMs' performance in the rare disease domain~\cite{chen2024rarebench}. We utilize 1,197 publicly available cases, each with at least three symptom codes and corresponding diagnostic information extracted from electronic health records (EHRs).

\textbf{\textsc{MIMIC-IV-Ext-Rare}}\ \ MIMIC-IV (version 3.0) contains EHR data from the Beth Israel Deaconess Medical Center (BIDMC) in the United States, spanning 2008 to 2022~\cite{johnson2023mimic}, with disease codes following ICD-9 and ICD-10 standards. We map these codes to rare disease identifiers from OMIM\footnote{\url{https://omim.org/}} and Orphanet\footnote{\url{https://www.orpha.net/}}, extracting patients with multiple hospital admissions while excluding cases with incomplete information. This yields MIMIC-IV-Ext-Rare, a dataset of 4,760 rare disease patients with 18,522 admission EHRs curated for medication recommendation tasks in rare disease contexts.

%%%%%%%%%%%%%%%%%%%%%%%%%%%%%%%%%%%%%%%%%%%%%%%%%%%%%%%%%%%%%%%%%%%%%%%%

\section{Overview of \textsc{RareAgents}}

\begin{algorithm}[h!]
	\caption{Pseudo-code of \textsc{RareAgents}}
	\label{alg:rareagents_mdt}
	\footnotesize
	\begin{small}
		\KwIn{$\mathcal{R}$: patient profile, $APA$: attending physician agent, $\mathcal{SP}$: specialist pool, $\mathcal{T}$: medical tool functions, $R$: the maximum number of rounds}
		\KwOut{Final decision $\mathcal{A}$ for rare disease patient}
		Initialize $f^{c} \leftarrow False$, $r \leftarrow 0$, $\mathcal{O}^{(0)}(\mathcal{R}) \leftarrow \varnothing$ \\
		
		\textcolor{darkgreen}{// \textit{Multi-Disciplinary Team (MDT) 
			formation}}\\
		$\text{MDT} \leftarrow \text{LLM}_{APA}(\mathcal{R}, \mathcal{SP})$ \\
		\textcolor{darkgreen}{// \textit{Specialist physician agents in MDT initialization}} \\
		\ForEach{$s \in \text{MDT}$}{
			$f^{c}_{s} \gets \text{False}$, $\Delta\mathcal{O}^{(0)}_{s} \gets \mathcal{R}$
		}
		\While{not $f^{c}$ and $r < R$}{
			$f^{c} \leftarrow True$ \\
			
			\ForEach{$s \in \text{MDT}$}{
				\If{not $f^{c}_{s}$}{
					\textcolor{darkgreen}{// \textit{Dynamic self-memory retrieval}}\\
					$\mathcal{MR}_{s} \leftarrow \text{LLM}_{s}.retrieve(\mathcal{R})$ \hfill $\triangleright$ Eq. (2)\\
					\textcolor{darkgreen}{// \textit{Medical tools utilization}}\\
					$\mathcal{TR}_{s} \leftarrow \text{LLM}_{s}.useTools(\mathcal{T}(\mathcal{R}))$ \hfill $\triangleright$ Eq. (3) \\
					\textcolor{darkgreen}{// \textit{Opinion synchronization}}\\
					$\mathcal{O}^{(r)}_{s} \leftarrow \text{LLM}_{s}(\Delta\mathcal{O}^{(r)}_{s}, \mathcal{MR}_{s}, \mathcal{TR}_{s})$ \\
					$\Delta\mathcal{O}^{(r+1)}_{s} \leftarrow \text{MEETING}(\mathcal{O}^{(r)}_{s}, \text{MDT})$ \\
					\If{$\Delta\mathcal{O}^{(r+1)}_{s}$ = $\varnothing$}{
						$\mathcal{O}^{(r)}(\mathcal{R}) \leftarrow \mathcal{O}^{(r)}(\mathcal{R})\cup\mathcal{O}^{(r)}_{s}$, $f^{c}_{s} \leftarrow True$
					}
					\Else{
						$f^{c} \leftarrow False$
					}
				}
			}
			$\mathcal{O}^{(r+1)}(\mathcal{R}) \leftarrow \mathcal{O}^{(r)}(\mathcal{R})$, $r \leftarrow r + 1$
		}
		\textcolor{darkgreen}{// \textit{Summary of MDT's final discussion result}} \\
		$\mathcal{DR} \leftarrow \text{LLM}_{APA}.summary(\mathcal{O}^{(r)}(\mathcal{R}))$ \hfill $\triangleright$ Eq. (1) \\
		$\mathcal{MR} \leftarrow \text{LLM}_{APA}.retrieve(\mathcal{R})$ \hfill $\triangleright$ Eq. (2)\\
		$\mathcal{TR} \leftarrow \text{LLM}_{APA}.useTools(\mathcal{T}(\mathcal{R}))$ \hfill $\triangleright$ Eq. (3) \\
		\textcolor{darkgreen}{// \textit{Final Decision}}\\
		$\mathcal{A} \leftarrow \text{LLM}_{APA}(\mathcal{R}, \mathcal{DR}, \mathcal{MR}, \mathcal{TR})$ \hfill $\triangleright$ Eq. (4)\\
		\textcolor{darkgreen}{// \textit{Agents memory update}} \\
		$\text{LLM}_{APA}.update(\mathcal{R}, \mathcal{A})$ \\
		\ForEach{$s \in \text{MDT}$}{
			$\text{LLM}_{s}.update(\mathcal{R}, \mathcal{A})$
		}
		\Return{$\mathcal{A}$}
	\end{small}
\end{algorithm}

\begin{figure*}[ht]
	\centering
	\includegraphics[width=2.0\columnwidth]{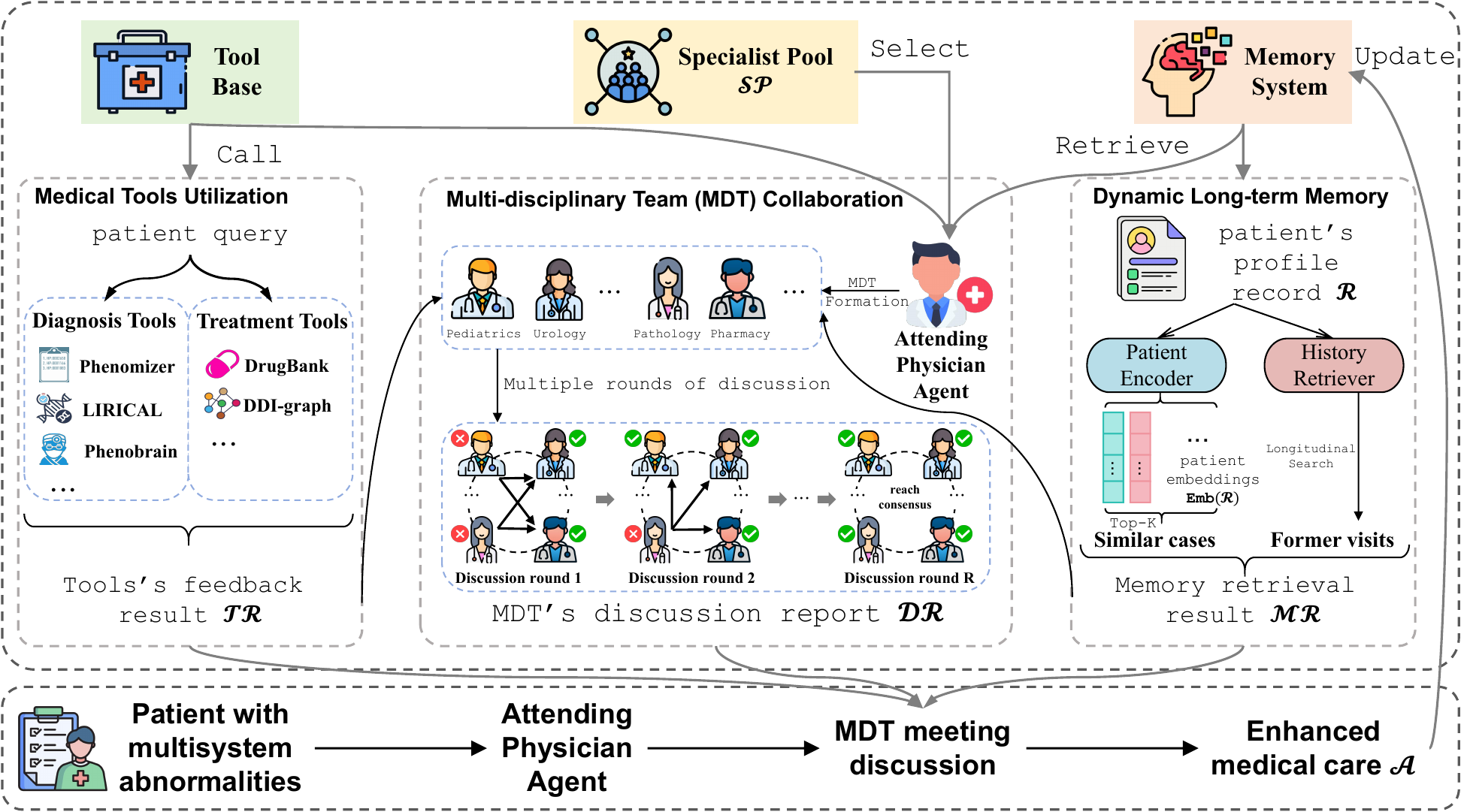}
	\caption{Overview of the \textbf{RareAgents}: For patients with multiple organ / multiple system abnormalities, the Attending Physician Agent selects specialists from a predefined pool to form an MDT, which reaches consensus through iterative discussions. Each physician agent is equipped with: a) Dynamic long-term memory to retrieve and update similar cases or prior records; and b) Tools to assist in diagnosis or treatment decisions.}
	\label{fig:overall}
\end{figure*}

This section introduces the proposed \textbf{RareAgents} framework for rare disease diagnosis and treatment. Algorithm~\ref{alg:rareagents_mdt} and Figure~\ref{fig:overall} provide detailed algorithm pseudo-code and overview of the framework, respectively. The \textit{RareAgents} framework is composed of three core modules: \textbf{(1) Multi-disciplinary Team (MDT) Collaboration:} The attending physician agent selects the most relevant specialists from a predefined specialist pool based on the patient's clinical information to form an MDT. These special physician agents engage in multiple rounds of discussion to reach a consensus on the diagnosis and treatment plan. \textbf{(2) Dynamic Long-term Memory:} Each agent, whether the attending physician or a specialist, maintains a personalized long-term memory. These memories, built from past consultation processes, serve as dynamic experience bases that can be retrieved and updated continuously to assist decision-making. \textbf{(3) Medical Tool Utilization:} Throughout the reasoning process, all physician agents can access and utilize online rare disease diagnostic programs and treatment tools to enhance their decision-making capabilities.

\subsection{Multi-disciplinary Team Collaboration}
Previous implementations of LLM-based MDTs often have the LLMs autonomously define the roles and responsibilities of various specialists~\cite{tang-etal-2024-medagents, kimmdagents}. In contrast, our approach mirrors real-world clinical practice by leveraging specialist departments commonly involved in rare disease cases~\cite{xie2023patient}. Under human specialist physicians' guidance, we constructed a Specialist Pool ($\mathcal{SP}$), which consists of 41 distinct clinical departments. Detailed definitions are provided in the supplementary material. The entire MDT consultation process is divided into three stages: \textbf{(i) MDT Formation} (Line 3-6 of Algorithm~\ref{alg:rareagents_mdt}): The attending physician agent assembles a patient-centric MDT. \textbf{(ii) Expert Consensus} (Line 15-21 of Algorithm~\ref{alg:rareagents_mdt}): Specialist agents within the MDT engage in multi-turn discussions (up to a maximum of $R$ rounds) to reach a consensus opinion $\mathcal{O}(\mathcal{R})$ based on patient's information $\mathcal{R}$. \textbf{(iii) Report Generation} (Line 24 of Algorithm~\ref{alg:rareagents_mdt}): The attending physician agent synthesizes the opinions from all MDT members to generate a final discussion report $\mathcal{DR}$, where
\begin{equation}
	\label{eq:dr}
	\mathcal{DR} = \text{SUMMARY}(\bigcup_{r=0}^{R} \bigcup_{s \in \text{MDT}} \mathcal{O}^{(r)}_{s}(\mathcal{R})).
\end{equation}

\subsection{Dynamic Long-term Memory}

In real-world clinical practice, physicians rely on both personal experience and historical patient records within the healthcare system for decision-making~\cite{trafton2018doctors}. Llama 3.1~\cite{dubey2024llama}, with its expanded context window from 8K to 128K tokens, provides a significantly larger capacity for developing long-term memory. Inspired by these, we design a \textbf{dynamic long-term memory mechanism} for the physician agents in \textit{RareAgents}, enabling them to store, retrieve, and update memories like human physicians (Line 12, 25, 30-32 of Algorithm~\ref{alg:rareagents_mdt}). Agents can facilitate personalized diagnosis and treatment based on historical interactions. For diagnosis, we use the rare disease patient embeddings ($Emb(*)$) from RareBench~\cite{chen2024rarebench} to dynamically retrieve the top-$k$ most similar cases from the patient database. In subsequent experiments, we select $k=5$. For treatment, we leverage the longitudinal nature of patient records in the MIMIC-IV-Ext-Rare dataset, where each patient may have multiple admission records. During the $n$-th admission, the physician agent retrieves the patient’s records from the previous $n-1$ visits. Denote $\mathcal{MR}$ as the result of dynamic long-term memory retrieval, where 
\begin{equation}
	\begin{aligned}
		\mathcal{MR}_{diagnosis}\left( \mathcal{R}\right) &= \arg\max_{\text{Top-K}}\left( Emb(\mathcal{R})\right),\\
		\mathcal{MR}_{treatment}( \mathcal{R}^{(n)}) &= \mathcal{R}^{(1:n-1)}\cup \mathcal{A}_{treatment}^{(1:n-1)}.
	\end{aligned}
	\label{eq:mr}
\end{equation}

\subsection{Medical Tools Utilization}

Physicians frequently use various tools to assist decision-making in clinical practice~\cite{kawamoto2005improving}. Similarly, the physician agents in \textit{RareAgents} have access to diagnostic and therapeutic tools to enhance their clinical reasoning capabilities (Line 14, 26 of Algorithm~\ref{alg:rareagents_mdt}). Llama 3.1’s built-in tool integration and function-calling capabilities enable the agents to interact with external environments dynamically~\cite{dubey2024llama}.  In this research, diagnostic tools include Phenomizer~\cite{kohler2017phenomizer2}, LIRICAL~\cite{robinson2020lirical}, and Phenobrain~\cite{phenobrain}, all of which are accessible via APIs or web interfaces. Therapeutic tools are knowledge bases like DrugBank (for drug information) and DDI-graph (for drug-drug interaction relationships). Detailed tool functions are provided in the supplementary material. Let $\mathcal{T} = \{ T_1, T_2, ... \}$ denote the set of medical tool functions, and $\mathcal{TR}$ represents the aggregated output from the tools' feedback, where 
\begin{equation}
	\label{eq:tr}
	\mathcal{TR} = \text{CONCAT}(\bigcup_{T_{i} \in \mathcal{T}}T_i(\mathcal{R})).
\end{equation} 
Finally, \textit{RareAgents} synthesize the results from MDT consensus, dynamic long-term memory, and tools' feedback to generate the final decision $\mathcal{A}$: 
\begin{equation}
	\label{eq:final_answer}
	\mathcal{A} = \text{LLM}(\mathcal{R}, \mathcal{DR}, \mathcal{MR}, \mathcal{TR}).
\end{equation}

\begin{table*}[t]
	\centering
	\begin{minipage}{0.66\textwidth} % 左侧大表格占0.69宽度
		\centering
		\small
		\setlength{\tabcolsep}{2pt} % 减少列间空白
		\renewcommand{\arraystretch}{0.92} % 压缩行高
			\begin{tabular}{@{}c@{\hspace{0pt}}ccc@{\hspace{1pt}}c|cc@{\hspace{8pt}}c@{\hspace{0pt}}c@{}}
				\toprule
				\multirow{2}{*}{\textbf{Model}} & \multicolumn{4}{c|}{\textbf{Diagnosis} on \textbf{\textsc{RareBench}}} & \multicolumn{4}{c}{\textbf{Treatment} on \textbf{\textsc{MIMIC-IV-Rare}}} \\
				\cmidrule(lr){2-5} \cmidrule(lr){6-9}
				& {\textbf{Hit@1}} & {\textbf{Hit@3}} & {\textbf{Hit@10}} & {\textbf{MR}$\downarrow$} &  {\textbf{Jaccard}} & {\textbf{F1}} & {\textbf{DDI}$\downarrow$} & {\textbf{\#MED}} \\
				\midrule
				%				\rowcolor{gray!20}
				\multicolumn{9}{l}{\textit{\textbf{General \& Medical LLMs (zero-shot CoT)}}} \\
				GPT-4o       &    0.4169        &   \underline{0.5815}         &   0.7068         &   \underline{2.0}         &    \underline{0.3282}        &    \underline{0.4693}        &    0.0907        &    12.10        \\
				GPT-3.5      &     0.3968       &    0.5079        &   0.6007         &   3.0         &   0.2277         &    0.3451        &   0.0856         &    8.72        \\
				UltraMedical-70B & 0.4002 & 0.5639 & 0.6424 & \underline{2.0} & 0.2606 & 0.3922 & 0.0739 & 13.08 \\
				OpenBioLLM-70B & 0.3885 & 0.5388 & 0.6182 & \underline{2.0} & 0.1504 & 0.2465 & \underline{0.0615} & 14.73 \\
				UltraMedical-8B & 0.3425 & 0.4294 & 0.4787 & \textgreater 10 & 0.1613 & 0.2549 & 0.0840 & 9.14 \\
				OpenBioLLM-8B & 0.1495  & 0.1763 & 0.1997 & \textgreater 10 & 0.0997 & 0.1715 & \textbf{0.0519} & 20.96\\
				\midrule
				%				\rowcolor{gray!20}
				\multicolumn{9}{l}{\textit{\textbf{o1-like LLMs (zero-shot)}}} \\
				DS-R1-Distill-Llama-70B & 0.3509 & 0.5221 & 0.6291 & 3.0 & 0.2924 & 0.4267 & 0.0901 & 11.87 \\
				DS-R1-Distill-Llama-8B & 0.3158 & 0.4511 & 0.5171 & 6.0 & 0.2109 & 0.3251 & 0.0803 & 9.05 \\
				Baichuan-M1-14B & 0.3175 & 0.5313 & 0.6241 & 3.0 & 0.2188 & 0.3381 & 0.0734 & 11.36 \\
				HuatuoGPT-o1-70B & 0.3584 & 0.5305 & 0.6232 &  3.0 & 0.2536 & 0.3819 & 0.0837 & 10.57 \\
				\midrule
				%				\rowcolor{gray!20}
				\multicolumn{9}{l}{\textit{\textbf{Llama-3.1-8B-Instruct (Medical Agent framework)}}} \\
				Single-Agent &    0.3041        &   0.4578         &     0.5698       &   5.0         &   0.2104         &    0.3229        &      0.0951      &       9.68     \\
				MedAgents    &    0.3734        &   0.4879         &     0.5698       &    4.0        &   0.2285         &   0.3505         &      0.0997      &    9.60        \\
				MDAgents     &    0.3233        &   0.4453         &     0.5271       &     7.0      &    0.2311        &   0.3539         &   0.0715   & 10.92      \\
				\cellcolor{lightblue}\textbf{RareAgents} (MDT only)     &  0.3826         &   0.5013         &    0.6007        &     3.0       &     0.2376       &    0.3630        &    0.0957        &       11.74     \\
				\cellcolor{lightblue}\textbf{RareAgents}      &  \underline{0.4511}         &   0.5647         &    \underline{0.7377}        &     \underline{2.0}       &     0.3052       &    0.4475        &    0.0820        &       12.98     \\
				\midrule
				%				\rowcolor{gray!20}
				\multicolumn{9}{l}{\textit{\textbf{ Llama-3.1-70B-Instruct (Medical Agent framework)}}} \\
				Single-Agent &    0.3751        &   0.5397         &     0.6658       &    3.0        &      0.2543      &     0.3736       &      0.0907      &   10.97         \\
				MedAgents    &   0.4010         &   0.5163         &    0.6449        &     3.0       &     0.2607       &    0.3905        &   0.0974         &   11.20         \\
				MDAgents     &   0.4042         &    0.5640        &    0.6586        &     \underline{2.0}       &    0.2961        &   0.4349         &      0.0813      &   12.41         \\
				\cellcolor{lightblue}\textbf{RareAgents} (MDT only)     &  0.4177         &   0.5455         &    0.6800        &     \underline{2.0}       &     0.3089       &    0.4468        &    0.0950        &       13.40     \\
				\cellcolor{lightblue}\textbf{RareAgents}      &   \textbf{0.5589}         &   \textbf{0.6867}         &   \textbf{0.7811}         &         \textbf{1.0}   &   \textbf{0.4108}         &  \textbf{0.5563}          &    0.0796        &    13.17        \\
				\bottomrule
			\end{tabular}
%		}
		\caption{Results of various LLMs and medical agent frameworks on the differential diagnosis of RareBench-Public and the medication recommendation of MIMIC-IV-Ext-Rare. \textbf{Bold} indicates best performance, while \underline{underlined} denotes second-best.}
		\label{tab:main_result}
	\end{minipage}
	\hfill
	\begin{minipage}{0.33\textwidth} % 右侧小表格占0.3宽度
		\centering
		\setlength{\tabcolsep}{3pt} % 减少列间空白
		\renewcommand{\arraystretch}{0.92} % 压缩行高
		\footnotesize
			\begin{tabular}{@{\hspace{0pt}}p{1.5cm}cccc@{\hspace{0pt}}}
				\toprule
				\textbf{Model} & \textbf{Hit@1} & \textbf{Hit@3} & \textbf{Hit@10} & \textbf{MR}$\downarrow$  \\
				\midrule
				\small{Phenomizer}   &      0.0844      &    0.2072        &       0.3835     &     \textgreater 10       \\
				\small{LIRICAL}      &   0.1637         &   0.2840         &    0.4152        &    \textgreater 10        \\
				\small{BASE\_IC}      &   0.2047         &   0.3434         &    0.5322        &    8.0        \\
				\footnotesize{Phen2Disease}      &   0.2105         &   0.3266         &    0.5129        &    10.0        \\
				\small{Phenobrain}   &   0.2857         &   0.4670         &  0.6341          &   4.0        \\
				\midrule
				\cellcolor{lightblue}\small{\textbf{RareAgents}} \par % 调整行间距
				{(8B)} &  \underline{0.4511}         &   \underline{0.5647}         &    \underline{0.7377}        &     \underline{2.0} \\
				\cellcolor{lightblue}\small{\textbf{RareAgents}} \par % 调整行间距
				{(70B)} &  \textbf{0.5589}         &   \textbf{0.6867}         &   \textbf{0.7811}         &         \textbf{1.0}        \\
				\bottomrule
			\end{tabular}
%		}
		\caption{Performance of 5 Domain-specific SOTA Models for Diagnosis. }
		\label{tab:diagnosis_baselines}
%		\vspace{1em} % 增加两个小表格之间的间距
		
			\begin{tabular}{@{\hspace{0pt}}p{1.5cm}cccc@{\hspace{0pt}}}
				\toprule
				\textbf{Model} & \textbf{Jacc.} & \textbf{F1} & \textbf{DDI}$\downarrow$ & \textbf{\#MED} \\
				\midrule
				LR      &     0.3564       &     0.5020       &     0.0686       &     8.30       \\
				LEAP    &     0.2959       &     0.4341       &    \textbf{0.0485}        &     5.92       \\
				RETAIN  &    0.3527        &     0.5056       &    0.0626        &    13.08        \\
				G-Bert  &    \underline{0.4030}        &     \underline{0.5554}       &    0.0751        &    14.61        \\
				GAMENet &    0.3731        &     0.5195       &    0.0650        &    10.00        \\
				SafeDrug&    0.3903        &     0.5426       &    0.0733        &    12.88        \\
				COGNet  &    0.3883        &     0.5367       &    0.0751        &    14.28        \\
				MICRON  &     0.3887       &     0.5417       &    0.0729        &    12.91        \\
				MoleRec &    0.3975       &     0.5498       &    0.0714        &    12.15        \\
				RAREMed &    0.3800        &     0.5268       &    \underline{0.0622}        &    8.75        \\
				\midrule
				\cellcolor{lightblue}\small{\textbf{RareAgents}} \par % 调整行间距
				{(70B)} &  \textbf{0.4108}  &  \textbf{0.5563}          &    0.0796        &    13.17 \\
				\bottomrule
			\end{tabular}

		\caption{Performance of 10 Domain-specific SOTA Models for Treatment.}
		\label{tab:treatment_baselines}
	\end{minipage}
\end{table*}

\section{Experimental Setup and Main Results}

\subsection{Evaluation Metrics}

\subparagraph{Differential Diagnosis} The diagnostic task is evaluated using two primary metrics: top-k recall (Hit@k, where k=1, 3, 10) and median rank (MR). Hit@k measures diagnostic accuracy by checking if the actual disease is among the top-k predictions, while MR represents the median position of the correct diagnosis across all cases.

\textbf{Medication Recommendation}\ \ The therapeutic task is assessed with four metrics: Jaccard coefficient (Jaccard), F1-score (F1), Drug-Drug Interaction rate (DDI), and the average number of recommended medications (\#MED). Jaccard measures the overlap between the recommended and ground truth medication sets, normalized by their union. F1 quantifies recommendation precision and recall, with higher values indicating better performance. DDI reflects the frequency of potential adverse interactions among recommended drugs, with lower values indicating safer prescriptions. \#MED evaluates the consistency between the number of recommended medications and those prescribed by clinicians. Detailed formulas of metrics are provided in the supplementary material.

\subsection{Baselines}

\textbf{Domain-specific SOTA models}\ \ For the differential diagnosis task, the domain-specific SOTA models include Phenomizer~\cite{kohler2017phenomizer2}, LIRICAL~\cite{robinson2020lirical}, BASE\_IC, Phen2Disease~\cite{zhai2023phen2disease}, and Phenobrain~\cite{phenobrain}. For the medication recommendation task, we leverage ten models: Logistic Regression (LR), LEAP, RETAIN~\cite{choi2016retain}, G-Bert~\cite{shang2019pre}, GAMENet~\cite{shang2019gamenet}, SafeDrug~\cite{yang2021safedrug}, COGNet~\cite{wu2022conditional}, MICRON~\cite{yang2021change}, MoleRec~\cite{yang2023molerec}, and RAREMed~\cite{zhao2024leave}. Notably, these models for medication recommendation require training on the dataset. We conduct \textbf{5-fold cross-validation} based on the number of patients in MIMIC-IV-Ext-Rare and report the average results. In each fold, 20\% of the data is used as the test set, while the remaining 80\% is split into 80\% training and 20\% validation subsets. Supplementary material provides additional details on these baselines and their configurations.

\textbf{General and Medical LLMs}\ \ General LLMs include GPT-4o and GPT-3.5-turbo-0125. The medical LLMs include OpenBioLLM and UltraMedical, both fine-tuned on medical datasets using Llama-3 (8B and 70B). O1-like LLMs include DeepSeek-R1-Distill-Llama (8B and 70B), Baichuan-M1-14B, and HuatuoGPT-o1-70B. All of these models are evaluated in a zero-shot setting with the temperature parameter set to 0. Non-o1-like LLMs utilize Chain-of-Thought (CoT)~\cite{wei2022chain} to enhance reasoning.

\textbf{Open-Source Medical Multi-Agents}\ \ For open-source medical multi-agent frameworks, we select MedAgents~\cite{tang-etal-2024-medagents} and MDAgents~\cite{kimmdagents}, both implemented initially using GPT-4 APIs. We have adapted them to operate on the local Llama-3.1 models.

\subsection{Main Results}

Table~\ref{tab:main_result},~\ref{tab:diagnosis_baselines}, and ~\ref{tab:treatment_baselines} present the performance of all models on RareBench-Public for differential diagnosis and MIMIC-IV-Ext-Rare for medication recommendation. 

\textbf{Differential Diagnosis}\ \ RareAgents (Llama-3.1-70B) outperform all baselines across all evaluation metrics. Even though RareAgents (Llama-3.1-8B) ranks second in some metrics such as Top-1 Recall (Hit@1), it demonstrates significant improvements over other medical agent frameworks. Interestingly, LLMs' performance already surpasses that of domain-specific SOTA models. Among the fine-tuned LLMs, UltraMedical performs better than the base Llama-3.1, while OpenBioLLM shows a decline in performance. This suggests that fine-tuned models may not generalize well to all medical tasks, because their effectiveness is highly 
dependent on the fine-tuning data and methods.

\textbf{Medication Recommendation}\ \ \textit{RareAgents} (Llama-3-70B) achieves the best performance across all metrics except for DDI. The dataset's inherent DDI and average number of drugs recommended per case (\#MED) are 0.0755 and 11.27, respectively. While OpenBioLLM achieves the lowest DDI rate, it performs poorly in Jaccard and F1. Its higher \#MED indicates a tendency to recommend more irrelevant medications. For other metrics, existing LLMs and multi-agent frameworks remain inferior to the performance of domain-specific SOTA models trained on the dataset. Notably, \textit{RareAgents} demonstrates competitive performance through a plug-and-play framework.

\begin{table*}[ht]
	\centering
	\begin{minipage}{0.69\textwidth} % 左侧大表格占0.69宽度
		\centering
		\footnotesize
		\setlength{\tabcolsep}{2pt} % 全局减少列间空白
		\renewcommand{\arraystretch}{0.92} % 压缩行高
			\begin{tabular}{@{\hspace{0pt}}ll@{\hspace{0pt}}cc@{\hspace{1pt}}c|ccl@{\hspace{0pt}}l@{\hspace{0pt}}}
				\toprule
				\multirow{2}{*}{\textbf{Model}} & \multicolumn{4}{c|}{\textbf{Diagnosis} on \textbf{\textsc{RareBench-Public}}} & \multicolumn{4}{c}{\textbf{Treatment} on \textbf{\textsc{MIMIC-IV-Ext-Rare}}} \\
				\cmidrule(lr){2-5} \cmidrule(lr){6-9}
				& {\textbf{Hit@1}} & {\textbf{Hit@3}} & {\textbf{Hit@10}} & {\textbf{MR}$\downarrow$} & {\textbf{Jaccard}} & {\textbf{F1}} & {\textbf{DDI}$\downarrow$} & {\textbf{\#MED}} \\
				\midrule
%				\rowcolor{gray!20}
				\multicolumn{9}{l}{\textit{\textbf{Llama-3.1-8B-Instruct}}} \\
				\textbf{w/o MDT} &    0.4394 (\textcolor{darkred}{$\downarrow 2.6 \%$})       &   \textbf{0.5973}         &     0.7343       &   \textbf{2.0}         &    0.2856        &    0.4244        &    0.0850 (\textcolor{darkred}{$\uparrow 3.7 \%$})       &   12.91        \\
				\textbf{w/o Memory}     &   0.3952 (\textcolor{darkred}{$\downarrow 12.4 \%$})        &    0.5581        &    0.6951        &        3.0    &    0.2422         &    0.3689        &   \textbf{0.0723} (\textcolor{darkgreen}{$\downarrow 11.8 \%$})       &     12.94      \\
				\textbf{w/o Tools} &  0.4361 (\textcolor{darkred}{$\downarrow 3.3 \%$})         &   0.5113         &    0.7143        &  3.0          &    0.2644         &        0.3951    &    0.1012 (\textcolor{darkred}{$\uparrow 23.4 \%$})       &    11.92       \\
				\cellcolor{lightblue}\textbf{RareAgents}      &  \textbf{0.4511}         &   \underline{0.5647}         &    \textbf{0.7377}        &     \textbf{2.0}       &     \textbf{0.3052}       &    \textbf{0.4475}        &    \underline{0.0820}        &       12.98      \\
				\midrule
%				\rowcolor{gray!20}
				\multicolumn{9}{l}{\textit{\textbf{Llama-3.1-70B-Instruct}}} \\
				\textbf{w/o MDT} &    0.5171 (\textcolor{darkred}{$\downarrow 7.5 \%$})       &    0.6416        &   0.7377         &     \textbf{1.0}       &   0.3828          &    0.5292        &   0.0859 (\textcolor{darkred}{$\uparrow 7.9 \%$})        &   13.04        \\
				\textbf{w/o Memory}    &   0.4336 (\textcolor{darkred}{$\downarrow 22.4 \%$})         &   0.5564         &   0.6976         &       2.0     &   0.3185         &      0.4584      &    0.0884 (\textcolor{darkred}{$\uparrow 11.1 \%$})        &    13.20       \\
				\textbf{w/o Tools}   &    0.5221 (\textcolor{darkred}{$\downarrow 6.6 \%$})       &    0.6558        &   0.7469         &    \textbf{1.0}         &  0.3662           &   0.5090         &   0.0961 (\textcolor{darkred}{$\uparrow 20.7 \%$})        &   13.25        \\
				\cellcolor{lightblue}\textbf{RareAgents}      &   \textbf{0.5589}         &   \textbf{0.6867}         &   \textbf{0.7811}         &         \textbf{1.0}    &    \textbf{0.4108}         &  \textbf{0.5563}          &    \textbf{0.0796}        &    13.17            \\
				\bottomrule
			\end{tabular}
		\caption{Ablation study results for the impact of each module in \textit{RareAgents}.}
		\label{tab:ablation_study}
	\end{minipage}
	\hfill
	\begin{minipage}{0.3\textwidth}
		\centering
		\includegraphics[width=\linewidth]{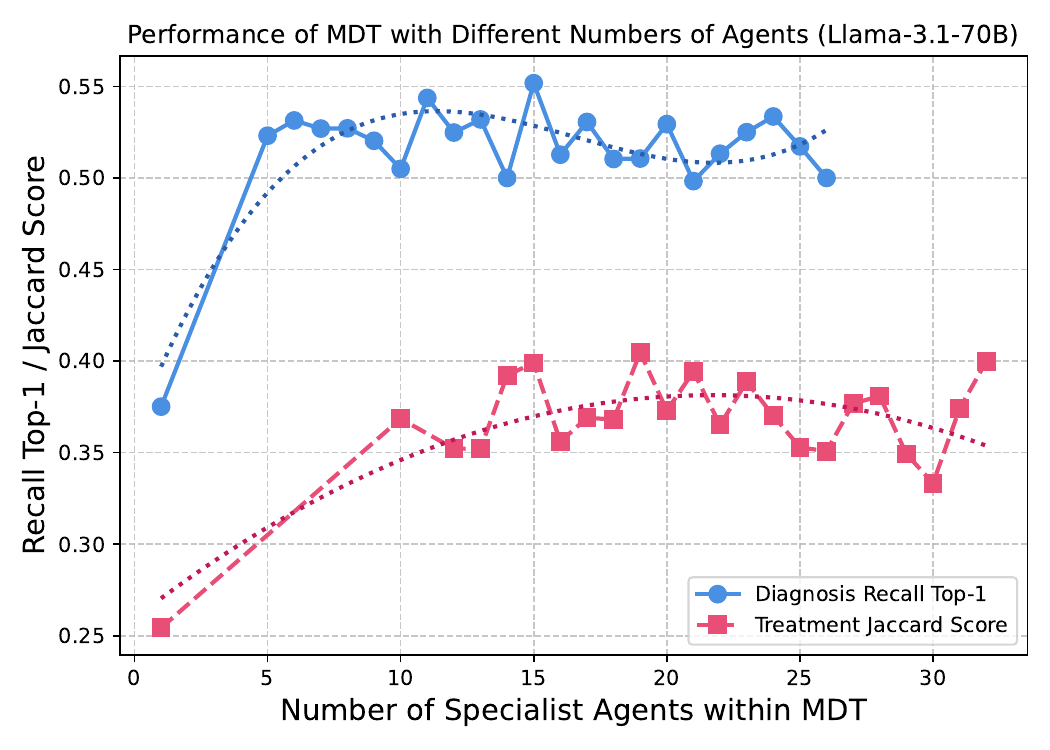}
		\captionof{figure}{Agents Number in MDT.}
		\label{fig:mdt_number}
	\end{minipage}
	\end {table*}
	
\section{Analysis and Discussion}
\label{sec:bibtex}

\begin{figure*}[ht]
	\centering
	\begin{minipage}{0.32\textwidth}
		\centering
		\includegraphics[width=\linewidth]{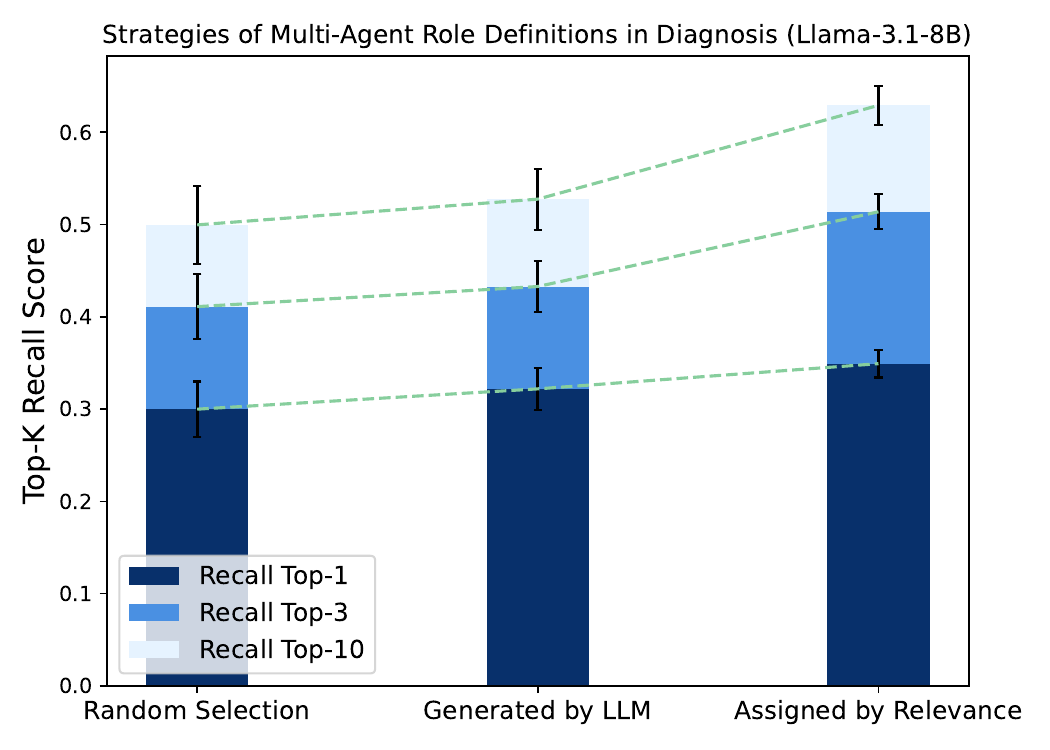}
		\caption{Diagnosis MDT settings.}
		\label{fig:new_diagnosis_mdt}
	\end{minipage}%
	\hfill
	\begin{minipage}{0.32\textwidth}
		\centering
		\includegraphics[width=\linewidth]{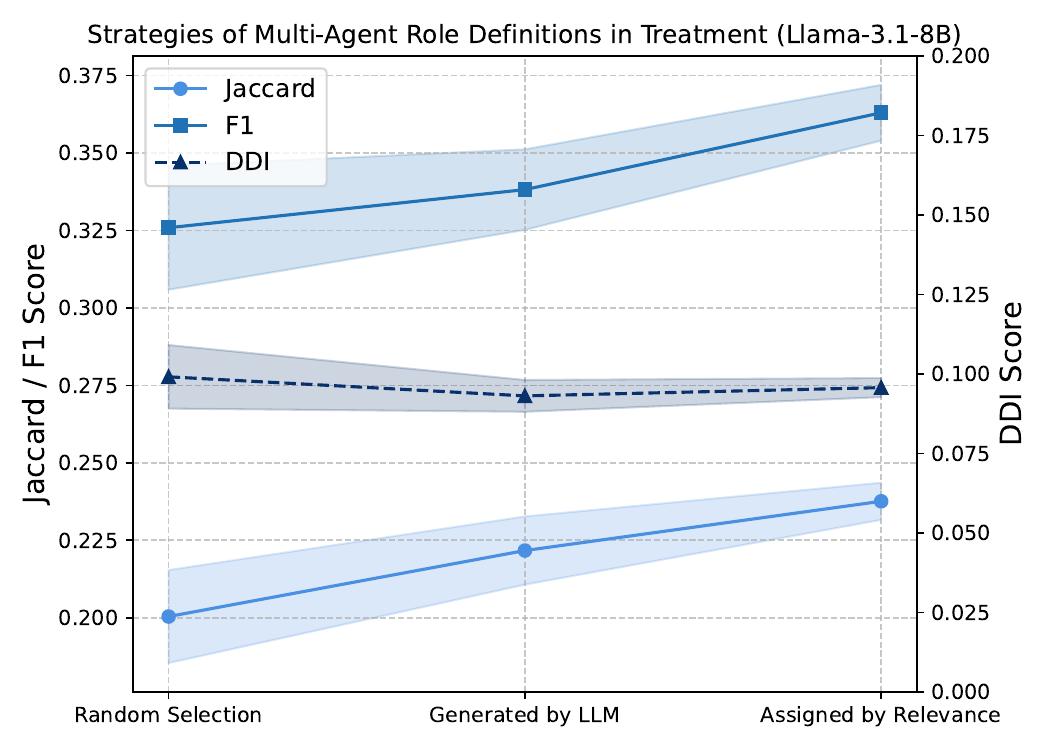}
		\caption{Treatment MDT settings.}
		\label{fig:new_treament_mdt}
	\end{minipage}%
	\hfill
	\begin{minipage}{0.32\textwidth}
		\centering
		\includegraphics[width=\linewidth]{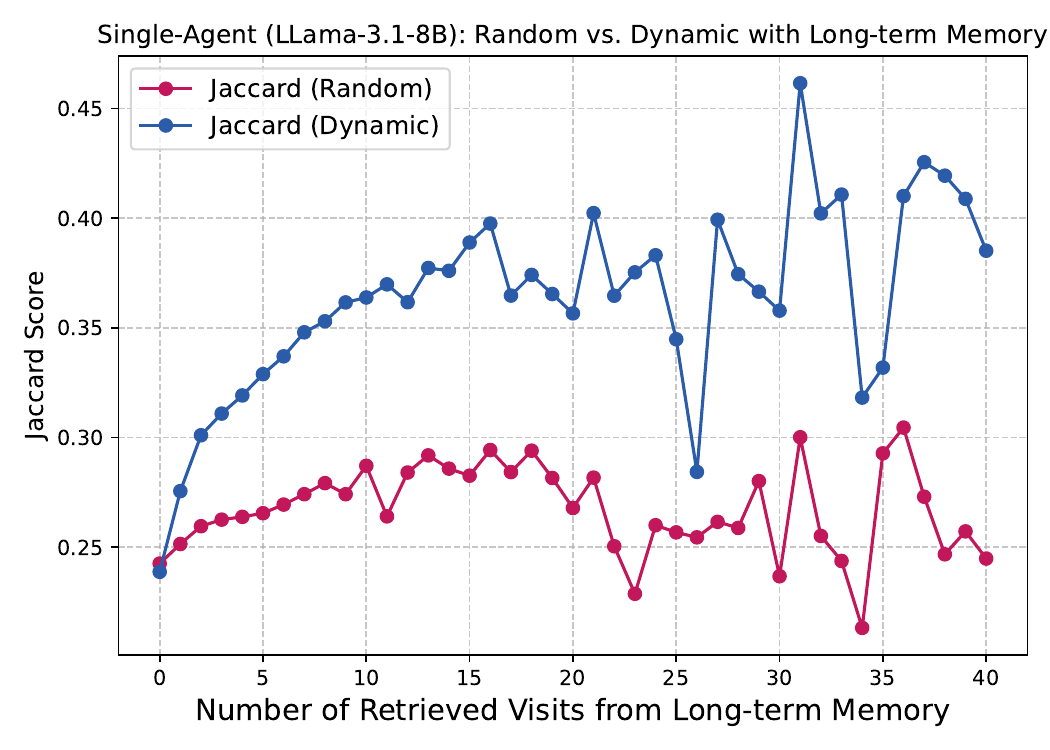}
		\caption{Jaccard by memory settings.}
		\label{fig:treatment_memory_1}
	\end{minipage}%
\end{figure*}

\subsection{Ablation Study}

\textit{RareAgents} consists of three key components:  Multi-disciplinary Team (MDT) collaboration, dynamic long-term memory, and medical tools utilization. To quantify the contribution of each module, we conduct ablation experiments by removing one component at a time, with results shown in Table~\ref{tab:ablation_study}. The findings reveal that removing any single component leads to a performance decline to varying degrees. Among them, when the memory module is removed, the performance drop is most significant. This is attributed to the complexity of rare diseases. The memory module provides the necessary context, helping the LLM distinguish rare conditions from more common ones, thus avoiding the pitfalls of a cold start in reasoning. In the medication recommendation task, the removal of the tools module results in a significant increase in DDI rate. This is because drug knowledge bases deliver specialized pharmacological insights, which effectively reduce DDI and enhance prescription safety.

\subsection{Advanced Nature of MDT in RareAgents}

To further evaluate the efficacy of MDT within \textit{RareAgents}, we conduct experiments isolating the MDT component from the memory mechanisms and external tools. As reported in Table~\ref{tab:main_result}, the configuration \textbf{RareAgents(MDT only)} consistently outperforms other medical agents. This result highlights not only the robustness of the MDT within \textit{RareAgents} but also its pivotal role in navigating the intricate challenges posed by rare diseases. On average, \textit{RareAgents} (Llama-3.1-70B) engage 12.53 specialists for differential diagnosis and 22.22 for medication recommendation. As shown in Figure~\ref{fig:mdt_number}, MDT performance peaks around these agent numbers.

Moreover, we explore three strategies for assigning specialist roles: (1) autonomously generated by the LLM~\cite{tang-etal-2024-medagents, kimmdagents}, (2) randomly selected from a predefined specialist pool,  and (3) assigned based on the most relevant departmental expertise. 
All strategies employ the same number of specialist agents. Figures~\ref{fig:new_diagnosis_mdt} and~\ref{fig:new_treament_mdt} demonstrate that assigning specialists based on departmental relevance consistently outperforms the other two strategies. This advantage arises from the expert-curated role definitions, which are grounded in domain-specific knowledge and enable deeper contextual understanding.

To evaluate the effectiveness of the dynamic retrieval mechanism in long-term memory, we compare it with a baseline that randomly selects the same number of cases. Using the medication recommendation task as an example (Figure~\ref{fig:treatment_memory_1}), dynamic memory mechanism significantly outperforms the baseline, showing that precise contextual relevance matters more than sheer retrieval volume. In contrast, random retrieval yields limited gains, even with more cases. 

\subsection{Case Study}

Given space limitations, we provide detailed case studies and extensive comparisons in the supplementary material.

%%%%%%%%%%%%%%%%%%%%%%%%%%%%%%%%%%%%%%%%%%%%%%%%%%%%%%%%%%%%%%%%%%%%%%%%

\section{Conclusion}

This paper presents \textbf{RareAgents}, a patient-centered framework designed to facilitate personalized diagnosis and treatment for rare diseases through the integration of multidisciplinary team collaboration, dynamic long-term memory, and medical tools. As a plug-and-play framework, \textit{RareAgents} demonstrates superior performance on Llama-3.1 (8B and 70B), surpassing domain-specific state-of-the-art models, general, medical and o1-like LLMs, as well as medical multi-agent frameworks. Furthermore, we contribute MIMIC-IV-Ext-Rare, a curated rare disease patients dataset, providing a valuable resource for future research.

\section*{Acknowledgments}

This study was supported by grants from the National Science Foundation of China (T2541010), the National Key R\&D Program of China (2024YFF1207103), and Beijing National Research Center for Information Science and Technology (BNRist). The funders had no roles in study design, data collection and analysis, publication decisions, or manuscript preparation.

\bibliography{aaai2026}

%File: anonymous-submission-latex-2026.tex
% \documentclass[letterpaper]{article} % DO NOT CHANGE THIS
% \usepackage[submission]{aaai2026}  % DO NOT CHANGE THIS
% \usepackage{times}  % DO NOT CHANGE THIS
% \usepackage{helvet}  % DO NOT CHANGE THIS
% \usepackage{courier}  % DO NOT CHANGE THIS
% \usepackage[hyphens]{url}  % DO NOT CHANGE THIS
% \usepackage{graphicx} % DO NOT CHANGE THIS
\urlstyle{rm} % DO NOT CHANGE THIS
\def\UrlFont{\rm}  % DO NOT CHANGE THIS
\frenchspacing  % DO NOT CHANGE THIS
\setlength{\pdfpagewidth}{8.5in} % DO NOT CHANGE THIS
\setlength{\pdfpageheight}{11in} % DO NOT CHANGE THIS
%
% These are recommended to typeset algorithms but not required. See the subsubsection on algorithms. Remove them if you don't have algorithms in your paper.
%\usepackage{algorithm}
%\usepackage{algorithmic}

% new
% \usepackage{tabularx}
% \usepackage{booktabs}
% \usepackage{siunitx}
% \usepackage{multirow}
% \usepackage{makecell}
% \usepackage{colortbl}
% \usepackage{amssymb}
% \usepackage{pifont}
% \usepackage[linesnumbered,ruled,vlined]{algorithm2e}
% \usepackage{amsmath}
% \usepackage{subcaption}
% \usepackage{tikz}

% \usepackage{dashrule}
% \usepackage[most]{tcolorbox}
\usetikzlibrary{shadows, backgrounds}

\definecolor{lightblue}{rgb}{1,1,1}
\definecolor{darkgreen}{rgb}{0, 0, 0}
\definecolor{darkred}{rgb}{0,0,0}
% \newcommand{\xmark}{\ding{55}} 

% These are are recommended to typeset listings but not required. See the subsubsection on listing. Remove this block if you don't have listings in your paper.
%\usepackage{newfloat}
%\usepackage{listings}
%\DeclareCaptionStyle{ruled}{labelfont=normalfont,labelsep=colon,strut=off} % DO NOT CHANGE THIS
%\lstset{%
%	basicstyle={\footnotesize\ttfamily},% footnotesize acceptable for monospace
%	numbers=left,numberstyle=\footnotesize,xleftmargin=2em,% show line numbers, remove this entire line if you don't want the numbers.
%	aboveskip=0pt,belowskip=0pt,%
%	showstringspaces=false,tabsize=2,breaklines=true}
%\floatstyle{ruled}
%\newfloat{listing}{tb}{lst}{}
%\floatname{listing}{Listing}
%
% Keep the \pdfinfo as shown here. There's no need
% for you to add the /Title and /Author tags.
\pdfinfo{
/TemplateVersion (2026.1)
}

\setcounter{secnumdepth}{0} %May be changed to 1 or 2 if section numbers are desired.

\section{Limitations and Future Work}
In this study, we utilize pre-extracted symptoms and clinical codes from patients' original electronic health records (EHRs), focusing solely on textual data. While this approach provides valuable insights, it is important to acknowledge that the diagnosis and treatment of rare diseases often benefit from integrating multimodal data, such as medical imaging and genotypic information. Although \textit{RareAgents} is designed as a flexible, plug-and-play framework that supports easy integration and adaptation, the current implementation does not include domain-specific fine-tuning of the underlying large language models (LLMs) for medical or rare disease contexts. In future work, we plan to address these limitations by incorporating more comprehensive patient data and performing targeted fine-tuning on state-of-the-art open-source LLMs, to achieve more accurate diagnostic and therapeutic outcomes.

\section{Ethical Considerations}
Although \textit{RareAgents} has demonstrated promising performance in tasks related to differential diagnosis and medication recommendations for rare diseases on real-world datasets such as RareBench and MIMIC-IV, further validation on external datasets and feedback from medical professionals are essential. Additionally, it is important to acknowledge the inherent limitations of LLMs, including potential biases and hallucinations in their outputs. Therefore, it is critical to emphasize that LLMs should currently be viewed only as supplementary tools. For clinical decision-making, guidance from qualified medical professionals is indispensable, especially for specific diagnostic or treatment decisions.

\section{Data Appendix}
In our research, we utilize two public datasets, adhering to the highest ethical standards. Both datasets are free from content that could compromise patient privacy or disclose personally identifiable information. The RareBench dataset~\cite{chen2024rarebench} is released under the Creative Commons Attribution 4.0 International License (CC BY). MIMIC-IV~\cite{johnson2023mimic} is governed by the PhysioNet Credentialed Health Data License 1.5.0. The MIMIC-IV-Ext-Rare dataset, derived from MIMIC-IV, is strictly intended for research purposes and will be shared on PhysioNet under the same licensing terms as the original dataset.

\section{Data Processing Details}

Due to MIMIC-IV's official requirements, MIMIC-IV-Ext-Rare is currently undergoing the necessary review process on the PhysioNet platform and will be available soon. To ensure full reproducibility, we have included detailed data processing steps alongside our submitted code. These steps are outlined below:
\begin{enumerate}
	\item ORPHA-ICD-10 Mapping: We used the en\_product1.json file from the Orphadata website. From this file, we identified mappings between ICD-10 and ORPHA codes where the label was "E (Exact mapping: the two concepts are equivalent)." This step allowed us to extract the relevant ORPHA codes.
	\item Rare Disease Selection: We leveraged the comprehensive list of 9,260 rare diseases from RareBench~\cite{chen2024rarebench}, and supplemented it with mappings to equivalent OMIM and ICD-9 codes (as the initial step focused on ICD-10).
	\item Patient Filtering: Inspired by the approach in RAREMed~\cite{zhao2024leave}, we filtered patients by retaining only those with at least two hospital admissions to ensure a more robust representation of patient history.
\end{enumerate}

\section{Case Study}
\label{sec:app_case_study}
This section provides two concrete examples demonstrating the differences in performance between Single-Agent and \textit{RareAgents}, both based on the Llama-3.1-70B model.

\subsubsection{Case 1 (Differential Diagnosis)} 
The first example involves a patient diagnosed with \textbf{Brugada Syndrome}. The Single-Agent (Llama-3.1-70B) performs a step-by-step analysis of the patient's symptoms, systematically categorizing them. However, despite its detailed analysis, it fails to include the correct diagnosis within its top 10 predicted diseases. In contrast, \textit{RareAgents} (Llama-3.1-70B) leverages a multidisciplinary team discussion approach, dynamically retrieving similar patient cases from long-term memory and diagnostic tools' feedback. As a result, it successfully identifies the correct diagnosis as the top prediction.

\subsubsection{Case 2 (Medication Recommendation)}

The second example focuses on a patient requiring medication recommendations from the MIMIC-IV-Ext-Rare dataset. The Single-Agent (Llama-3.1-70B) produces a list of 21 recommended drugs, of which only 8 align with the actual prescriptions made by physicians, missing 13 critical medications. In contrast, \textit{RareAgents} (Llama-3.1-70B), by incorporating multidisciplinary team discussions and dynamically retrieving prior patient visit records and treatment tools' feedback, achieves a significantly better result. It recommends 17 correct medications, with only 4 missing from the final recommendations.

\begin{table*}[h!]
	\centering
	
	\resizebox{\textwidth}{!}{%
		\begin{tabular}{llll}
			\toprule
			Pediatrics & Nuclear Medicine           & Pathology                & Nephrology             \\ 
			Urology    & Hepatobiliary Surgery      & Neurology                & Oncology               \\ 
			Hematology & Plastic Surgery            & Obstetrics and Gynecology & General Practice       \\ 
			Radiology  & Interventional Radiology   & Ophthalmology            & Gastroenterology       \\ 
			Neurosurgery & Cardiology               & General Surgery          & Infectious Diseases    \\ 
			Rheumatology & Thoracic Surgery         & Dermatology              & Rehabilitation Medicine \\ 
			Psychiatry & Clinical Nutrition         & Geriatrics               & Pharmacy               \\ 
			Pulmonology & Vascular Surgery          & Orthopedic Surgery       & Ultrasound Medicine    \\ 
			Dentistry  & Anesthesiology             & Cardiac Surgery          & Otorhinolaryngology    \\ 
			Endocrinology  & Laboratory Medicine    & Traditional Chinese Medicine & Breast Surgery       \\ 
			Allergy and Immunology & ~              & ~                        & ~                      \\ 
			\bottomrule
		\end{tabular}%
	}
	\caption{List of 41 Departments from Specialist Pool $\mathcal{SP}$.}
	\label{tab:specialist_departments}
\end{table*}

\section{Predefined Specialist Pool}
\label{sec:app_sp}

Under the guidance of human specialist physicians, we established a \textbf{Predefined Specialist Pool} ($\mathcal{SP}$) consisting of 41 different medical specialties, as listed in Table~\ref{tab:specialist_departments}. Specifically, each specialist agent is characterized by two primary attributes: \textbf{description} and \textbf{system\_message}. When forming a multidisciplinary team (MDT), the attending physician agent selects specialists based on the content of their descriptions, while the system\_message serves as the system prompt for the chosen agent.

\tikzset{
	mybox/.style={
		draw=black, 
		fill=blue!5, % 淡蓝色背景
		very thick, 
		rectangle, 
		rounded corners, 
		inner sep=10pt, 
		drop shadow % 添加阴影效果
	},
	fancytitle/.style={
		fill=blue!70, 
		text=white, 
		font=\bfseries, % 加粗字体
		rounded corners, 
		inner xsep=10pt, 
		inner ysep=3pt
	}
}

\begin{figure}[h]
	\centering
	\begin{tikzpicture}
		\node [mybox] (box){
			\begin{minipage}{0.9\columnwidth} % Restrict to one column width
				\footnotesize
				
%				\setstretch{1.3} % Line spacing
				
				\begin{flushleft}
					\textbf{\textless description\textgreater }
					
					A Pediatrician specializes in the medical care of infants, children, and adolescents, focusing on their healthy growth, development, and managing childhood illnesses. 
					
					\textbf{\textless  system\_message\textgreater }
					
					As a Pediatrician, you are dedicated to the medical care of infants, children, and adolescents. Your expertise involves diagnosing and treating a wide range of illnesses and conditions specific to younger patients. Your role is crucial in promoting healthy growth and development, providing preventative care, and offering guidance to parents on the well-being of their children. Demonstrate proficiency in pediatric assessments, immunizations, and managing both acute and chronic conditions. Communicate effectively with children and their families to ensure a comprehensive understanding of treatment plans and health maintenance strategies.

				\end{flushleft}
			\end{minipage}
		};
		\node[fancytitle, rounded corners, right=10pt] at (box.north west) {\includegraphics[width=0.35cm]{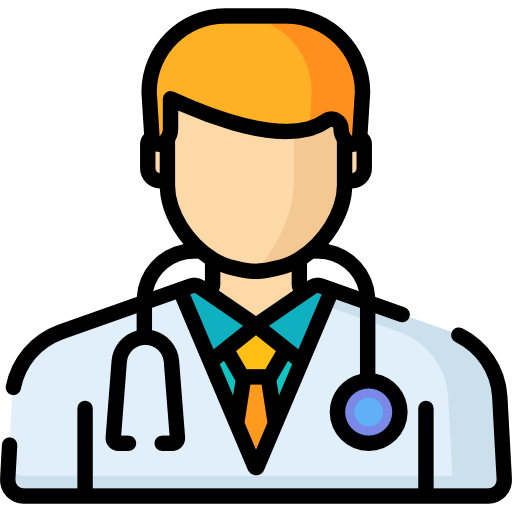} \ Pediatrics };
		\label{fig:prompt_template1}
	\end{tikzpicture}
	\caption{Example of Agent Role for Pediatrics in the Specialist Pool.}
	\label{fig:pediatrics_example}
	\vspace{12pt}
\end{figure}

Figure~\ref{fig:pediatrics_example} illustrates an example of a pediatric specialist agent, while detailed definitions for the remaining 40 specialties can be found in the code repository. 

\section{Medical tools}
\label{sec:app_mt}

\subparagraph{Phenomizer}Phenomizer
% \footnote{\url{https://compbio.charite.de/phenomizer/}} 
is a sophisticated web-based application designed to assist clinicians in the differential diagnosis of rare genetic diseases. We simulated interactions with the Phenomizer platform using a Python script. Specifically, the script selects "symmetric" as the similarity measure from the menu, sequentially inputs the symptom codes (The Human Phenotype Ontology codes), and retrieves the diagnostic results. For each predicted disease, the output includes an associated p-value. In our framework, \textit{RareAgents} utilizes a tool function to obtain the top 10 diseases predicted by Phenomizer.

\subparagraph{LIRICAL} LIRICAL
% \footnote{\url{https://lirical.readthedocs.io}}
(LIkelihood Ratio Interpretation of Clinical AbnormaLities) is a locally deployable tool that we executed using Java 17. It provides posterior probabilities for predicted diseases. Similarly, \textit{RareAgents} retrieves the top 10 diagnostic results generated by LIRICAL through a function call.

\subparagraph{Phenobrain} Phenobrain
% \footnote{\url{https://www.phenobrain.cs.tsinghua.edu.cn/pc}}
is a rare disease diagnostic tool available both as a web application and through an API. For each predicted disease, it provides a corresponding prediction score. \textit{RareAgents} interacts with Phenobrain via a dedicated API-access function to retrieve the top 10 predicted diseases.

\subparagraph{DrugBank} DrugBank is a comprehensive, freely accessible online database that provides detailed information on drugs and their targets, including chemical properties, pharmacology, mechanisms of action, and molecular biology data. In the medication recommendation task, for drugs appearing in the candidate list, \textit{RareAgents} retrieves more detailed information about each drug from DrugBank via function calls.

\subparagraph{DDI-Graph} DDI-Graph (Drug-Drug Interaction Graph) is a graph-based model representing drug-drug interactions (DDIs), where drugs are represented as nodes and interactions as edges. \textit{RareAgents} can query this graph data via function calls to retrieve pairwise interaction relationships between drugs. 
We utilize the same version of DrugBank and DDI-Graph as employed by RAREMed~\cite{zhao2024leave}.

\section{Metric Formulation}
\label{sec:app_metric}

\subparagraph{Top-k Recall} In the differential diagnosis task, let $D_{ground\_truth}$ and $D_{predicted}$ represent the actual diagnoses provided by physicians and the list of diseases predicted by the model, respectively. The Top-k Recall is defined as 1 if and only if $D_{ground\_truth} \in D_{predicted}^{[:Top-K]}$. When there are multiple ground truth diagnoses, only the highest-ranked one in the predicted list is considered.

\subparagraph{Jaccard coefficient} In the medication recommendation task, let $M_{ground\_truth}$ and $M_{predicted}$ represent the actual recommendations provided by physicians and the predictions made by the model, respectively. The Jaccard score is computed as follows:

\begin{equation*}
	\label{eq:jaccard}
	Jaccard = \frac{|M_{ground\_truth} \cap M_{predicted}|}{|M_{ground\_truth} \cup M_{predicted}|}.
\end{equation*}

\subparagraph{F1-score} The recall and precision are formulated as:
\begin{equation*}
	\label{eq:recall}
	Recall = \frac{|M_{ground\_truth} \cap M_{predicted}|}{|M_{ground\_truth}|},
\end{equation*}
\begin{equation*}
	\label{eq:precision}
	Precision = \frac{|M_{ground\_truth} \cap M_{predicted}|}{|M_{predicted}|}.
\end{equation*}
The F1-score is then calculated as:
\begin{equation*}
	\label{eq:f1}
	F1 = \frac{2\cdot Recall\times Precision}{Recall + Precision}.
\end{equation*}

\subparagraph{DDI rate} Let $A_{ddi}$ denote the adjacency matrix of the DDI-Graph, where $A_{ddi}^{(jk)} = 1$ indicates the presence of an interaction between drugs $j$ and $k$, and $A_{ddi}^{(jk)} = 0$ indicates no interaction. The DDI rate score is calculated as follows: 
\begin{equation*}
	\label{eq:ddi}
	DDI = \frac{\sum_{j,k\in  M_{predicted}}A_{ddi}^{(jk)}}{\sum_{j,k\in  M_{predicted}}1}.
\end{equation*}

\subparagraph{\#MED} Let $N_p$ denote the total number of patients,  $N_v^{(i)}$ represent the number of visits for the $i$-th patient, and $M_{predicted}^{(i, j)}$ indicate the set of medications recommended by the model for the $j$-th visit of the $i$-th patient. Then,
\begin{equation*}
	\label{eq:med}
	\#MED = \frac{\sum_{i=1}^{N_p} \sum_{j=1}^{N_v^{(i)}} |M_{predicted}^{(i, j)}|}{\sum_{i=1}^{N_p}\sum_{j=1}^{N_v^{(i)}} 1}.
\end{equation*}

\section{Evaluation Details}

For the medication recommendation task, since a predefined list of 122 candidate drugs is provided, the large language model's outputs are matched against drug names using regular expressions for precise identification.

For the differential diagnosis task, since no information about candidate diseases is provided, the large language model's outputs may differ in expression (e.g., abbreviations or aliases) from the standard diagnoses provided by physicians. To address this, we use GPT-4o as an external evaluation model to assess whether the standard diagnosis appears within the top 10 predicted diseases. If the standard diagnosis is found, GPT-4o also identifies the corresponding rank. The evaluation prompt is as follows:

\tikzset{
	mybox/.style={draw=black, very thick, rectangle, rounded corners, inner sep=10pt, inner ysep=13pt},
	fancytitle/.style={fill=black, text=white, rounded corners, inner xsep=7pt, inner ysep=3.5pt} % Adjusted inner xsep for more horizontal padding
}
\begin{figure}[h!]
	\centering
	\begin{tikzpicture}
		\node [mybox] (box){
			\begin{minipage}{0.9\columnwidth} % Restrict to one column width
				\footnotesize
				
%				\setstretch{1.3} % Line spacing
				
				\begin{flushleft}
					
					You are a specialist in the field of rare diseases.
					
					I will now give you ten predicted diseases if the predicted diagnosis is in the standard diagnosis. Please output the predicted rank, otherwise output "No", only output "No" or "1-10" numbers, if the predicted disease has multiple conditions, only output the top rank. Output only "No" or one number, no additional output.
					
					Predicted diseases: \texttt{\textbraceleft\textbraceleft predict\_diagnosis\_list\textbraceright\textbraceright}
					
					Standard diagnosis:\texttt{\textbraceleft\textbraceleft golden\_diagnosis\textbraceright\textbraceright}
					
				\end{flushleft}
			\end{minipage}
		};
		\node[fancytitle, rounded corners, right=10pt] at (box.north west) {Diagnosis Evaluation Prompt};
		
	\end{tikzpicture}
\end{figure}

\section{Details of Experiment Settings}

\textit{RareAgents} is built on the Llama-3.1 (8B and 70B)~\cite{dubey2024llama} and runs via vLLM (with parameters temperature=0 and seed=42)~\cite{kwon2023efficient} on a system with 4 NVIDIA A100 GPUs, each with 80 GB memory. In the differential diagnosis task, \textit{RareAgents} (Llama-3.1-70B) requires an average time of 36.31 seconds for multidisciplinary consultation per patient. In the medication recommendation task, the average time for multidisciplinary consultation per patient is 72.26 seconds.

\subsection{Task Prompt}

Below are the attending physician agent's system prompt and task-specific prompts. Full prompt configurations can be found in the code repository.

\begin{figure}[h!]
	\centering
	\begin{tikzpicture}
		\node [mybox] (box){
			\begin{minipage}{0.9\columnwidth} % Restrict to one column width
				\footnotesize
				
%				\setstretch{1.3} % Line spacing
				
				\begin{flushleft}
					
					You are a highly experienced physician. You will be provided with a complex clinical case that may involve atypical presentations or rare conditions. Carefully review the patient's symptoms, history, and any other relevant information.
					
				\end{flushleft}
			\end{minipage}
		};
		\node[fancytitle, rounded corners, right=10pt] at (box.north west) {Attending Physician Agent};
	\end{tikzpicture}
\end{figure}

\begin{figure}[h!]
	\centering
	\begin{tikzpicture}
		\node [mybox] (box){
			\begin{minipage}{0.9\columnwidth} % Restrict to one column width
				\footnotesize
				
%				\setstretch{1.3} % Line spacing
				
				\begin{flushleft}
					
					Based on the symptoms of the patient, list the diagnosis separately at the end in the following format:\\
					DIAGNOSIS:\\
					1. \textless Diagnosis 1\textgreater \\
					2. \textless Diagnosis 2\textgreater \\
					3. \textless Diagnosis 3\textgreater \\
					...\\
					10. \textless Diagnosis 10\textgreater 
					
				\end{flushleft}
			\end{minipage}
		};
		\node[fancytitle, rounded corners, right=10pt] at (box.north west) {Differential Diagnosis Prompt};
	\end{tikzpicture}
	\vspace{12pt}
\end{figure}

\begin{figure}[h!]
	\centering
	\begin{tikzpicture}
		\node [mybox] (box){
			\begin{minipage}{0.9\columnwidth} % Restrict to one column width
				\footnotesize
				
%				\setstretch{1.3} % Line spacing
				
				\begin{flushleft}
					
					Based on the diagnosis and procedures provided, please give the most appropriate combination of medications. Select medications only from the given list. List each medication on a separate line using the following format:\\
					TREATMENT:\\
					1. \textless Medication 1\textgreater \\
					2. \textless Medication 2\textgreater \\
					3. \textless Medication 3\textgreater \\
					...\\
					Only include medications from the provided list.
					
				\end{flushleft}
			\end{minipage}
		};
		\node[fancytitle, rounded corners, right=10pt] at (box.north west) {Medication Recommendation Prompt};
	\end{tikzpicture}
	\vspace{12pt}
\end{figure}

\begin{figure}[h]
	\centering
	\includegraphics[width=\columnwidth]{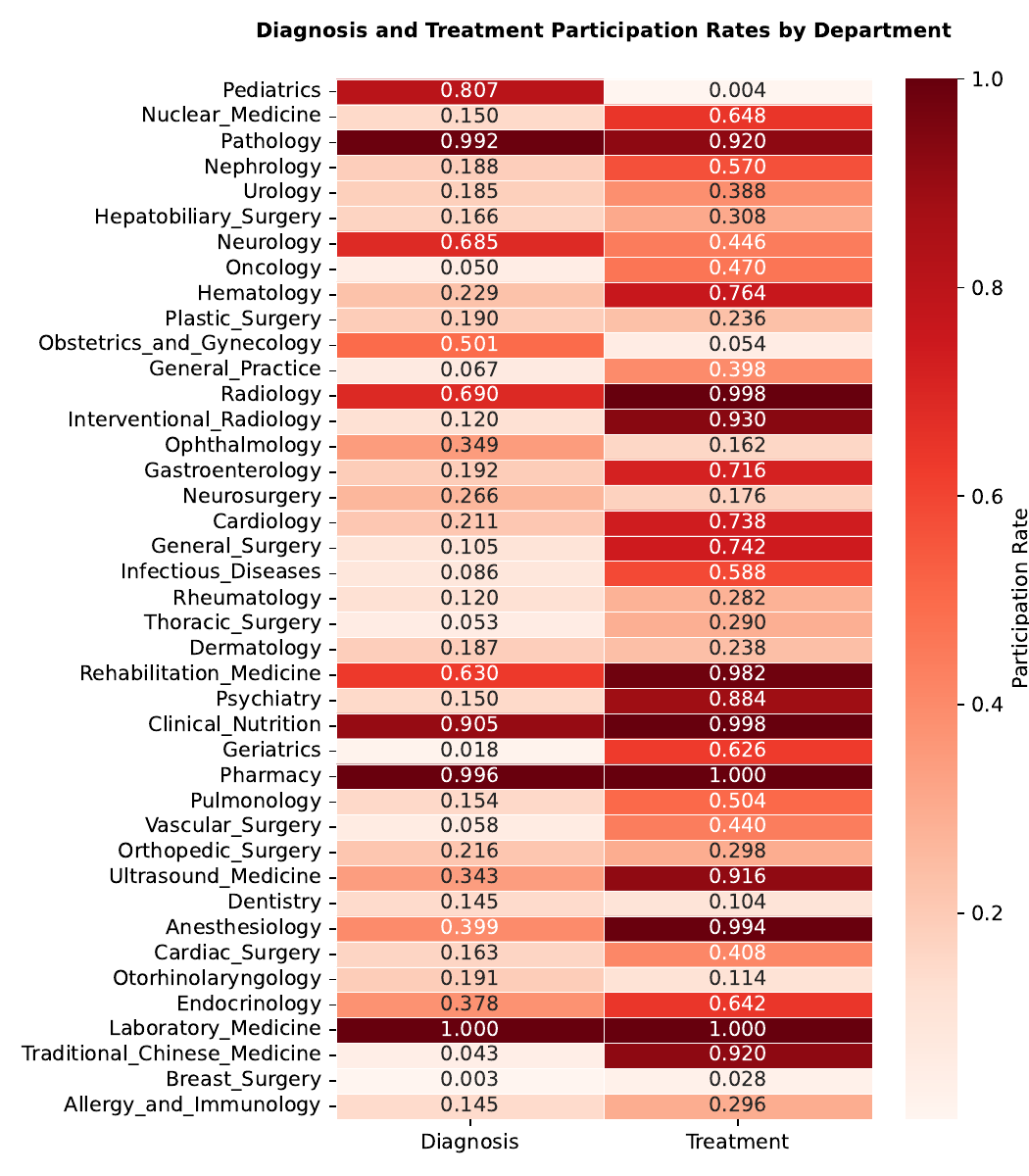}
	\caption{Heatmap of Participation Rates in Diagnostic and Treatment Tasks Across Specialist Departments.}
	\label{fig:sp_heatmap}
\end{figure}

\begin{figure}[h]
	\centering
	\includegraphics[width=\columnwidth]{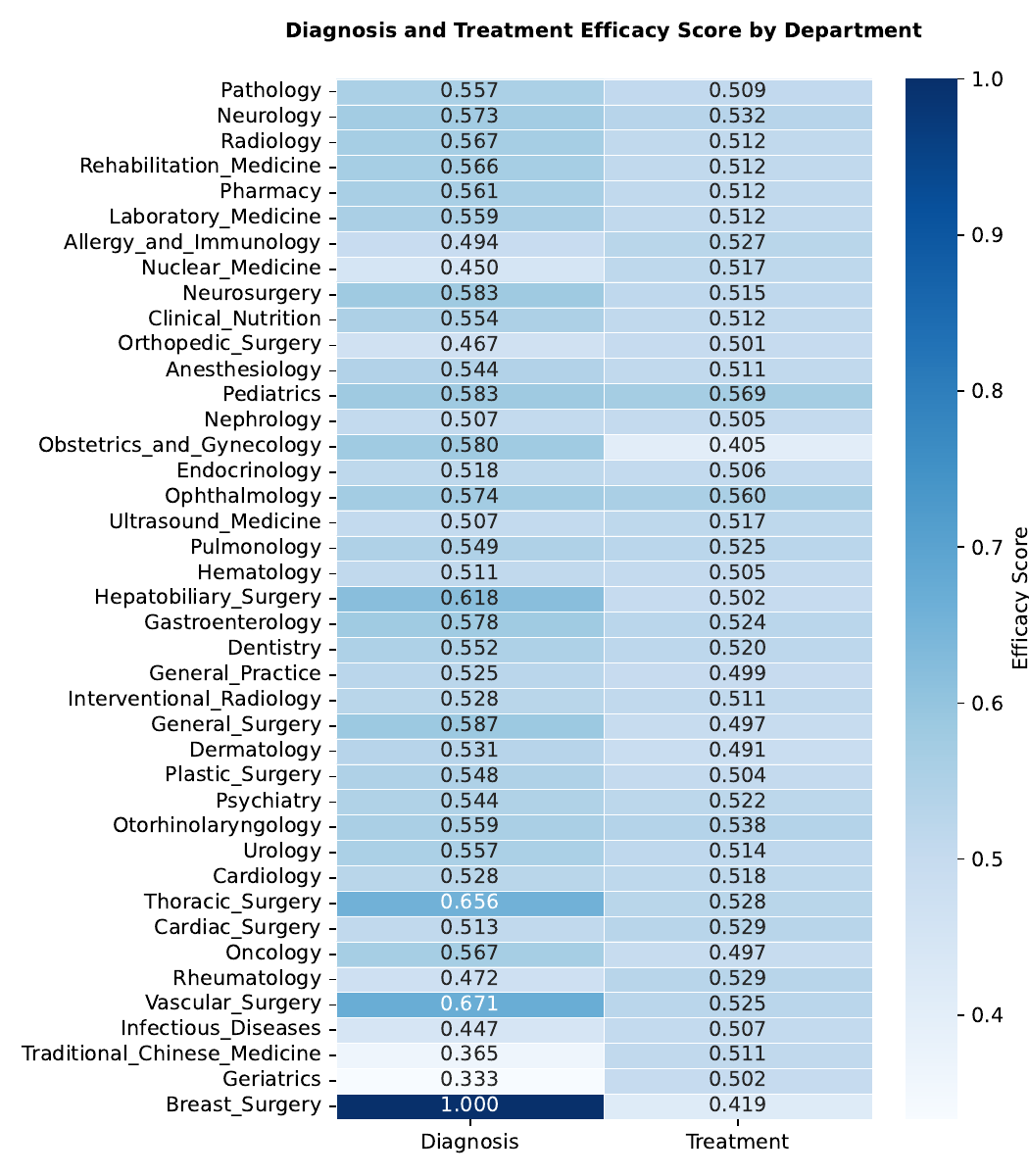}
	\caption{Heatmap of Efficacy Scores in Diagnostic and Treatment Tasks Across Specialist Departments.}
	\label{fig:new_sp_heatmap}
	\vspace{12pt}
\end{figure}

\subsection{Domain-Specific Baselines}
\label{sec:app_baseline}

Phen2Disease~\cite{zhai2023phen2disease} is a phenotype-driven model that ranks diseases and genes based on bidirectional maximum-matching semantic similarity calculations between patient and disease phenotype sets. In this study, we used the default settings of Phen2Disease, focusing exclusively on the disease ranking results.
BASE\_IC is a semantic similarity computation method based on Information Content (IC). It evaluates the informativeness of phenotypic terms by assessing their specificity and frequency, aiming to improve phenotype-disease matching and ranking. 
Descriptions of Phenomizer~\cite{kohler2009phenomizer1, kohler2017phenomizer2}, LIRICAL\cite{robinson2020lirical}, and Phenobrain~\cite{phenobrain} are provided in Appendix~\ref{sec:app_mt}.

Logistic Regression (LR) is an instance-based classifier with $L_2$ regularization, where inputs are represented as a multi-hot vector of length $|Diseases| + |Procedures|$.

LEAP (LEArn to Prescribe)~\cite{zhang2017leap} models medication recommendation as a sequential decision-making problem, using a recurrent decoder to capture drug dependencies. The hyperparameters in this study are set as follows: learning rate = 2e-4, number of epochs = 30. 

RETAIN~\cite{choi2016retain} employs a two-level neural attention mechanism to analyze electronic health record (EHR) data, identifying key historical visits and clinical variables that influence predictions. The hyperparameters are set as follows: learning rate = 2e-4, number of training epochs = 40.

G-Bert~\cite{shang2019pre} integrates hierarchical information from medical ontologies with EHR data for pretraining on medical code representation and drug recommendation tasks. The hyperparameters are set as follows: learning rate = 1e-4, dimension of node embeddings = 64, number of pretraining epochs = 100.

GAMENet~\cite{shang2019gamenet} combines a Graph Convolutional Network (GCN) and a Graph Augmented Memory Module to integrate a Drug-Drug Interaction (DDI) knowledge graph with longitudinal patient EHR data. The hyperparameters are set as follows: learning rate = 2e-4, dimension of node embeddings = 64, number of training epochs = 200.

SafeDrug~\cite{yang2021safedrug} encodes the molecular structure of drugs using a global Message Passing Neural Network (MPNN) and a local bipartite learning module. The hyperparameters are set as follows: learning rate = 5e-4, dimension of node embeddings = 64, number of training epochs = 200.

COGNet~\cite{wu2022conditional} is a conditional generative network for drug recommendation that dynamically determines whether to copy drugs from prior recommendations or predict new drugs based on the patient’s current diagnosis and historical records. The hyperparameters are set as follows: learning rate = 1e-3, max number of sentences in beam search = 4, dimension of node embeddings = 64, number of training epochs = 100.

MICRON~\cite{yang2021change} is a recurrent residual learning model for drug change prediction, capturing patient state transitions through efficient residual health representations. The hyperparameters are set as follows: learning rate = 2e-4, dimension of node embeddings = 64, number of training epochs = 40.

MoleRec~\cite{yang2023molerec} is a molecule substructure-aware model for drug combination recommendation. The hyperparameters are set as follows: learning rate = 5e-4, dropout ratio = 0.7, dimension of node embeddings = 64, number of training epochs = 50.

RAREMed~\cite{zhao2024leave} is a drug recommendation model designed for rare disease patients, employing a pretraining-finetuning paradigm. It incorporates self-supervised tasks (sequence alignment prediction and self-reconstruction) and a unified Transformer encoder for input sequences to capture complex relationships between diseases and medication codes. The hyperparameters are set as follows: learning rate = 1e-5, dropout ratio = 0.3, dimension of node embeddings = 512, number of pretraining epochs = 20.

\section{More Detailed Analysis}
\label{app:more_detailed_analysis}

\subsection{Specialists Collaboration Patterns}

To gain deeper insights into the collaboration and contribution distribution of specialist agents within \textit{RareAgents}, we conduct a detailed analysis of their interaction patterns. Figure~\ref{fig:sp_heatmap} displays the participation patterns of specialist agents within \textit{RareAgents} (Llama-3.1-70B). In the diagnostic task, the relatively even distribution of contributions across specialties underscores the necessity of diverse clinical insights for accurately identifying rare diseases. In contrast, the treatment task sees more concentrated input from select experts, reflecting the need for deep, specialized knowledge to navigate complex therapeutic protocols and ensure optimal patient outcomes. Furthermore, the structured role assignments facilitate better collaboration among agents, allowing them to leverage complementary expertise and provide a more holistic clinical assessment., Furthermore, we analyzed the efficacy scores of each specialist agent, measured by the average diagnostic Recall@1 and medication recommendation F1 across assigned cases. As shown in Figure~\ref{fig:new_sp_heatmap}, over 90\% of specialists achieved efficacy scores above 0.5 in both tasks, confirming their necessity and effectiveness.

\subsection{More Ablation Study about Memory Mechanism}

Existing medical multi-agent frameworks, such as MedAgents~\cite{tang-etal-2024-medagents} and MDAgents~\cite{kimmdagents}, do not incorporate a memory mechanism. To assess the impact of this feature, we performed an ablation study comparing conditions with and without the memory module under the Single-Agent baseline. The results, presented in the Table~\ref{tab:app_memory_module}, indicate that the inclusion of a memory mechanism yields a noticeable performance improvement. Specifically, while the memory-enhanced agent outperforms the baseline, RareAgents achieves the highest performance, underscoring the combined benefits of memory and multi-agent collaboration. This suggests that the memory mechanism not only aids in retaining contextual information but also complements the synergy among agents, which is crucial in rare disease diagnosis and treatment.

\subsection{Random vs. Dynamic of Memory Setting}

\begin{figure*}[ht]
		\centering
		\begin{subfigure}{0.32\textwidth}
			\centering
			\includegraphics[width=\linewidth]{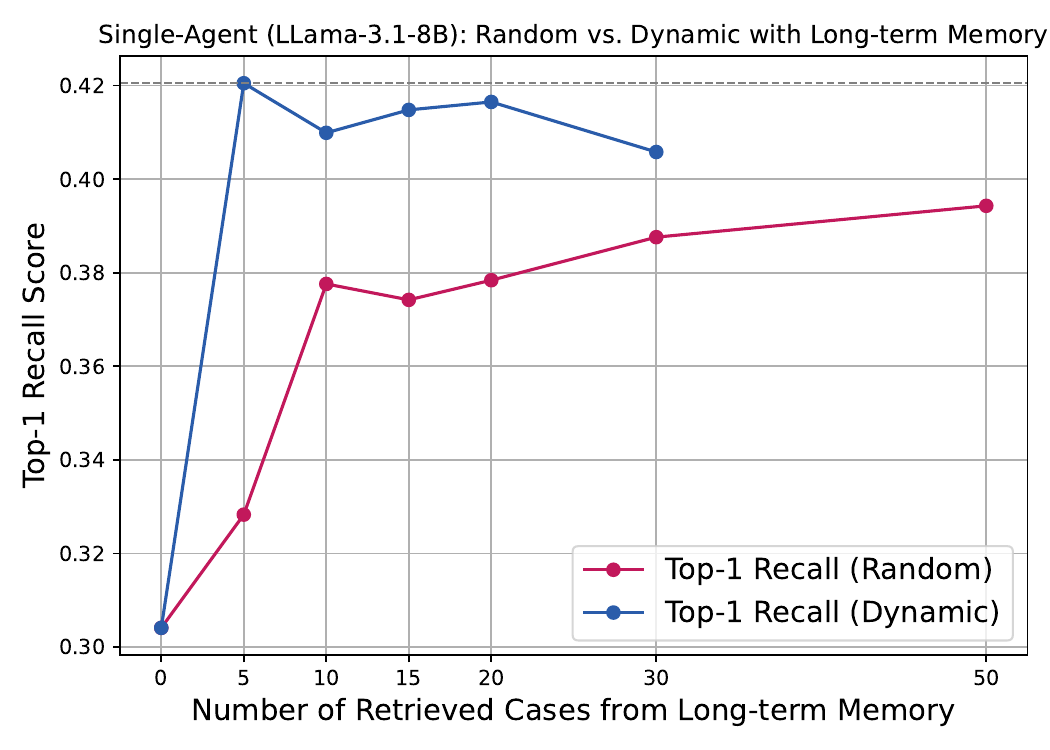}
			\caption{Top-1 Recall by memory settings.}
			\label{fig:diagnosis_memory_top1}
		\end{subfigure}%
		\hfill
		\begin{subfigure}{0.32\textwidth}
			\centering
			\includegraphics[width=\linewidth]{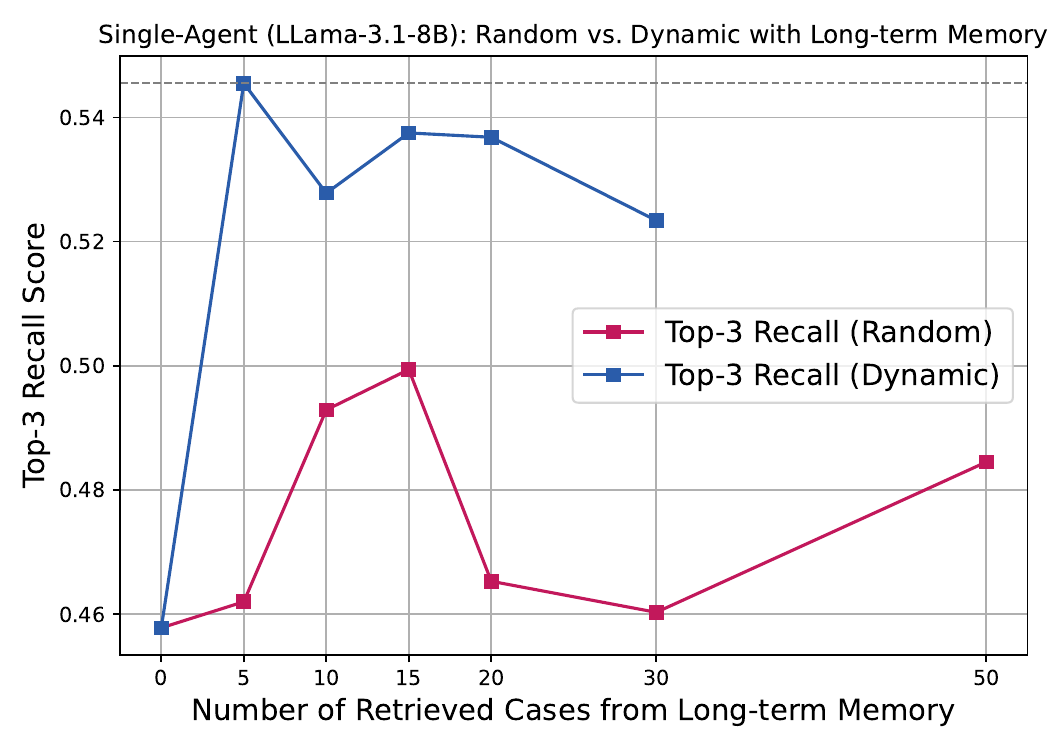}
			\caption{Top-3 Recall by memory settings.}
			\label{fig:diagnosis_memory_top3}
		\end{subfigure}%
		\hfill
		\begin{subfigure}{0.32\textwidth}
			\centering
			\includegraphics[width=\linewidth]{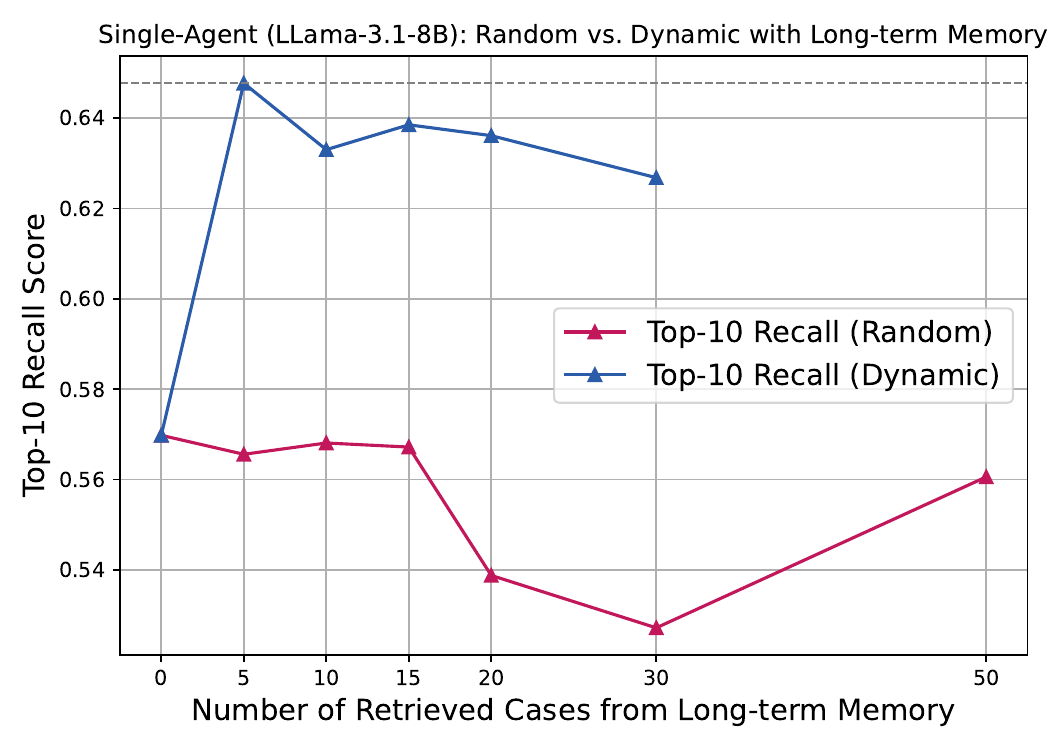}
			\caption{Top-10 Recall by memory settings.}
			\label{fig:diagnosis_memory_top10}
		\end{subfigure}%
		\hfill
		\begin{subfigure}{0.32\textwidth}
			\centering
			\includegraphics[width=\linewidth]{assets/treatment_memory_1.pdf}
			\caption{Jaccard Score by memory settings.}
			\label{fig:treatment_memory_1}
		\end{subfigure}%
		\hfill
		\begin{subfigure}{0.32\textwidth}
			\centering
			\includegraphics[width=\linewidth]{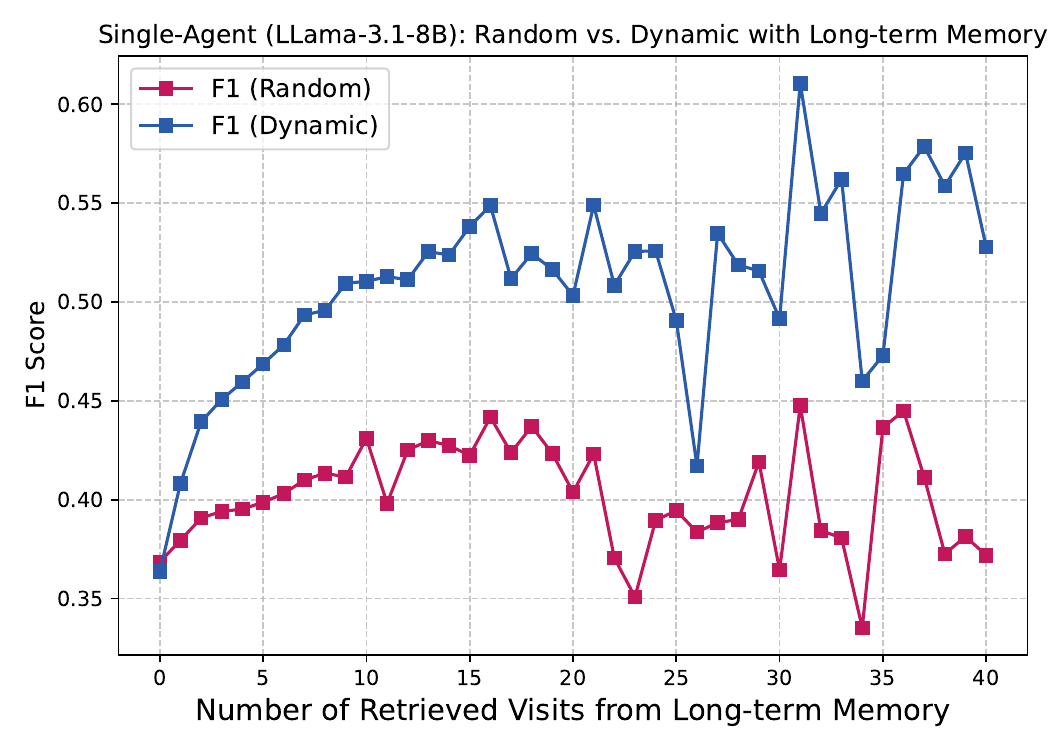}
			\caption{F1 Score by memory settings.}
			\label{fig:treatment_memory_2}
		\end{subfigure}
		\hfill
		\begin{subfigure}{0.32\textwidth}
			\centering
			\includegraphics[width=\linewidth]{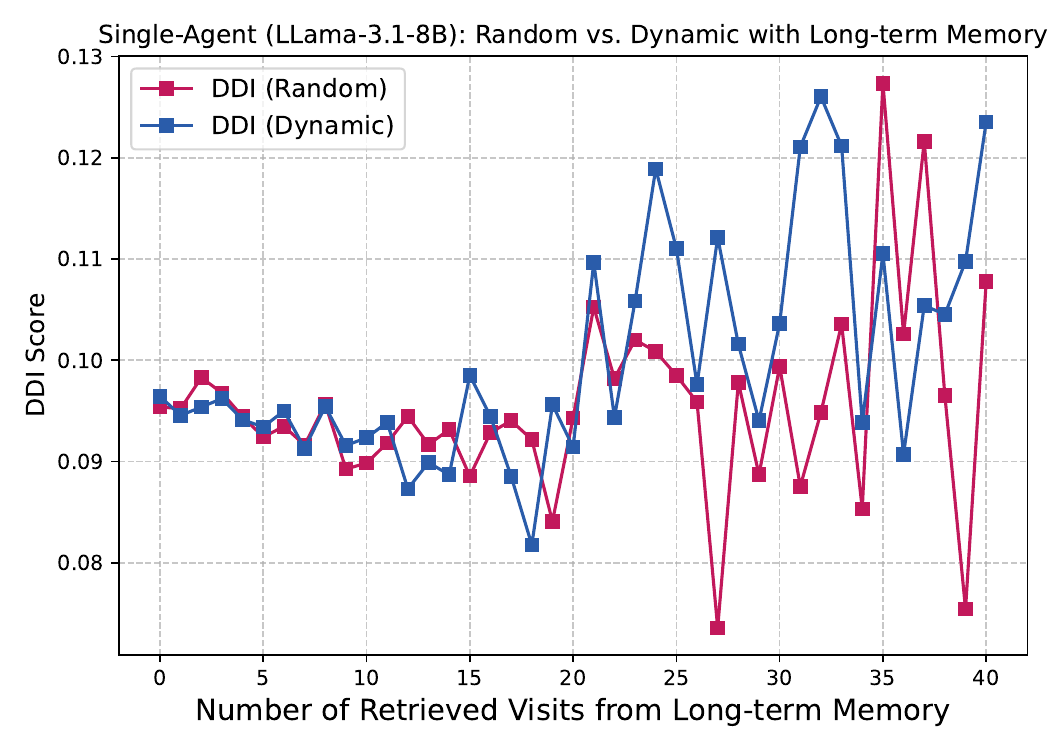}
			\caption{DDI Score by memory settings.}
			\label{fig:treatment_memory_3}
		\end{subfigure}
		\vspace{12pt}
		\caption{A detailed exploration of each module in \textit{RareAgents}: (i) MDT roles, (ii) Memory, (iii) Tools.}
		\label{fig:app_memory}
	\end{figure*}

To evaluate the efficacy of the dynamic retrieval mechanism in long-term memory, we compare it with a baseline approach that randomly selects an equivalent number of cases. The results are shown in Figure~\ref{fig:app_memory}, the dynamic memory mechanism achieves significant performance improvements by retrieving only a small number of highly relevant records (e.g., just 5 cases), emphasizing that precise context outweighs sheer volume. In contrast, random retrieval offers limited utility even when retrieving a vast number of cases.

Further analysis of the dynamic long-term memory for diagnosis reveals that the optimal diagnostic performance occurs at $k=5$. As the number of cases increases, the diagnostic performance may decline. This is because cases with similar symptoms are often retrieved within the first five instances, and incorporating more cases may introduce additional hallucinations in the LLM, reducing diagnostic reliability.

\subsection{Effectiveness of Different Medical Tools}

\begin{figure*}[ht]
		\centering
		\begin{subfigure}{0.42\textwidth}
			\centering
			\includegraphics[width=\linewidth]{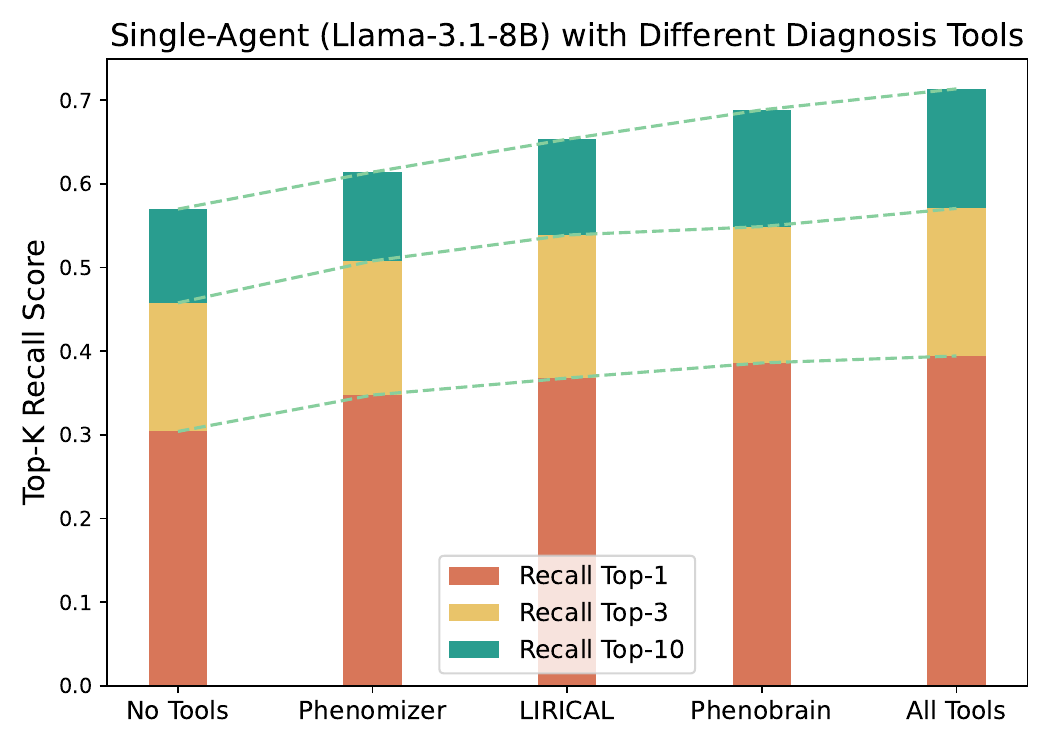}
			\caption{Treatment Performance under five Tools settings.}
			\label{fig:diagnosis_tools}
		\end{subfigure}
		\hfill
		\begin{subfigure}{0.42\textwidth}
			\centering
			\includegraphics[width=\linewidth]{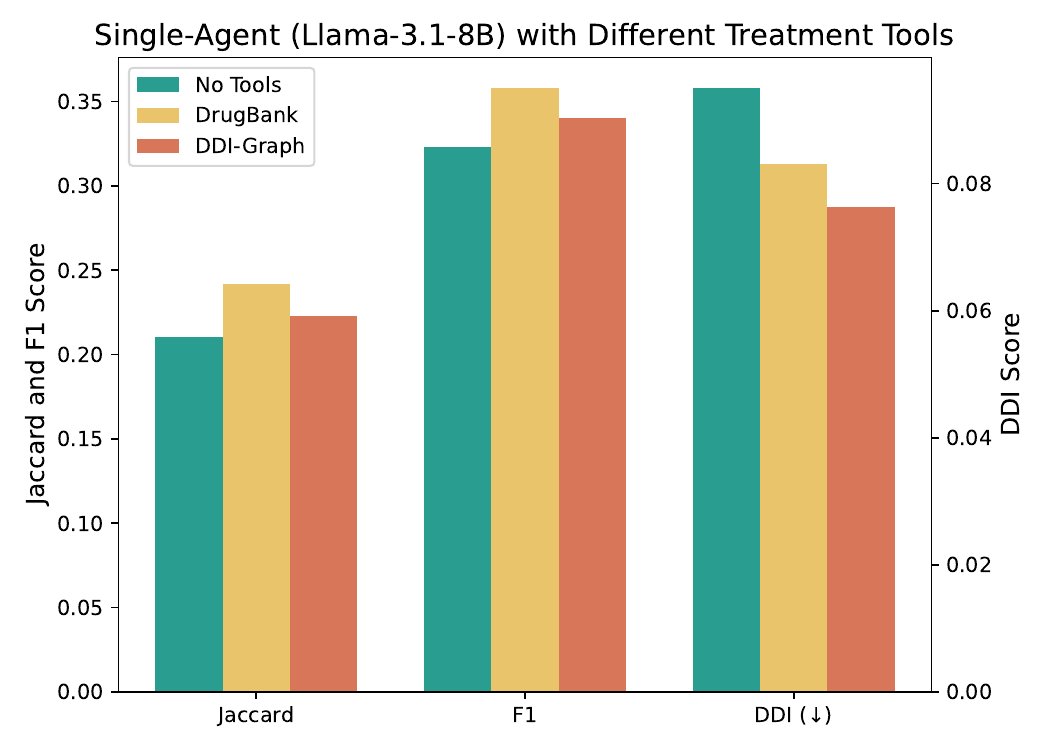}
			\caption{Treatment Performance under three Tools settings.}
			\label{fig:treatment_tools}
		\end{subfigure}
		\vspace{12pt}
		\caption{Detailed Analysis of the Tool Module in \textit{RareAgents}: Utilization of Single Tool Enhances Performance.}
		\label{fig:app_tools}
	\end{figure*}

We further assess the individual contributions of tools in a single-agent setup, where the agent is restricted to using one tool at a time. The results, shown in Figures~\ref{fig:app_tools}, indicate that each tool independently enhances the agent's performance, with more effective tools contributing greater enhancements. Their combined use further reinforces overall results by addressing different aspects of the task, highlighting the complementary nature of these resources. Combining all tools delivers the best overall results. Moreover, incorporating the DDI-graph significantly reduces the DDI rate in treatment, promoting medication safety and efficacy.

\begin{table*}[ht!]
	\centering
	\small

	% \resizebox{2.0\columnwidth}{!}{
		\begin{tabular}{clccc|ccll}
			\toprule
			\multirow{2}{*}{\textbf{Model}} & \multicolumn{4}{c|}{\textbf{Diagnosis} on \textbf{\textsc{RareBench-Public}}} & \multicolumn{4}{c}{\textbf{Treatment} on \textbf{\textsc{MIMIC-IV-Ext-Rare}}} \\
			\cmidrule(lr){2-5} \cmidrule(lr){6-9}
			& {\textbf{Hit@1}} & {\textbf{Hit@3}} & {\textbf{Hit@10}} & {\textbf{MR}($\downarrow$)} & {\textbf{Jaccard}} & {\textbf{F1}} & {\textbf{DDI}($\downarrow$)} & {\textbf{\#MED}} \\
			\midrule
%			\rowcolor{gray!20}
			\multicolumn{9}{l}{\textit{\textbf{Llama-3.1-8B-Instruct}}} \\
			\textbf{Single Agent (w/o Memory)} &    0.3041       &   0.4578         &     0.5698       &   5.0         &    0.2104        &    0.3229        &    0.0951      &   9.68        \\
			\textbf{Single Agent (w/ Memory)}     &   0.4002 & 0.5063 & 0.6558 & 3.0 & 0.2458 & 0.3813 & 0.0928 & 11.73      \\
			\cellcolor{lightblue}\textbf{RareAgents}      &  \textbf{0.4511}         &   \textbf{0.5647}         &    \textbf{0.7377}        &     \textbf{2.0}       &     \textbf{0.3052}       &    \textbf{0.4475}        &    \textbf{0.0820}        &       12.98      \\
			\midrule
%			\rowcolor{gray!20}
			\multicolumn{9}{l}{\textit{\textbf{Llama-3.1-70B-Instruct}}} \\
			\textbf{Single Agent (w/o Memory)} &    0.3751 & 0.5397 & 0.6658 & 3.0 & 0.2543 & 0.3736 & 0.0907 & 10.97        \\
			\textbf{Single Agent (w/ Memory)}    &   0.4887 & 0.6057 & 0.6934 & 2.0 & 0.3345 & 0.4731 & 0.0891 & 13.27       \\
			\cellcolor{lightblue}\textbf{RareAgents}      &   \textbf{0.5589}         &   \textbf{0.6867}         &   \textbf{0.7811}         &         \textbf{1.0}    &    \textbf{0.4108}         &  \textbf{0.5563}          &    \textbf{0.0796}        &    13.17            \\
			\bottomrule
		\end{tabular}
		% }
		\caption{More Ablation Study about Memory Mechanism}
	\label{tab:app_memory_module}
	\end {table*}
	
	\begin{figure*}[ht]
		\centering
		\includegraphics[width=2.0\columnwidth]{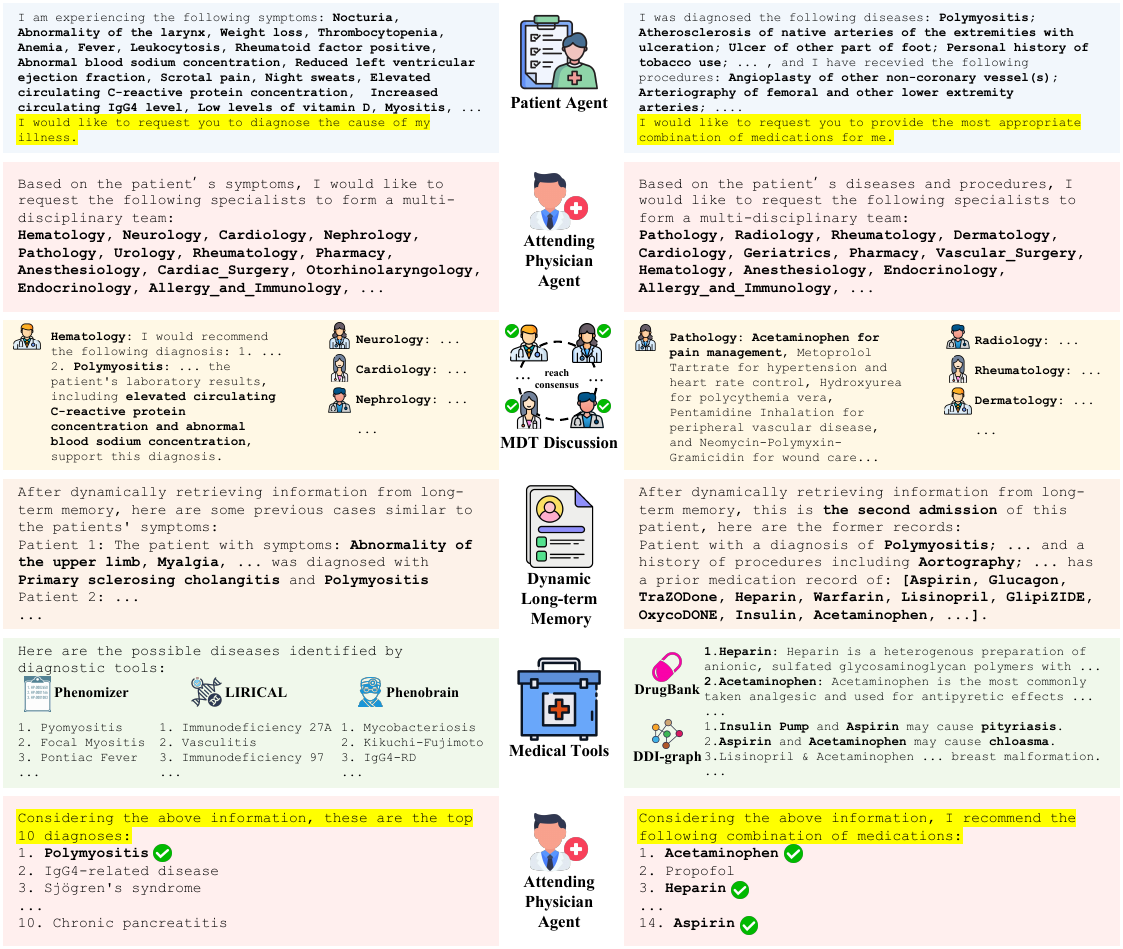}
		\caption{Illustrative examples of our proposed \textbf{RareAgents} framework for diagnosing and treating rare diseases are presented. The left panel features a patient from RareBench-Public requiring differential diagnosis based on symptoms, while the right panel shows a case from MIMIC-IV-Ext-Rare involving medication recommendations based on diseases and procedures. \textit{RareAgents} integrates multidisciplinary team (MDT) collaboration, dynamic long-term memory, and the utilization of medical tools to provide patient-centered, personalized care.}
		\label{fig:example}
	\end{figure*}

\tikzset{
	mybox/.style={
		draw=black, 
		fill=yellow!20, % 浅黄色背景
		very thick, 
		rectangle, 
		rounded corners, 
		inner sep=10pt, 
		drop shadow % 添加阴影效果
	},
	fancytitle/.style={
		fill=black!55, 
		text=white, 
		font=\bfseries, % 加粗字体
		rounded corners, 
		inner xsep=10pt, 
		inner ysep=3pt
	}
}

\begin{figure*}[h!]
	\centering
	\begin{tikzpicture}
		\node [mybox] (box){
			\begin{minipage}{1.9\columnwidth} % Restrict to one column width
				\footnotesize
				
%				\setstretch{1.3} % Line spacing
				
				\begin{flushleft}
					\includegraphics[width=0.4cm]{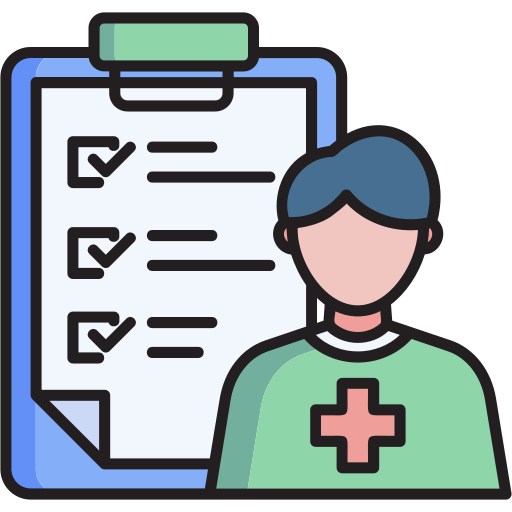} \ \textbf{Patient Agent:}
					I am experiencing the following symptoms: Urinary hesitancy, Urinary incontinence, Prostatitis, Syncope, Muscle weakness, Elevated circulating creatine kinase concentration, Exercise intolerance, Loss of consciousness, Pedal edema, ST segment elevation, Increased circulating creatine kinase MB isoform, Dysuria, Amaurosis fugax. I would like to request you to diagnose the cause of my illness.
					
					\hdashrule[0.5ex]{146mm}{1pt}{2mm}
					
					\includegraphics[width=0.45cm]{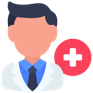} \ \textbf{Single-Agent (Llama-3.1-70B): }
					Based on the symptoms provided, here's a detailed analysis and the top 10 most likely diagnoses:\\
					The patient's symptoms can be broadly categorized into several groups: urinary symptoms (urinary hesitancy, urinary incontinence, prostatitis, dysuria), cardiovascular symptoms (syncope, loss of consciousness, ST segment elevation, increased circulating creatine kinase MB isoform), muscular symptoms (muscle weakness, elevated circulating creatine kinase concentration, exercise intolerance), and neurological symptoms (amaurosis fugax). Pedal edema could be related to either cardiovascular or renal issues.\\
					Given the complexity and the combination of these symptoms, it's essential to consider conditions that could potentially explain multiple symptom groups simultaneously. Here's a step-by-step breakdown:\\
					1. **Cardiovascular Symptoms**: The presence of ST segment elevation and increased circulating creatine kinase MB isoform strongly suggests myocardial infarction (heart attack) or acute coronary syndrome. However, the combination with other symptoms like muscle weakness and elevated creatine kinase concentration points towards conditions affecting both the heart and skeletal muscles.\\
					2. **Muscular Symptoms**: Elevated creatine kinase levels and muscle weakness could indicate a myopathy. When combined with cardiovascular symptoms, conditions like myocarditis or cardiomyopathy with skeletal muscle involvement become plausible.\\
					3. **Urinary Symptoms**: These could be indicative of a urinary tract infection, benign prostatic hyperplasia (BPH), or other conditions affecting the prostate or urinary system. However, in the context of the other symptoms, a systemic condition affecting multiple body systems is more likely.\\
					4. **Neurological Symptoms**: Amaurosis fugax (temporary loss of vision in one eye) could be indicative of a transient ischemic attack (TIA) or a vascular issue affecting the eye.\\
					Considering the combination of these symptoms, here are the top 10 most likely diagnoses:\\
					DIAGNOSIS:\\
					\includegraphics[width=0.26cm]{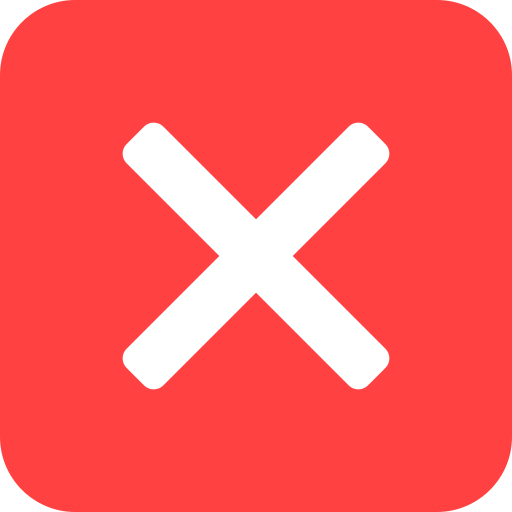} \ 1. **Myocardial Infarction with Cardiac Sarcoidosis**: A condition where granulomatous inflammation affects the heart, potentially leading to heart failure, arrhythmias, and symptoms overlapping with those described.\\
					\includegraphics[width=0.26cm]{assets/wrong.png} \ 2. **Duchenne Muscular Dystrophy with Cardiac Involvement**: Although typically diagnosed in childhood, some cases may present later, and cardiac involvement is common, which could explain the combination of muscle weakness, elevated CK, and cardiac symptoms.\\
					\includegraphics[width=0.26cm]{assets/wrong.png} \ 3. **Myocarditis with Skeletal Muscle Involvement**: Inflammation of the heart muscle that could also affect skeletal muscles, explaining the elevated CK levels and muscle weakness.\\
					\includegraphics[width=0.26cm]{assets/wrong.png} \ 4. **Cardiac Amyloidosis**: A condition where abnormal proteins deposit in the heart, potentially leading to heart failure and symptoms that could overlap with those described, including neuropathy and muscle weakness.\\
					\includegraphics[width=0.26cm]{assets/wrong.png} \ 5. **Ehlers-Danlos Syndrome with Cardiac Involvement**: A genetic disorder affecting connective tissue, which could explain the diverse range of symptoms including cardiac issues, muscle weakness, and potentially the urinary symptoms.\\
					\includegraphics[width=0.26cm]{assets/wrong.png} \ 6. **Mitochondrial Myopathy, Encephalopathy, Lactic Acidosis, and Stroke-like Episodes (MELAS) Syndrome**: A mitochondrial disorder that affects many areas of the body, including the heart, muscles, and nervous system.\\
					\includegraphics[width=0.26cm]{assets/wrong.png} \ 7. **Fabry Disease**: A genetic disorder that results from the buildup of a particular type of fat in the body's cells, affecting many areas including the heart, kidneys, and nervous system.\\
					\includegraphics[width=0.26cm]{assets/wrong.png} \ 8. **Acute Coronary Syndrome with Rhabdomyolysis**: A condition where a heart attack is complicated by muscle breakdown, leading to elevated CK levels and muscle weakness.\\
					\includegraphics[width=0.26cm]{assets/wrong.png} \ 9. **Cardiac Sarcoidosis with Neurosarcoidosis**: Sarcoidosis affecting both the heart and nervous system, potentially explaining the amaurosis fugax and other neurological symptoms.\\
					\includegraphics[width=0.26cm]{assets/wrong.png} \ 10. **Danon Disease**: A rare genetic disorder characterized by cardiomyopathy, skeletal myopathy, and other systemic symptoms.\\
					Each of these conditions is rare and complex, and a definitive diagnosis would require extensive diagnostic testing, including genetic analysis, cardiac evaluation, muscle biopsy, and potentially other specialized tests.

				\end{flushleft}
			\end{minipage}
		};
		\node[fancytitle, rounded corners, right=10pt] at (box.north west) {Case 1};
	\end{tikzpicture}
\end{figure*}

\begin{figure*}[ht]
	\centering
	\begin{tikzpicture}
		\node [mybox] (box){
			\begin{minipage}{1.9\columnwidth} % Restrict to one column width
				\footnotesize
				
%				\setstretch{1.3} % Line spacing
				
				\begin{flushleft}
					\textbf{RareAgents (Llama-3.1-70B): }
					
					\includegraphics[width=0.45cm]{assets/apa.png} \ \textbf{Attending Physician Agent: } Based on the patient’s symptoms, I would like to request the following specialists to form a multi-disciplinary team: 
					Urology, Cardiology, Neurology, Pathology, Nuclear Medicine, Radiology, Interventional Radiology, Ophthalmology, Rehabilitation Medicine, Pharmacy, Vascular Surgery, Ultrasound Medicine, Anesthesiology, Cardiac Surgery, Laboratory Medicine.\\
					\texttt{\textbraceleft\textbraceleft Multiple rounds of discussions by multidisciplinary team (MDT)\textbraceright\textbraceright}\\
					...\\ 
					\includegraphics[width=0.35cm]{assets/doctor1.png} \ \textbf{Cardiology Physician Agent: } Based on the patient's symptoms, medical history, and test results, I would recommend the following diagnosis and treatment plan: 1. **Acute Myocardial Infarction (AMI) with Rhabdomyolysis**: The patient's symptoms of chest pain, shortness of breath, and elevated troponin levels are consistent with an acute myocardial infarction. The presence of rhabdomyolysis, as indicated by elevated creatine kinase levels, suggests muscle damage. 2. **Myasthenia Gravis with Cardiac Involvement**: The patient's symptoms of muscle weakness, fatigue, and difficulty swallowing, along with the presence of anti-acetylcholine receptor antibodies, suggest myasthenia gravis. The cardiac involvement is indicated by the patient's symptoms of palpitations and shortness of breath....\\
					\includegraphics[width=0.35cm]{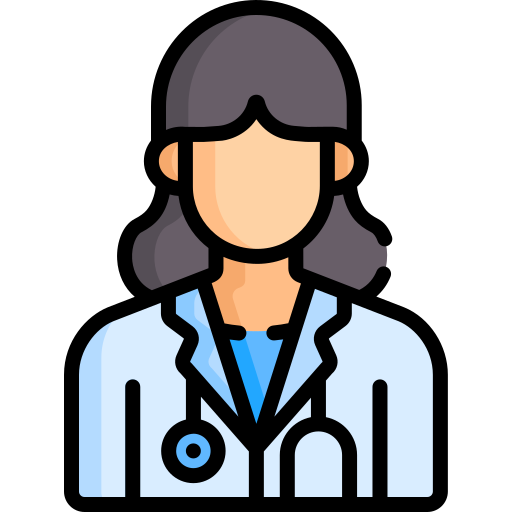} \ \textbf{Neurology Physician Agent: } Based on the diagnostic tools' results, the most likely diagnosis is Brugada Syndrome 4, which is a genetic disorder that affects the heart's electrical system and can cause abnormal heart rhythms, including ventricular fibrillation. The patient's symptoms, such as syncope, muscle weakness, and elevated circulating creatine kinase concentration, are consistent with this diagnosis. However, it's important to note that the patient's symptoms also suggest other possible diagnoses, such as multiple sclerosis, muscular dystrophy, and myasthenia gravis, which should be considered in the differential diagnosis. Therefore, a comprehensive diagnostic workup, including genetic testing, electromyography, and imaging studies, is necessary to confirm the diagnosis and rule out other possible conditions.\\
					\includegraphics[width=0.35cm]{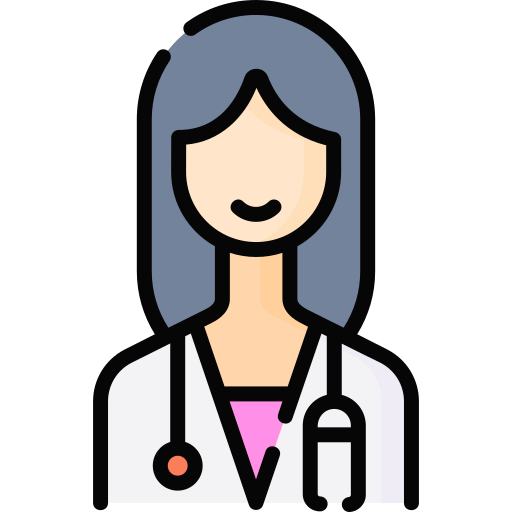} \ \textbf{Pathology Physician Agent: } Based on the patient's symptoms, I would recommend the following diagnosis:\\
					1. **Acute Myocardial Infarction (AMI) with Rhabdomyolysis**: The patient's symptoms of ST segment elevation, elevated circulating creatine kinase concentration, and increased circulating creatine kinase MB isoform suggest an acute myocardial infarction. The presence of muscle weakness, exercise intolerance, and elevated creatine kinase levels also suggest rhabdomyolysis. 2. **Myasthenia Gravis with Cardiac Involvement**: The patient's symptoms of muscle weakness, exercise intolerance, and syncope could be indicative of myasthenia gravis, an autoimmune disorder that affects the neuromuscular junction. The presence of cardiac involvement is suggested by the ST segment elevation and elevated creatine kinase levels....\\
					\includegraphics[width=0.35cm]{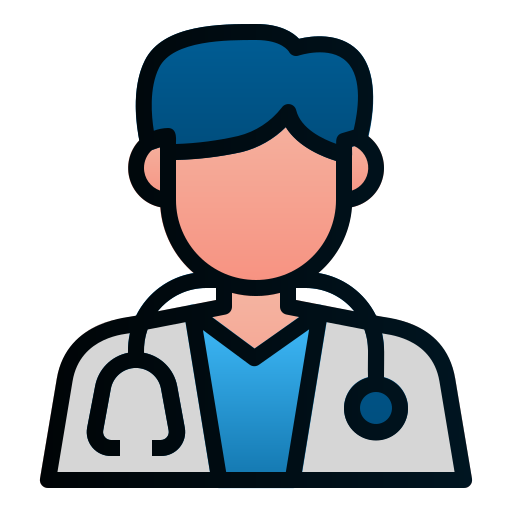} \ \textbf{Urology Physician Agent: } Based on the diagnostic tools' results, the most likely diagnosis is Brugada Syndrome 4, which is a genetic disorder that affects the heart's electrical system and can cause abnormal heart rhythms, including ventricular fibrillation. The patient's symptoms, such as syncope, muscle weakness, and ST segment elevation, are consistent with this diagnosis. However, it is essential to note that the diagnostic tools' results are not definitive, and a comprehensive evaluation by a healthcare professional is necessary to confirm the diagnosis. Additional testing, such as genetic testing, imaging studies, and electrophysiological studies, may be necessary to determine the underlying cause of the patient's symptoms. In conclusion, based on the diagnostic tools' results, Brugada Syndrome 4 is the most likely diagnosis, but further evaluation and testing are necessary to confirm the diagnosis and rule out other potential causes of the patient's symptoms.\\
					...\\

				\end{flushleft}
			\end{minipage}
		};
		\node[fancytitle, rounded corners, right=10pt] at (box.north west) {Case 1};
	\end{tikzpicture}
\end{figure*}

\begin{figure*}[ht]
	\centering
	\begin{tikzpicture}
		\node [mybox] (box){
			\begin{minipage}{1.9\columnwidth} % Restrict to one column width
				\footnotesize
				
%				\setstretch{1.3} % Line spacing
				
				\begin{flushleft}
					\textbf{RareAgents (Llama-3.1-70B): }
					
					\texttt{\textbraceleft\textbraceleft Dynamic retrieval of patients with similar symptoms from long-term memory\textbraceright\textbraceright}\\
					\includegraphics[width=0.4cm]{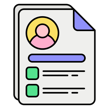} \ \textbf{Memory Retrieve Results: } Consider these previous cases for reference: \\
					1. Patient with symptoms: Trismus, Hyperhidrosis, Syncope, Vomiting, Loss of consciousness, Right bundle branch block, ST segment elevation was diagnosed with Brugada syndrome.\\
					2. Patient with symptoms: Urinary urgency, Urinary hesitancy, Urinary incontinence, Syncope, Stroke, Slurred speech, Constipation, Cerebral atrophy, Dysdiadochokinesis, Abnormal spinal cord morphology, Positive Romberg sign, Abnormal cerebral white matter morphology, Increased intracranial pressure, Cerebral ischemia, Increased CSF protein concentration, EMG: neuropathic changes, Babinski sign, Difficulty climbing stairs, Loss of consciousness, Lower limb muscle weakness, Abnormal prostate morphology, Sleep apnea, CSF pleocytosis, Dysesthesia, Abasia, Abnormal male reproductive system physiology, Schwannoma, Pollakisuria, Dysuria, Neoplasm of the lung, Erectile dysfunction, Abnormal lumbar spine morphology was diagnosed with Multiple system atrophy.\\
					3. Patient with symptoms: Syncope, Palpitations, ST segment elevation was diagnosed with Brugada Syndrome.\\ 
					4. Patient with symptoms: Syncope, Cardiomyopathy, Ventricular fibrillation, Cardiac arrest, Right bundle branch block, Myocarditis was diagnosed with Cateeholaminergic polymorphic ventricular tachycardia.\\
					5. Patient with symptoms: Ventricular fibrillation, Pneumonia, Dyspnea, Exercise intolerance, Viral hepatitis, ST segment elevation, Cough was diagnosed with Brugada syndrome.\\
					\texttt{\textbraceleft\textbraceleft Feedback after the function calls the diagnostic tool\textbraceright\textbraceright}\\
					
					\includegraphics[width=0.35cm]{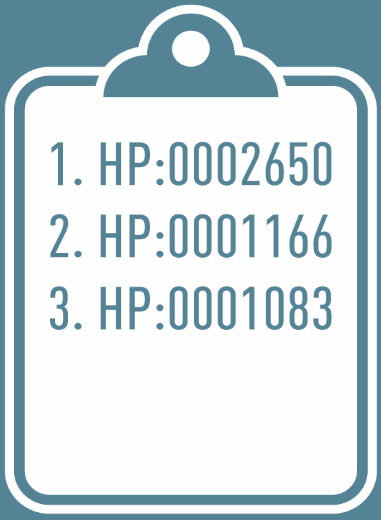} \ \textbf{Phenomizer: } 1. 109820 BLADDER DIVERTICULUM, p-value: 0.0160. 2. MULTIPLE SCLEROSIS, SUSCEPTIBILITY TO, p-value: 0.0374. 3. MUSCLE GLYCOGENOSIS, X-LINKED, p-value: 0.0374. 4.  \#125310 CEREBRAL ARTERIOPATHY, AUTOSOMAL DOMINANT, WITH SUBCORTICAL INFARCTSAND LEUKOENCEPHALOPATHY; CADASIL;DEMENTIA, HEREDITARY MULTI-INFARCT TYPE;CASIL, p-value: 0.0881. 5.  \#616231 MYOPATHY, VACUOLAR, WITH CASQ1 AGGREGATES; VMCQA, p-value: 0.0881. 6.  \#616094 MUSCULAR DYSTROPHY-DYSTROGLYCANOPATHY (LIMB-GIRDLE), TYPE C, 12; MDDGC12;MUSCULAR DYSTROPHY-DYSTROGLYCANOPATHY, LIMB-GIRDLE, POMK-RELATED, p-value: 0.0881. 7.  \#611876 BRUGADA SYNDROME 4; BRGDA4, p-value: 0.1207. 8.  \#609620 SHORT QT SYNDROME 1; SQT1, p-value: 0.1207. 9.  \#612347 JERVELL AND LANGE-NIELSEN SYNDROME 2; JLNS2, p-value: 0.1207. 10. MUSCULAR DYSTROPHY, CARDIAC TYPE, p-value: 0.1207\\
					
					\includegraphics[width=0.35cm]{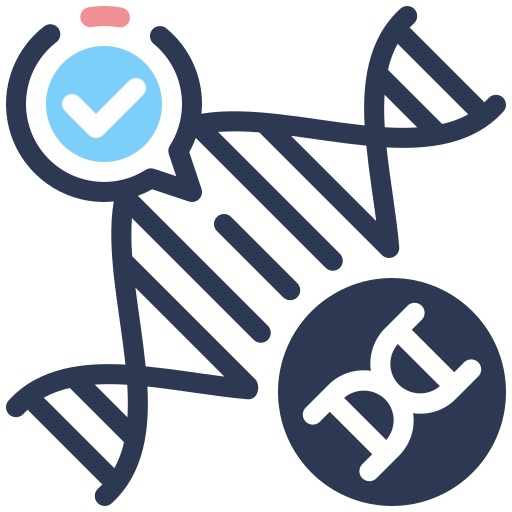} \ \textbf{LIRCAL: } 1. Glycogen storage disease II, posterior probability: 72.30 \%. 2. Danon disease, posterior probability: 0.00 \%. 3. Polyglucosan body myopathy 1 with or without immunodeficiency, posterior probability: 0.00 \%. 4. Myopathy, myofibrillar, 7, posterior probability: 0.00 \%. 5. Muscular dystrophy, limb-girdle, autosomal recessive 25, posterior probability: 0.00 \%. 6. Neuronal intranuclear inclusion disease, posterior probability: 0.00 \%. 7. Progressive external ophthalmoplegia with mitochondrial DNA deletions, autosomal recessive 5, posterior probability: 0.00 \%. 8. Progressive external ophthalmoplegia with mitochondrial DNA deletions, autosomal dominant 4, posterior probability: 0.00 \%. 9. Lipodystrophy, congenital generalized, type 4, posterior probability: 0.00 \%. 10. Mitochondrial DNA depletion syndrome 11, posterior probability: 0.00 \%.\\
					
					\includegraphics[width=0.35cm]{assets/lirical.png} \ \textbf{Phenobrain: } 1. Cerebralautosomal dominant angiopathy with subcortical infarcts and leukoencephalopathy, score: 0.9998560115190785. 2. Malakoplakia, score: 0.9992800575953923. 3. Medium chain acyl-CoA dehydrogenase deficiency, score: 0.9992080633549316. 4. Eisenmenger syndrome, score: 0.9991360691144708. 5. Pure autonomic failure, score: 0.99906407487401. 6. Porphyria/Porphyria, score: 0.9987041036717063. 7. Adrenoleukodystrophy, score: 0.9985601151907847. 8. Multiple sclerosis, susceptibility to, score: 0.9983441324694025. 9. Acute intermittent porphyria/Porphyria, acute intermittent, score: 0.9983081353491721.\\

				\end{flushleft}
			\end{minipage}
		};
		\node[fancytitle, rounded corners, right=10pt] at (box.north west) {Case 1};
	\end{tikzpicture}
\end{figure*}

\begin{figure*}[ht]
	\centering
	\begin{tikzpicture}
		\node [mybox] (box){
			\begin{minipage}{1.9\columnwidth} % Restrict to one column width
				\footnotesize
				
%				\setstretch{1.3} % Line spacing
				
				\begin{flushleft}
					\textbf{RareAgents (Llama-3.1-70B): }
					
					\texttt{\textbraceleft\textbraceleft Comprehensive decision-making based on MDT discussions, memory retrieval results, and tool feedback\textbraceright\textbraceright}\\
					\includegraphics[width=0.45cm]{assets/apa.png} \ \textbf{Attending Physician Agent: } Considering the above information, these are the top 10 diagnoses:\\
					\includegraphics[width=0.26cm]{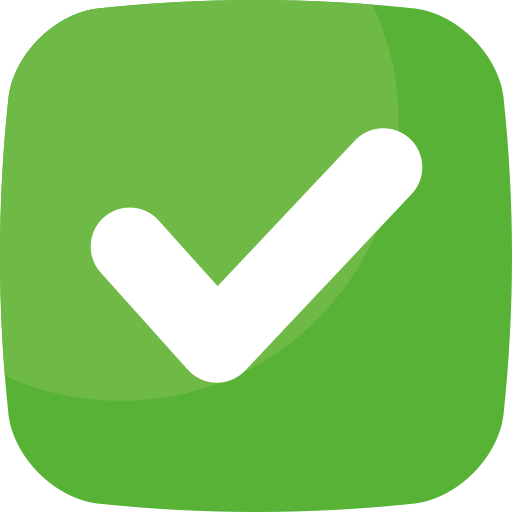} \ 1. Brugada Syndrome. \\
					2. Muscular dystrophy-dystroglycanopathy (congenital with brain and eye anomalies), type A, 3  \\
					3. Multiple system atrophy; MSA/Multiple system atrophy/Multiple system atrophy 1, susceptibility to  \\
					4. Catecholaminergic polymorphic ventricular tachycardia, CPVT/Catecholaminergic polymorphic ventricular tachycardia/Ventricular tachycardia, catecholaminergic polymorphic, 1, with or without atrial dysfunction and/or dilated cardiomyopathy  \\
					5. NKX6-2-related autosomal recessive hypomyelinating leukodystrophy/Spastic ataxia 8, autosomal recessive, with hypomyelinating leukodystrophy  \\
					6. Muscular dystrophy, limb-girdle, autosomal recessive 25  \\
					7. Glycogen storage disease II  \\
					8. Danon disease  \\
					9. Polyglucosan body myopathy 1 with or without immunodeficiency  \\
					10. Myopathy, myofibrillar, 7

				\end{flushleft}
			\end{minipage}
		};
		\node[fancytitle, rounded corners, right=10pt] at (box.north west) {Case 1};
	\end{tikzpicture}
\end{figure*}

\begin{figure*}[h!]
	\centering
	\begin{tikzpicture}
		\node [mybox] (box){
			\begin{minipage}{1.9\columnwidth} % Restrict to one column width
				\footnotesize
				
%				\setstretch{1.3} % Line spacing
				
				\begin{flushleft}
					\includegraphics[width=0.4cm]{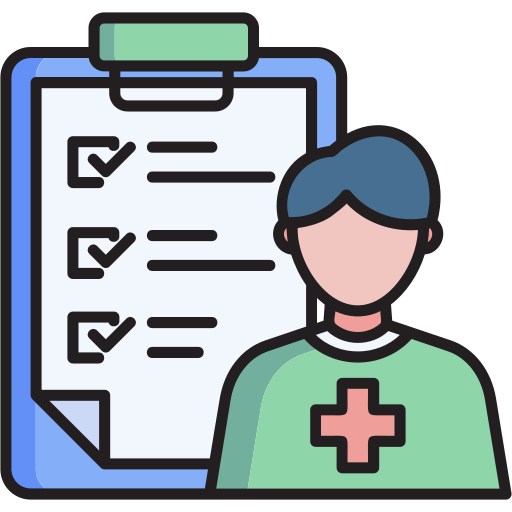} \ \textbf{Patient Agent:}
					I was diagnosed the following diseases: Encounter for antineoplastic chemotherapy; Primary central nervous system lymphoma, unspecified site, extranodal and solid organ sites; Other specified disorders of metabolism; Diabetes insipidus; Nonspecific reaction to tuberculin skin test without active tuberculosis; Epilepsy, unspecified, without mention of intractable epilepsy; Dysthymic disorder; Arthropathy, unspecified, site unspecified; Sleep disturbance, unspecified; Personal history of other infectious and parasitic diseases; Long-term (current) use of steroids, and I have received the following procedures: Injection or infusion of cancer chemotherapeutic substance. I would like to request you to provide the most appropriate combination of medications for me. \\
					\texttt{\textbraceleft\textbraceleft 122 medication candidates\textbraceright\textbraceright}\\
					
					\hdashrule[0.5ex]{146mm}{1pt}{2mm}
					
					\includegraphics[width=0.45cm]{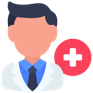} \ \textbf{Single-Agent (Llama-3.1-70B): }
					TREATMENT:  \\
					\includegraphics[width=0.26cm]{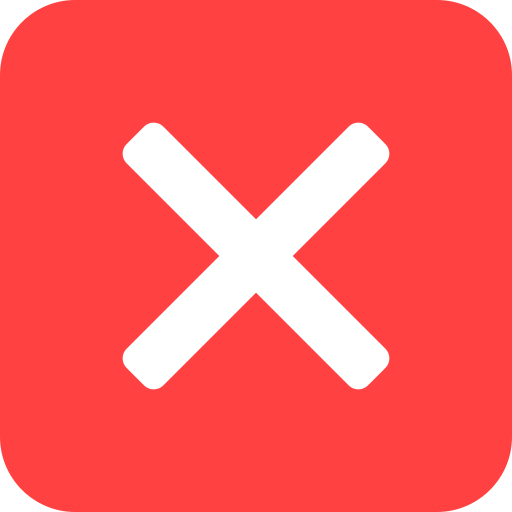} \ 1. *NF* Epirubicin  \\
					\includegraphics[width=0.26cm]{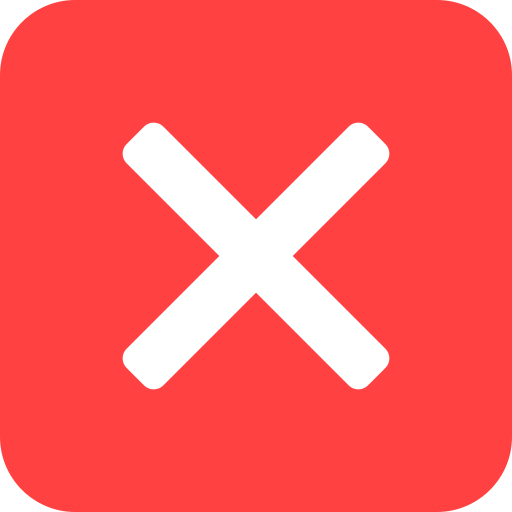} \ 2. Mesna  \\
					\includegraphics[width=0.26cm]{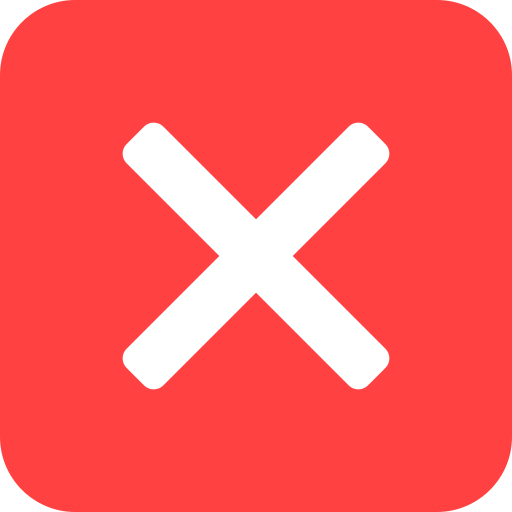} \ 3. Levothyroxine Sodium  \\
					\includegraphics[width=0.26cm]{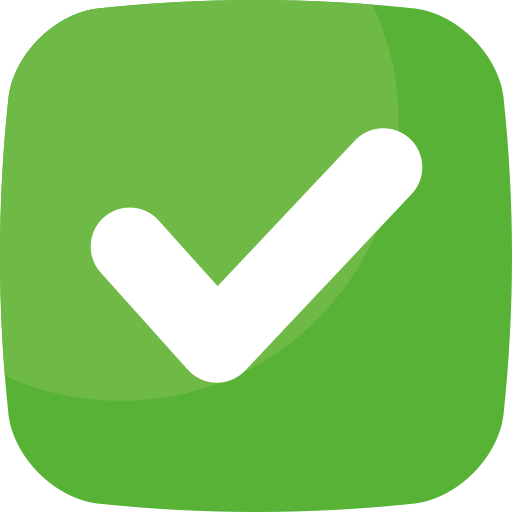} \ 4. Desmopressin Nasal  \\
					\includegraphics[width=0.26cm]{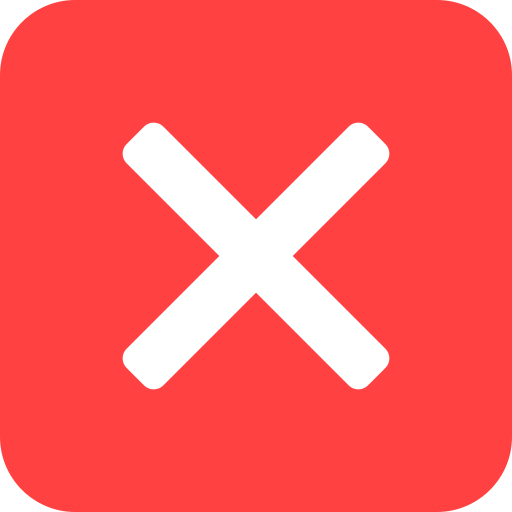} \ 5. Phenoxybenzamine HCl  \\
					\includegraphics[width=0.26cm]{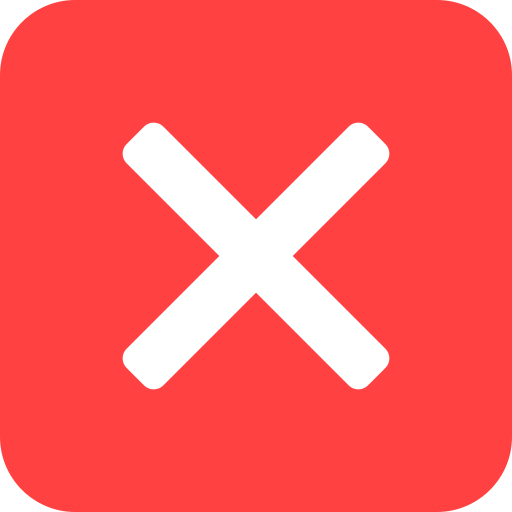} \ 6. Metoprolol Tartrate  \\
					\includegraphics[width=0.26cm]{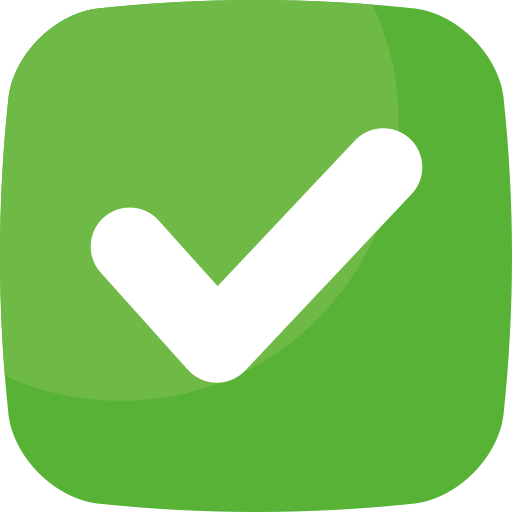} \ 7. Lorazepam  \\
					\includegraphics[width=0.26cm]{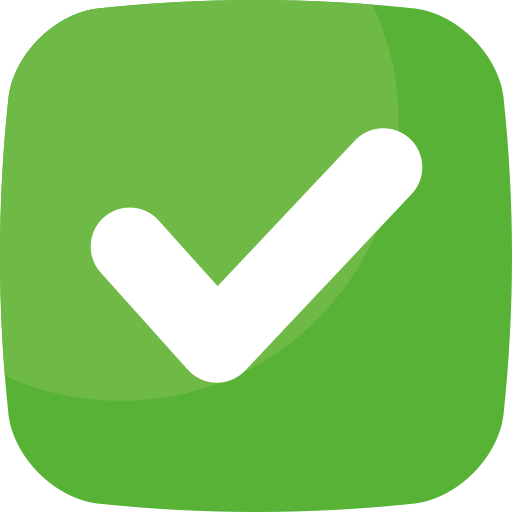} \ 8. LeVETiracetam  \\
					\includegraphics[width=0.26cm]{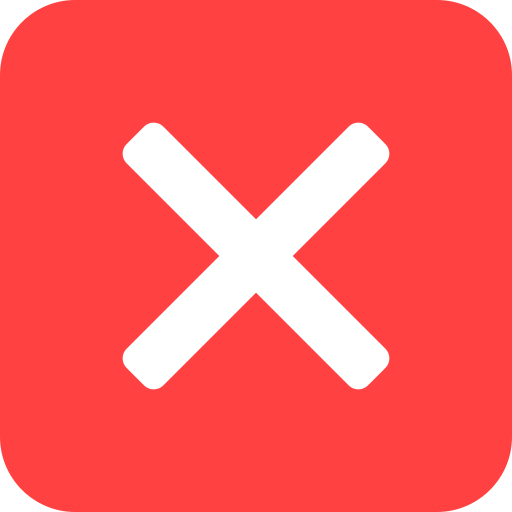} \ 9. phenobarbital  \\
					\includegraphics[width=0.26cm]{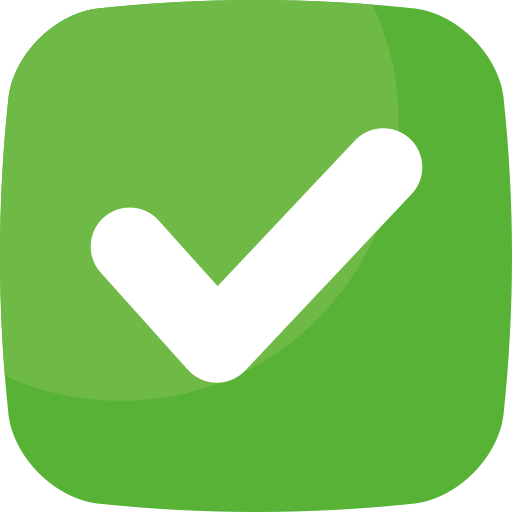} \ 10. Sertraline  \\
					\includegraphics[width=0.26cm]{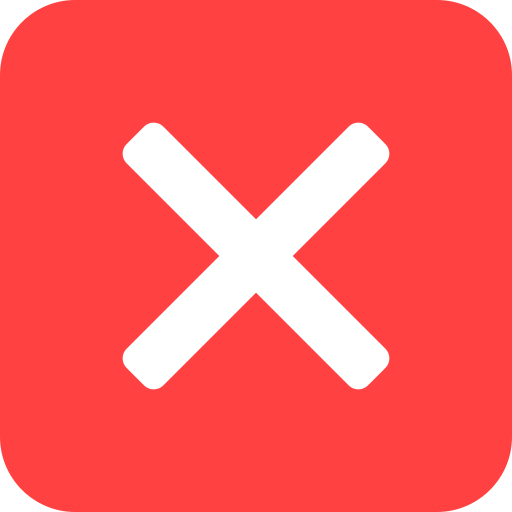} \ 11. prednisolone  \\
					\includegraphics[width=0.26cm]{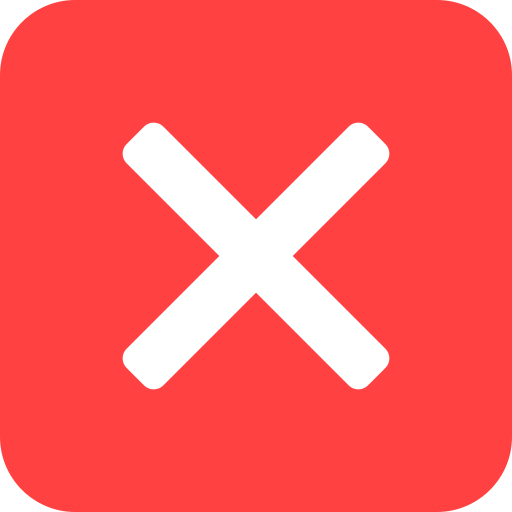} \ 12. fludrocortisone  \\
					\includegraphics[width=0.26cm]{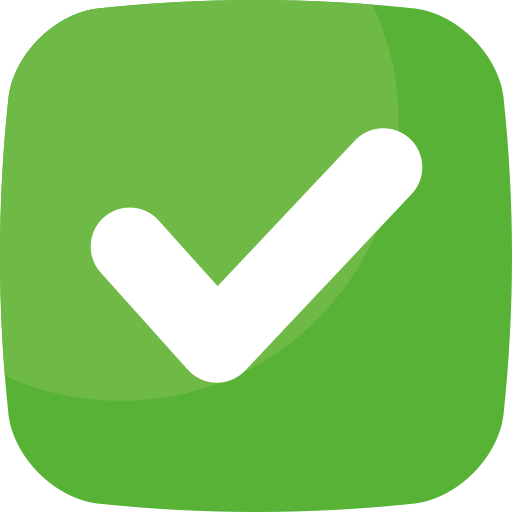} \ 13. Potassium Chloride  \\
					\includegraphics[width=0.26cm]{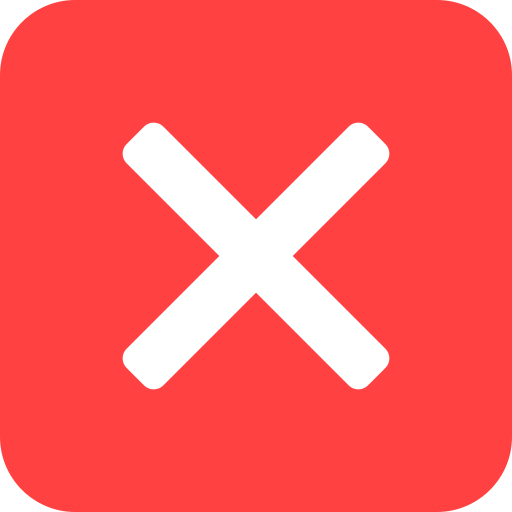} \ 14. Potassium Acetate  \\
					\includegraphics[width=0.26cm]{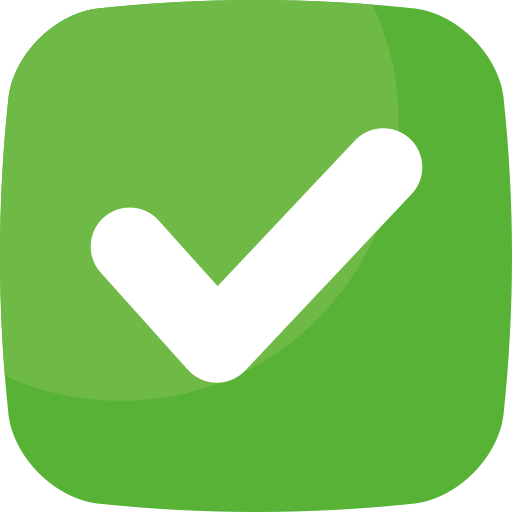} \ 15. Sodium Bicarbonate  \\
					\includegraphics[width=0.26cm]{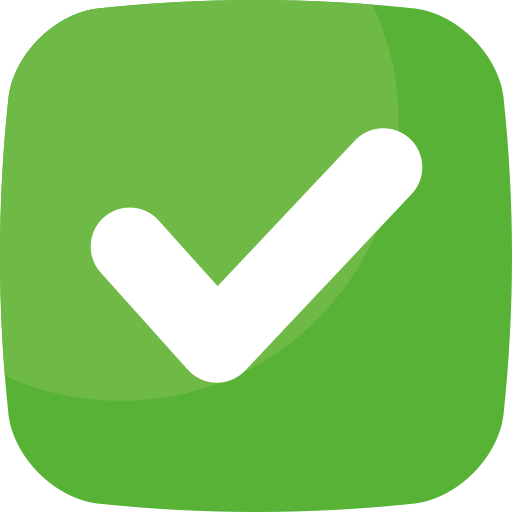} \ 16. Ondansetron  \\
					\includegraphics[width=0.26cm]{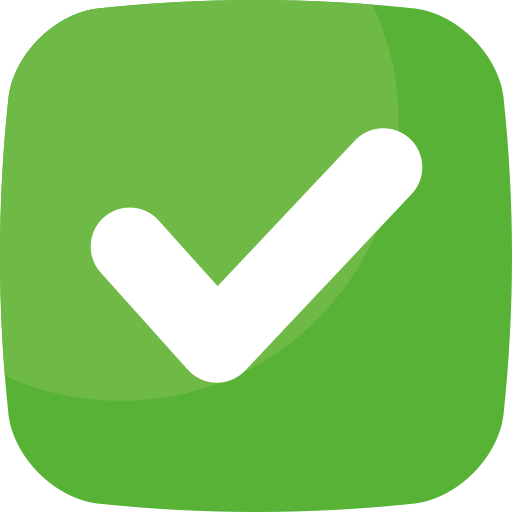} \ 17. Prochlorperazine  \\
					\includegraphics[width=0.26cm]{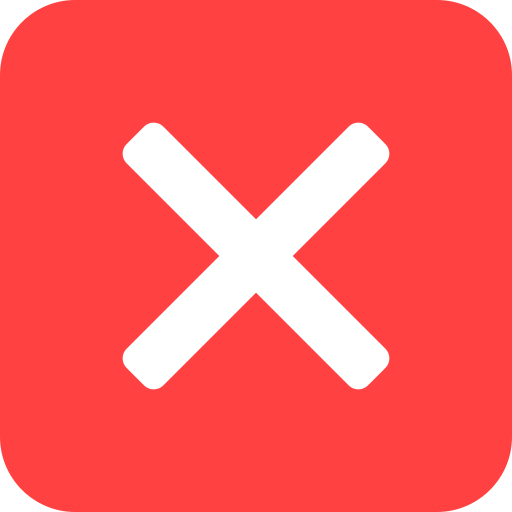} \ 18. Dantrolene Sodium  \\
					\includegraphics[width=0.26cm]{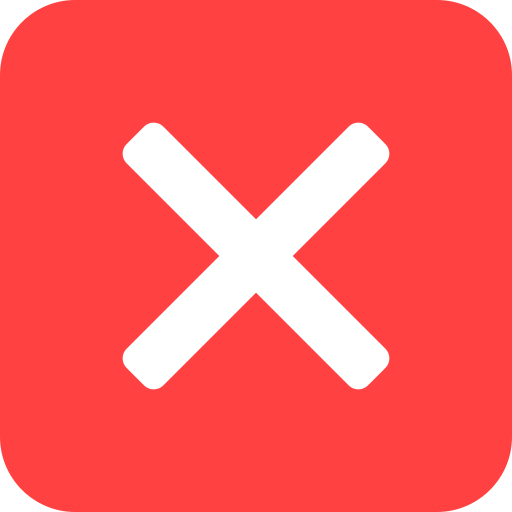} \ 19. Melphalan  \\
					\includegraphics[width=0.26cm]{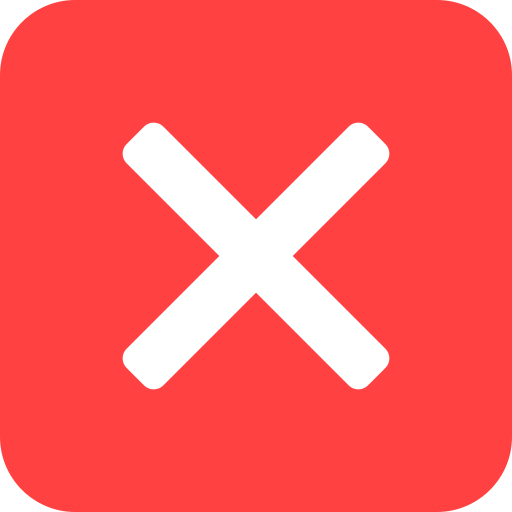} \ 20. interferon alfa-2b 6 million unit/mL  \\
					\includegraphics[width=0.26cm]{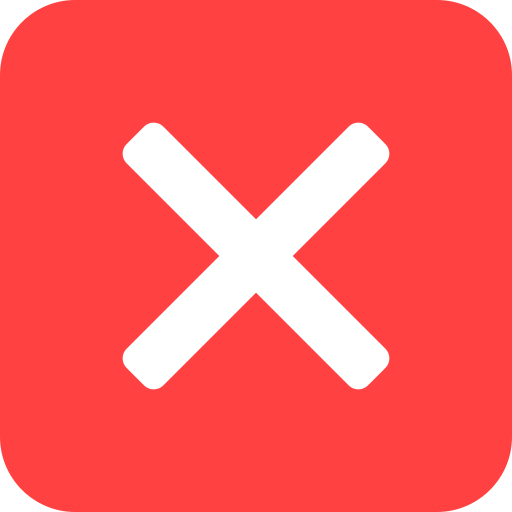} \ 21. Aldesleukin \\
					\textcolor{red}{\texttt{\textbraceleft\textbraceleft Absence of ground truth medication: Ranitidine, 0.9\% Sodium Chloride,  5\% Dextrosecandidates, Propranolol, Heparin, Lisinopril, DiphenhydrAMINE, Calcium Carbonate, traZODONE, OxycoDONE (Immediate Release), PredniSONE, Methotrexate, Isoniazid\textbraceright\textbraceright}}

				\end{flushleft}
			\end{minipage}
		};
		\node[fancytitle, rounded corners, right=10pt] at (box.north west) {Case 2};
	\end{tikzpicture}
\end{figure*}

\begin{figure*}[ht]
	\centering
	\begin{tikzpicture}
		\node [mybox] (box){
			\begin{minipage}{1.9\columnwidth} % Restrict to one column width
				\footnotesize
				
%				\setstretch{1.3} % Line spacing
				
				\begin{flushleft}
					\textbf{RareAgents (Llama-3.1-70B): }
					
					\includegraphics[width=0.45cm]{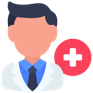} \ \textbf{Attending Physician Agent: } Based on the patient’s symptoms, I would like to request the following specialists to form a multi-disciplinary team: 
					Nuclear Medicine, Pathology, Nephrology, Urology, Neurology, Oncology, Hematology, Radiology, Interventional Radiology, Neurosurgery, Infectious Diseases, Rheumatology, Rehabilitation Medicine, Psychiatry, Clinical Nutrition, Pharmacy, Orthopedic Surgery, Hematology, Dentistry, Anesthesiology, Endocrinology, Laboratory Medicine, Traditional Chinese Medicine, Allergy and Immunology.\\
					\texttt{\textbraceleft\textbraceleft Multiple rounds of discussions by multidisciplinary team (MDT)\textbraceright\textbraceright}\\
					...\\ 
					\includegraphics[width=0.35cm]{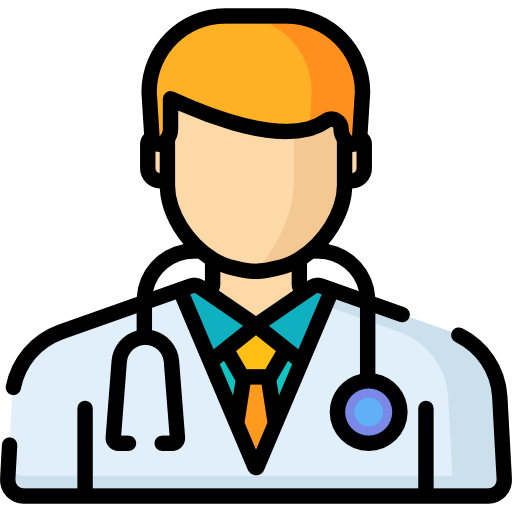} \ \textbf{Nuclear Medicine Physician Agent: } Based on the patient's diagnosis and medical history, I will provide my expert opinion on the treatment. As the patient has been diagnosed with primary central nervous system lymphoma, methotrexate is a common chemotherapeutic agent used to treat this condition. As the patient has epilepsy, LeVETiracetam is an antiepileptic medication that can help control seizures. As the patient has diabetes insipidus, desmopressin nasal spray can help regulate fluid balance and reduce urine production. As the patient has dysthymic disorder, sertraline is an antidepressant medication that can help manage symptoms of depression....\\
					\includegraphics[width=0.35cm]{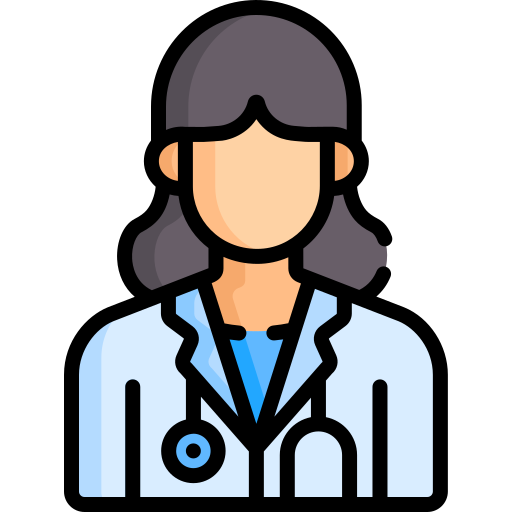} \ \textbf{Pathology Physician Agent: } Based on the patient's diagnosis and medical history, the following medications are recommended: 1. **Methotrexate**: As the patient is undergoing antineoplastic chemotherapy, methotrexate is a common chemotherapeutic agent used to treat various types of cancer, including lymphoma. 2. **LeVETiracetam**: As the patient has epilepsy, LeVETiracetam is an antiepileptic medication that can help control seizures. 3. **Desmopressin Nasal**: As the patient has diabetes insipidus, desmopressin nasal spray can help regulate water balance in the body. 4. **Sertraline**: As the patient has dysthymic disorder, sertraline is an antidepressant medication that can help manage symptoms of depression....\\
					\includegraphics[width=0.35cm]{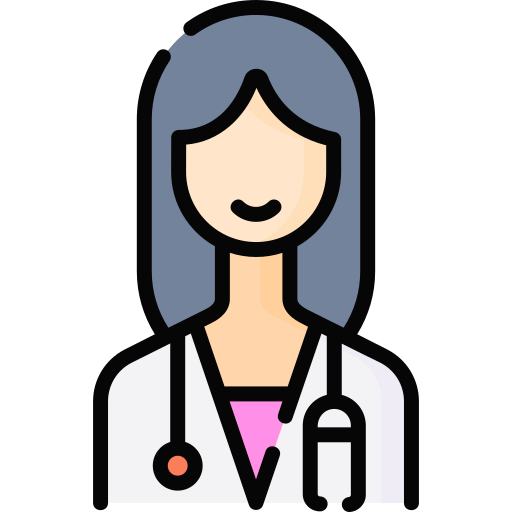} \ \textbf{Nephrology Physician Agent: } **Treatment Plan:** **Chemotherapy:** Continue with the current antineoplastic chemotherapy regimen as prescribed by the oncologist. **Diabetes Insipidus:** Desmopressin Nasal Spray: 1-2 sprays (10-20 mcg) intranasally at bedtime, as needed, to control nocturia and polyuria. **Epilepsy:** LeVETiracetam (Keppra): 500-1000 mg orally twice daily, as needed, to control seizures. **Dysthymic Disorder:** Sertraline (Zoloft): 50-100 mg orally once daily, as needed, to manage depressive symptoms. **Pain Management:** Acetaminophen (Tylenol): 650-1000 mg orally every 4-6 hours, as needed, for pain relief. **Steroid-Induced Side Effects:** Monitor for signs of steroid-induced side effects, such as hyperglycemia, hypertension, and osteoporosis....\\
					\includegraphics[width=0.35cm]{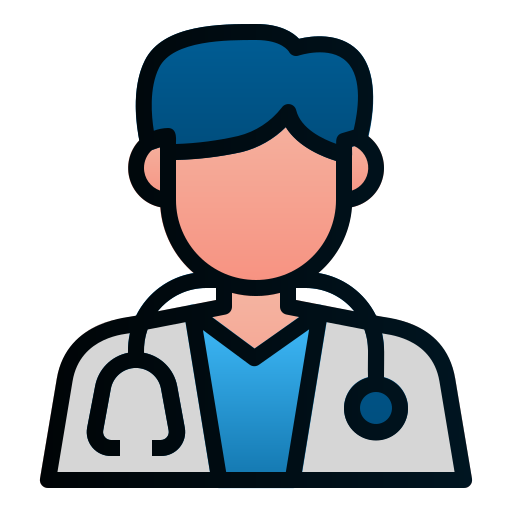} \ \textbf{Urology Physician Agent: } Based on the patient's diagnosis and medical history, I will provide a treatment plan that addresses the various conditions and medications. **Primary central nervous system lymphoma, unspecified site, extranodal and solid organ sites:** The patient is already receiving antineoplastic chemotherapy, which is the standard treatment for primary central nervous system lymphoma. I recommend continuing the chemotherapy regimen as prescribed. **Diabetes insipidus:** The patient is likely experiencing symptoms of diabetes insipidus due to the chemotherapy treatment. I recommend Desmopressin Nasal to manage the symptoms of diabetes insipidus. **Other specified disorders of metabolism:** The patient's metabolic disorders may be related to the chemotherapy treatment or underlying conditions. I recommend monitoring the patient's metabolic parameters and adjusting the treatment plan as needed. **Nonspecific reaction to tuberculin skin test without active tuberculosis:** The patient's reaction to the tuberculin skin test indicates a possible exposure to tuberculosis. I recommend monitoring the patient for signs of tuberculosis and considering prophylactic treatment if necessary....\\
					...\\

				\end{flushleft}
			\end{minipage}
		};
		\node[fancytitle, rounded corners, right=10pt] at (box.north west) {Case 2};
	\end{tikzpicture}
\end{figure*}

\begin{figure*}[ht]
	\centering
	\begin{tikzpicture}
		\node [mybox] (box){
			\begin{minipage}{1.9\columnwidth} % Restrict to one column width
				\footnotesize
				
%				\setstretch{1.3} % Line spacing
				
				\begin{flushleft}
					\textbf{RareAgents (Llama-3.1-70B): }
					
					\texttt{\textbraceleft\textbraceleft Dynamic retrieval of patient's former visits from long-term memory\textbraceright\textbraceright}\\
					\includegraphics[width=0.4cm]{assets/memory.png} \ \textbf{Memory Retrieve Results: } Consider these previous visits for reference: \\
					\textbf{First admission:} Patient with a diagnosis of Encounter for antineoplastic chemotherapy; Diabetes insipidus; Diffuse large B-cell lymphoma, extranodal and solid organ sites; Calculus of kidney; Personal history of nicotine dependence; Anxiety disorder, unspecified; Insomnia, unspecified; Essential (primary) hypertension; Presence of artificial hip joint, bilateral and a history of procedures including Introduction of Other Antineoplastic into Central Vein, Percutaneous Approach has a prior medication record of: ['Magnesium Sulfate', '5\% Dextrose', 'Sodium Bicarbonate', 'LamoTRIgine', 'Methotrexate', 'Heparin'].\\
					\textbf{Second admission:} Patient with a diagnosis of Primary central nervous system lymphoma, unspecified site, extranodal and solid organ sites; Cerebral edema; Diabetes insipidus; Other convulsions; Unspecified essential hypertension; Hip joint replacement; Personal history of tobacco use; Personal history of tuberculosis and a history of procedures including Injection or infusion of cancer chemotherapeutic substance has a prior medication record of: ['Sodium Bicarbonate', '0.9\% Sodium Chloride', 'Diazepam', 'LeVETiracetam', 'Heparin', '5\% Dextrose', 'Dexamethasone', 'TraZODone', 'Acetaminophen', 'Methotrexate', 'Propranolol', 'Calcium Carbonate', 'Senna', 'Sertraline'].\\
					\textbf{Third admission:} Patient with a diagnosis of Encounter for antineoplastic chemotherapy; Other malignant lymphomas, unspecified site, extranodal and solid organ sites; Diabetes insipidus; Unspecified essential hypertension; Hip joint replacement; Epilepsy, unspecified, without mention of intractable epilepsy; Anxiety state, unspecified; Sleep disturbance, unspecified and a history of procedures including Injection or infusion of cancer chemotherapeutic substance has a prior medication record of: ['traZODONE', 'Famotidine', 'Prochlorperazine', 'Dexamethasone', '0.9\% Sodium Chloride', 'Sodium Bicarbonate', 'Clonazepam', 'Senna', 'Ondansetron', '5\% Dextrose', 'Docusate Sodium', 'LeVETiracetam', 'Acetaminophen', 'Desmopressin Nasal', 'Lisinopril', 'Methotrexate'].\\ 
					...\\
					\texttt{\textbraceleft\textbraceleft Feedback after the function calls the therapeutic tool\textbraceright\textbraceright}\\
					
					\includegraphics[width=0.55cm]{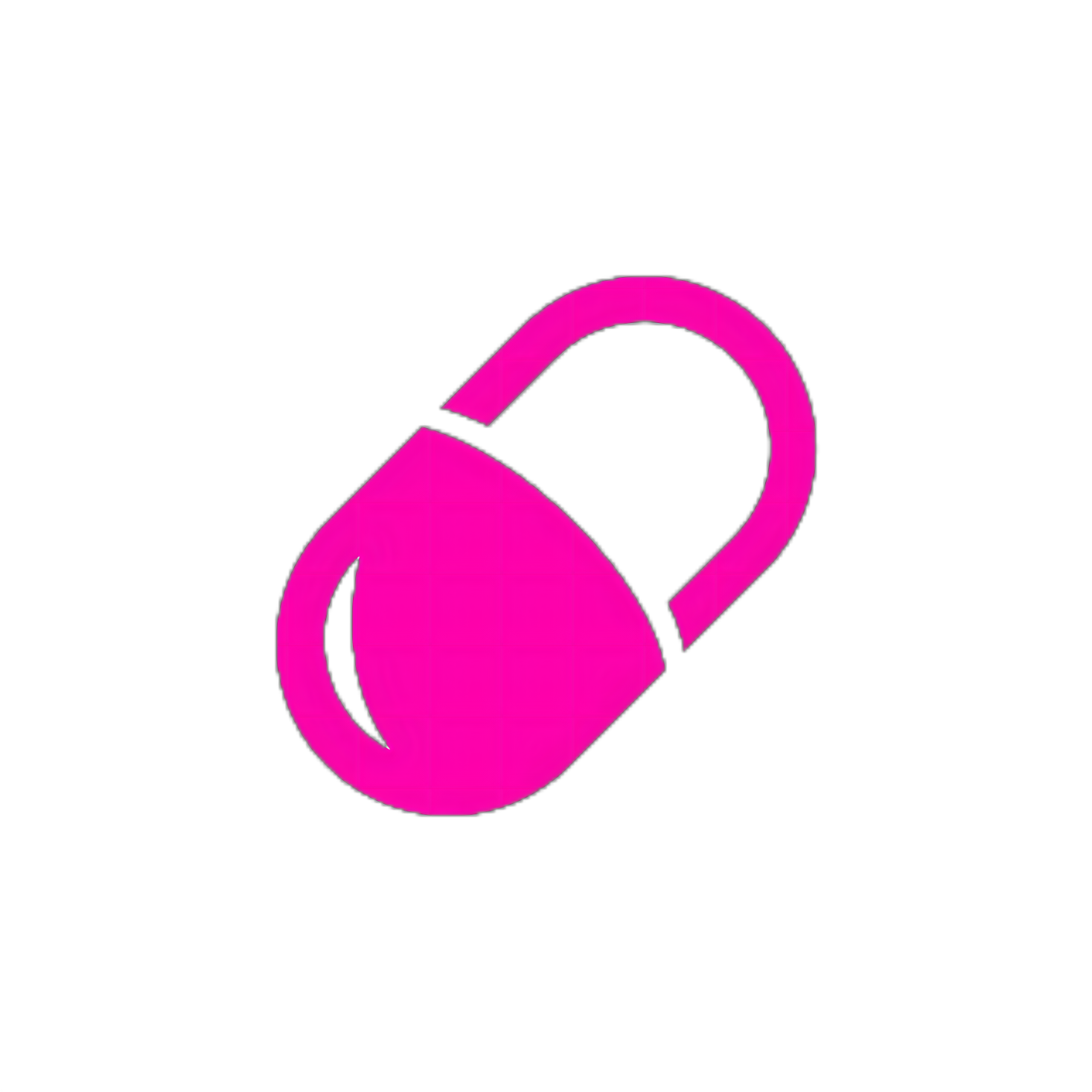} \ \textbf{DrugBank: } More detailed drug information is provided below:\\
					Ondansetron: A competitive serotonin type 3 receptor antagonist. It is effective in the treatment of nausea and vomiting caused by cytotoxic chemotherapy drugs, including cisplatin, and has reported anxiolytic and neuroleptic properties. \\
					Tolvaptan: Tolvaptan is used to treat low blood sodium levels (hyponatremia) associated with various conditions like congestive heart failure, cirrhosis, and syndrome of inappropriate antidiuretic hormones. \\
					Ranitidine: A non-imidazole blocker of those histamine receptors that mediate gastric secretion (H2 receptors). It is used to treat gastrointestinal ulcers. \\
					Penicillamine: Penicillamine is a pharmaceutical of the chelator class. The pharmaceutical form is D-penicillamine, as L-penicillamine is toxic (it inhibits the action of pyridoxine). It is an $\alpha$-amino acid metabolite of penicillin, although it has no antibiotic properties. \\
					Pyridoxine: Pyridoxine is the 4-methanol form of vitamin B6, an important water-soluble vitamin that is naturally present in many foods. As its classification as a vitamin implies, Vitamin B6 (and pyridoxine) are essential nutrients required for normal functioning of many biological systems within the body. While many plants and microorganisms are able to synthesize pyridoxine through endogenous biological processes, animals must obtain it through their diet. \\
					...\\
					
					\includegraphics[width=0.35cm]{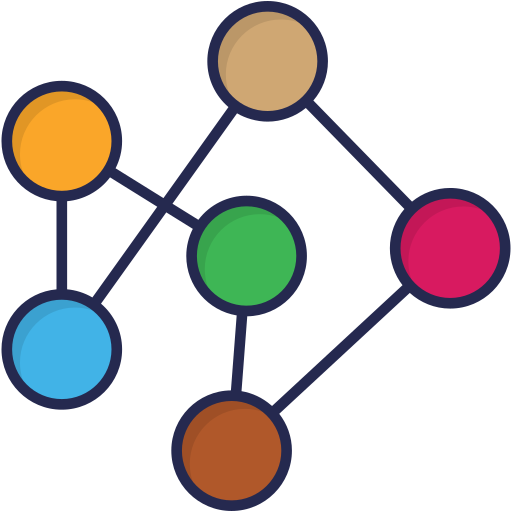} \ \textbf{DDI-graph: } The following are potential drug interactions:\\
					Prochlorperazine and Potassium Chloride may cause Myringitis. \\
					Sertraline and Prochlorperazine may cause meibomianitis. \\
					Mesna and Metoprolol Tartrate may cause corticosteroid therapy. \\
					Metoprolol Tartrate and Potassium Chloride may cause corticosteroid therapy. \\
					Sertraline and fludrocortisone may cause meibomianitis. \\
					... \\

				\end{flushleft}
			\end{minipage}
		};
		\node[fancytitle, rounded corners, right=10pt] at (box.north west) {Case 2};
	\end{tikzpicture}
\end{figure*}

\begin{figure*}[ht]
	\centering
	\begin{tikzpicture}
		\node [mybox] (box){
			\begin{minipage}{1.9\columnwidth} % Restrict to one column width
				\footnotesize
				
%				\setstretch{1.3} % Line spacing
				
				\begin{flushleft}
					\textbf{RareAgents (Llama-3.1-70B): }
					
					\texttt{\textbraceleft\textbraceleft Comprehensive decision-making based on MDT discussions, memory retrieval results, and tool feedback\textbraceright\textbraceright}\\
					\includegraphics[width=0.45cm]{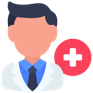} \ \textbf{Attending Physician Agent: } Considering the above information, I recommend the following combination of medications:\\
					\includegraphics[width=0.26cm]{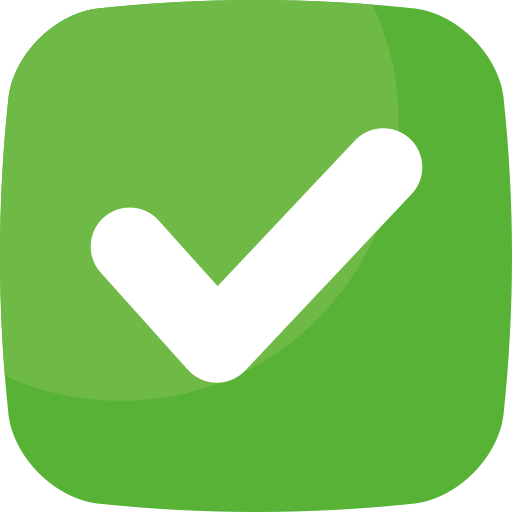} \ 1. Ondansetron  \\
					\includegraphics[width=0.26cm]{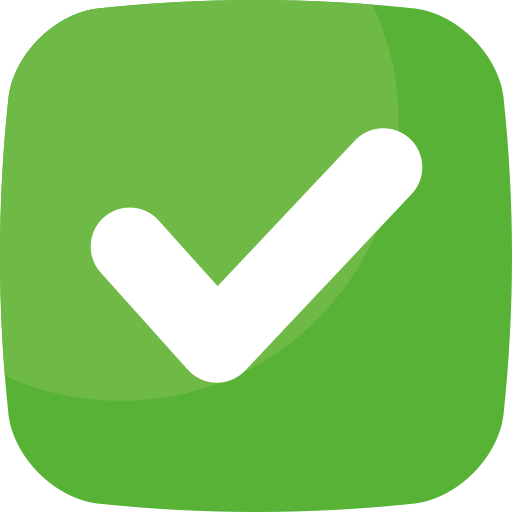} \ 2. Methotrexate  \\
					\includegraphics[width=0.26cm]{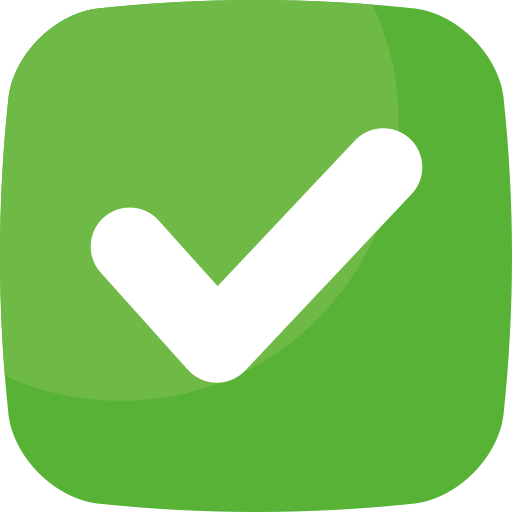} \ 3. LeVETiracetam  \\
					\includegraphics[width=0.26cm]{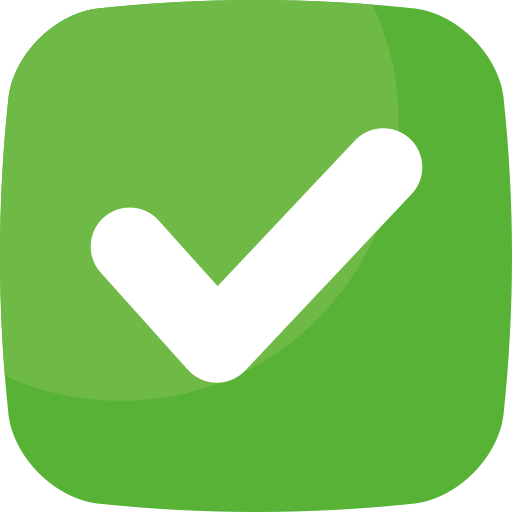} \ 4. Sertraline  \\
					\includegraphics[width=0.26cm]{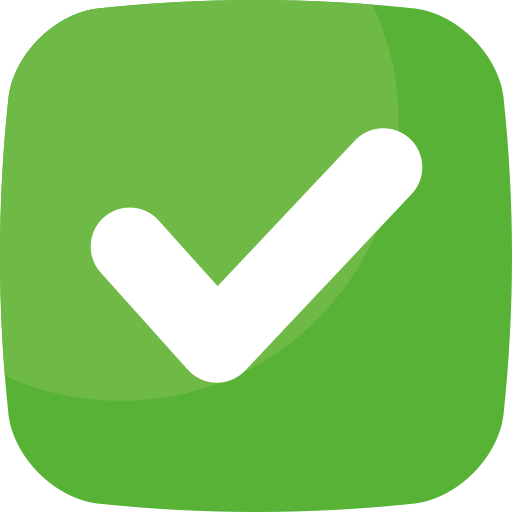} \ 5. Propranolol  \\
					\includegraphics[width=0.26cm]{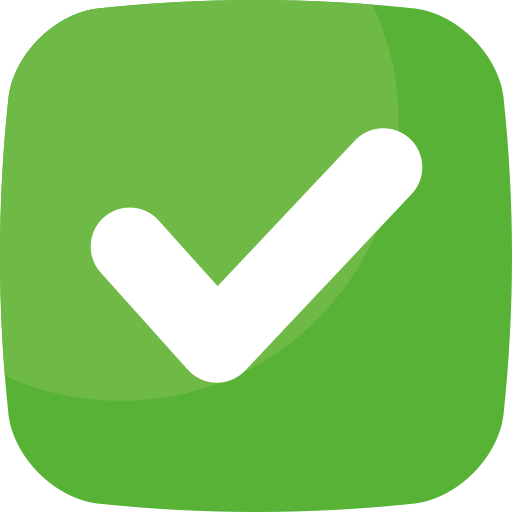} \ 6. Sodium Bicarbonate  \\
					\includegraphics[width=0.26cm]{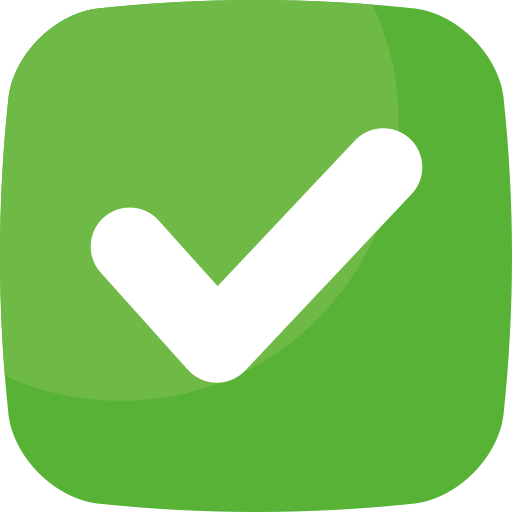} \ 7. 5\% Dextrose  \\
					\includegraphics[width=0.26cm]{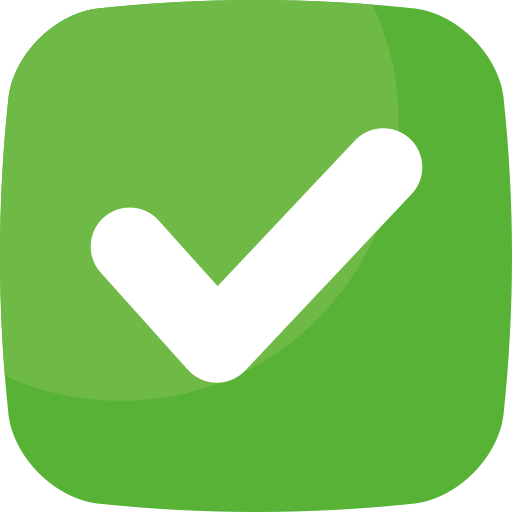} \ 8. 0.9\% Sodium Chloride  \\
					\includegraphics[width=0.26cm]{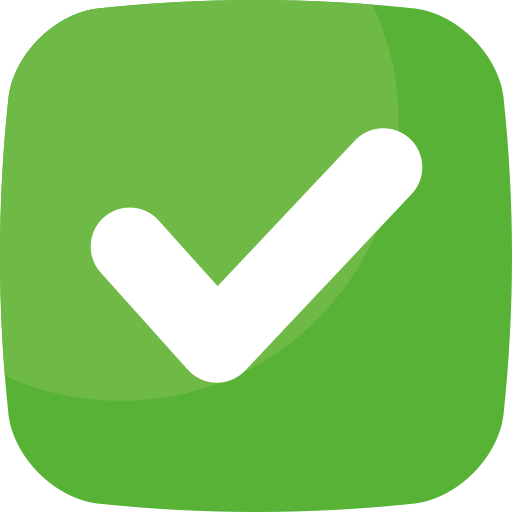} \ 9. Heparin  \\
					\includegraphics[width=0.26cm]{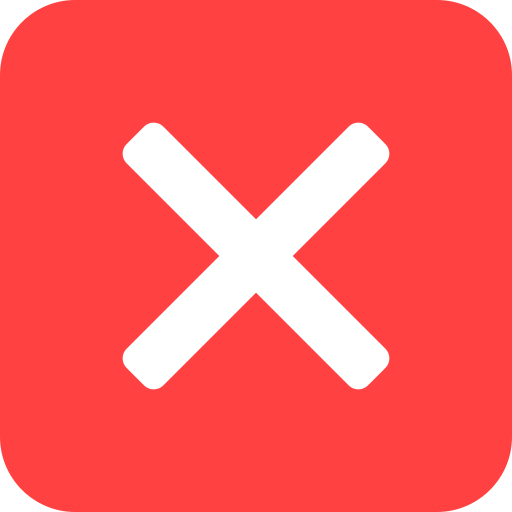} \ 10. Acetaminophen  \\
					\includegraphics[width=0.26cm]{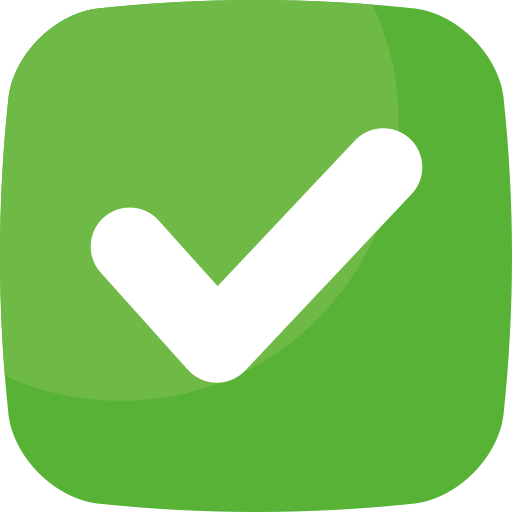} \ 11. OxycoDONE (Immediate Release)  \\
					\includegraphics[width=0.26cm]{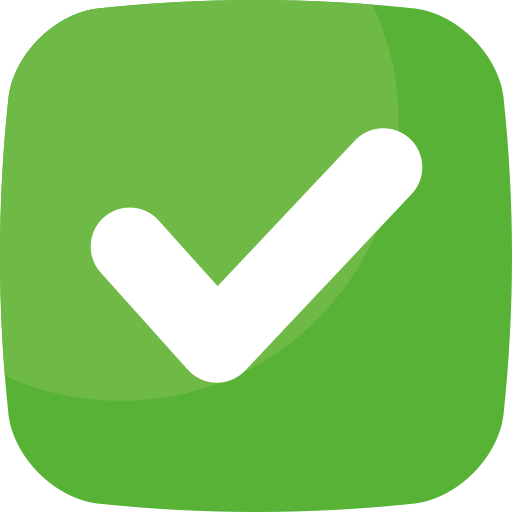} \ 12. traZODONE  \\
					\includegraphics[width=0.26cm]{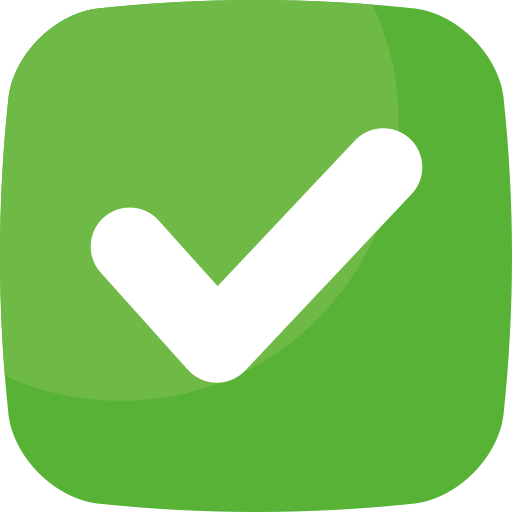} \ 13. Desmopressin Nasal  \\
					\includegraphics[width=0.26cm]{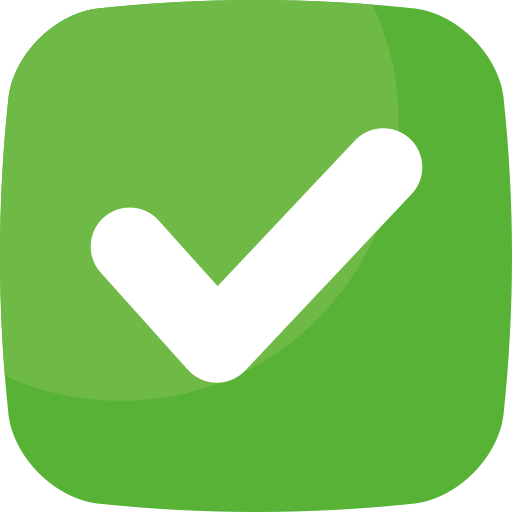} \ 14. Lisinopril  \\
					\includegraphics[width=0.26cm]{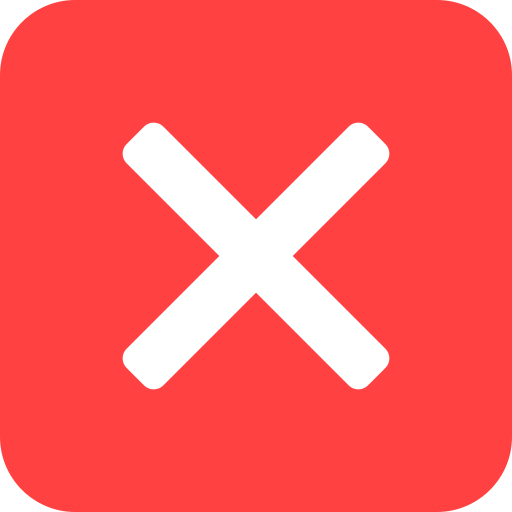} \ 15. Dexamethasone  \\
					\includegraphics[width=0.26cm]{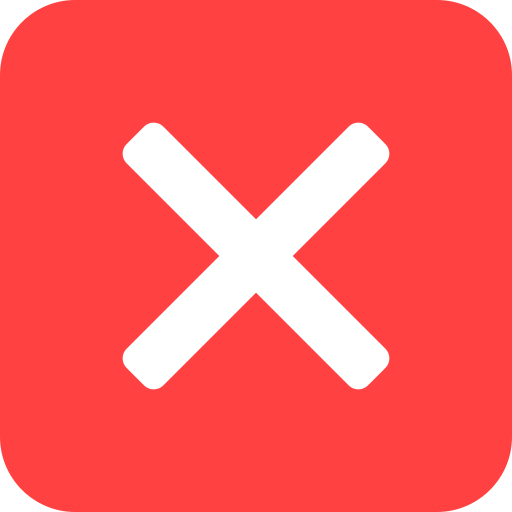} \ 16. Clonazepam  \\
					\includegraphics[width=0.26cm]{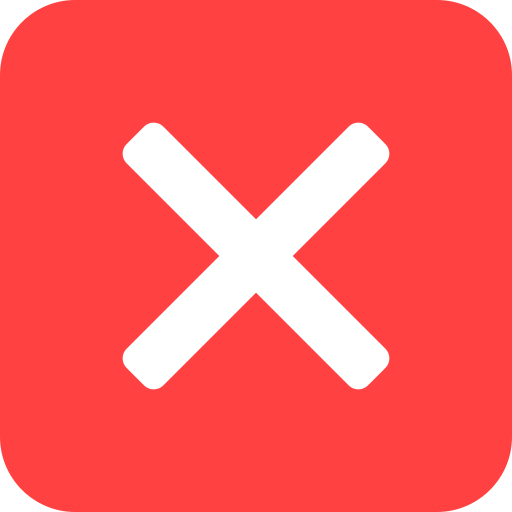} \ 17. Furosemide  \\
					\includegraphics[width=0.26cm]{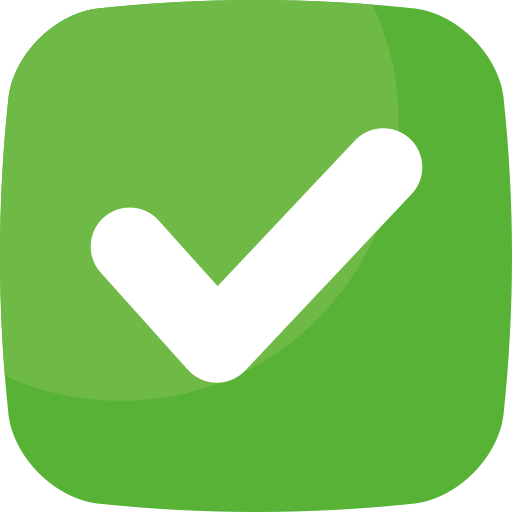} \ 18. Potassium Chloride Replacement (Oncology)  \\
					\includegraphics[width=0.26cm]{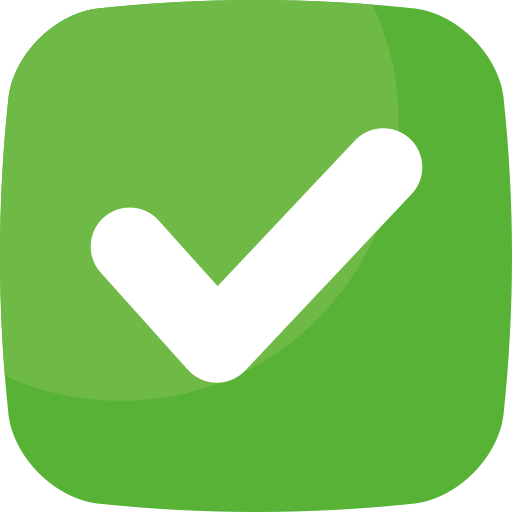} \ 19. Calcium Carbonate  \\
					\includegraphics[width=0.26cm]{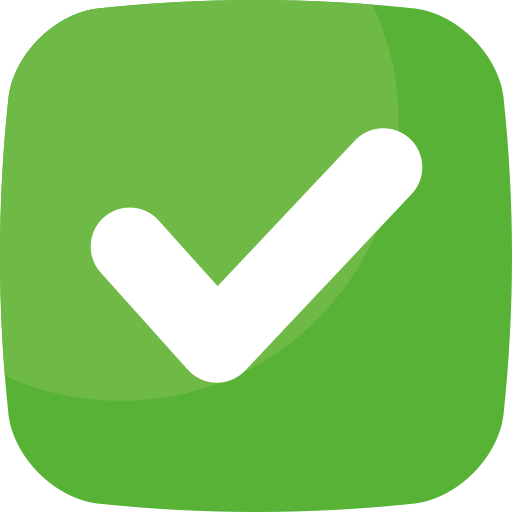} \ 20. Lorazepam  \\
					\includegraphics[width=0.26cm]{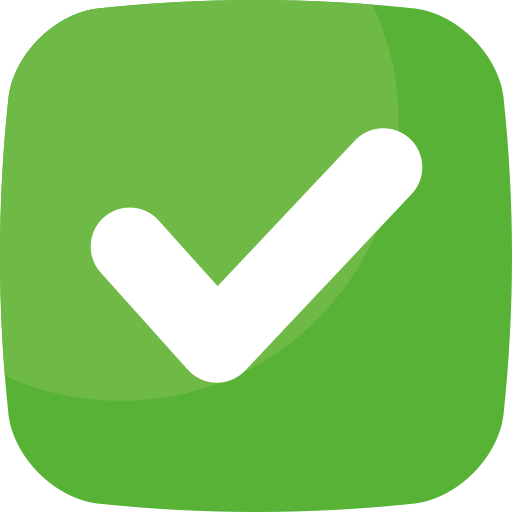} \ 21. PredniSONE \\
					\textcolor{red}{\texttt{\textbraceleft\textbraceleft Absence of ground truth medication: Ranitidine,  Prochlorperazine, DiphenhydrAMINE, Isoniazid\textbraceright\textbraceright}}

				\end{flushleft}
			\end{minipage}
		};
		\node[fancytitle, rounded corners, right=10pt] at (box.north west) {Case 2};
	\end{tikzpicture}
\end{figure*}

\end{document}